%% file: sample-sigconf-authordraft.tex
\PassOptionsToPackage{dvipsnames,table}{xcolor}
\documentclass[sigconf,balance=false]{acmart}

\usepackage{balance}
\AtBeginDocument{%
  }

\copyrightyear{2026}
\acmYear{2026}
\setcopyright{cc}
\setcctype{by}

\acmConference[MM '26]{Proceedings of the 34th ACM International Conference on Multimedia}{November 10--14, 2026}{Rio de Janeiro, Brazil}

\acmBooktitle{Proceedings of the 34th ACM International Conference on Multimedia (MM '26), November 10--14, 2026, Rio de Janeiro, Brazil}

\acmDOI{10.1145/3767308.3835435}

\acmISBN{979-8-4007-2213-4/2026/11}
 
\acmSubmissionID{2962}

\usepackage{subcaption}

\usepackage[ruled,vlined]{algorithm2e}

\usepackage{multirow}
\usepackage{pifont} 

\newcommand{\cmark}{\ding{51}} 
\newcommand{\xmark}{\ding{55}} 
\newcommand{\pmark}{\ding{108}} 
\newcommand{\myparagraph}[1]{\noindent{\bf{#1}}~~}
\definecolor{GroupGreen}{RGB}{210,240,210}
\definecolor{GroupBlue}{RGB}{210,225,245}
\definecolor{GroupGray}{gray}{0.92}

\usepackage[capitalize,noabbrev]{cleveref}

\begin{document}

\title{SlideGen: Collaborative Multimodal Agents for Scientific Slide Generation}


\author{Xin Liang}
\authornote{Both authors contributed equally to this research.}
\affiliation{%
  \institution{Stony Brook University}
  \city{Stony Brook}
  \state{New York}
  \country{USA}
}

\author{Zhilin Zhang}
\authornotemark[1]
\affiliation{%
  \institution{New York University}
  \city{New York}
  \state{New York}
  \country{USA}
}

\author{Xiang Zhang}
\affiliation{%
  \institution{University of British Columbia}
  \city{Vancouver}
  \state{British Columbia}
  \country{Canada}
}

\author{Haoran Su}
\affiliation{%
  \institution{New York University}
  \city{New York}
  \state{New York}
  \country{USA}
}

\author{Yiwei Xu}
\affiliation{%
  \institution{University of California, Los Angeles}
  \city{Los Angeles}
  \state{California}
  \country{USA}
}

\author{Siqi Sun}
\affiliation{%
  \institution{Fudan University}
  \city{Shanghai}
  \country{China}
}

\author{Chenyu You}
\authornote{Corresponding author.}
\affiliation{%
  \institution{Stony Brook University}
  \city{Stony Brook}
  \state{New York}
  \country{USA}
}

\renewcommand{\shortauthors}{Liang et al.}
 
 \renewcommand{\shortauthors}{Xin Liang et al.}


\input{sections/0_abstract}



\maketitle

\input{sections/1_intro}
\input{sections/2_related_work}
\input{sections/3_method}
\input{sections/4_experiments}
\input{sections/5_conclusion}

\balance
\bibliographystyle{ACM-Reference-Format}
\bibliography{sample-base}

\appendix
\clearpage
\input{sections/appendix}

\end{document}

%% file: sections/0_abstract.tex
\begin{abstract}
Creating presentation slides from scientific papers is not simply a matter of summarizing paragraphs. 
A presenter is required to decide what story to tell, which figures and equations to highlight, and how to arrange them into pages that are visually clear rather than crowded or repetitive. 
The need to jointly reason over long contexts and layout-sensitive design makes paper-to-slide generation a uniquely challenging multimodal task.
Most existing approaches, however, focus mainly on textual content selection, producing slides that often lack visual balance, narrative flow, or coherent integration of multimodal evidence.
In this work, we introduce SlideGen, a collaborative vision-language multi-agent framework that coordinates narrative planning, multimodal grounding, and layout composition. 
SlideGen assigns specialized agents to outline the presentation structure, align supporting figures and tables with key claims, generate speaker notes, and compose editable PPTX slides through a diverse layout library. 
By refining layouts at the deck level, the system produces slide decks that are both faithful to the source paper and effective as presentations.
To evaluate slide generation beyond text fidelity, we propose geometry-aware density (GAD), a metric that captures visual clutter, sparsity, and fragmentation, and shows strong agreement with human judgments. 
Evaluated across four complementary dimensions on our 200-paper benchmark, SlideGen consistently and significantly improves layout balance, content coverage, and text coherence, outperforming competitive baselines in paper-to-slide generation.
Our findings suggest that effective slide generation requires multimodal design reasoning, and that agent collaboration offers a principled bridge between document understanding and scientific communication.
\end{abstract}

%% file: sections/1_intro.tex
\section{Introduction}
\label{sec:intro}

Scientific slides are the primary medium through which research is communicated, e.g., in conference talks, classrooms, and tutorials. 
Unlike papers, slides are expected to distill complex technical content into a small number of visually organized pages, where clarity depends not only on \emph{what} is presented, but also on \emph{how} multimodal content is structured and arranged. 
An effective slide deck therefore requires narrative planning, selective emphasis, and deliberate visual design~\citep{hu2013ppsgen,fu2022doc2ppt}.

\begin{table}[t]
\caption{Comparison with existing slide generation systems across key dimensions. SlideGen stands apart in terms of all five dimensions. The complete table can be found in Appendix \cref{tab:auto-slide-comparison}.}
\vspace{-5pt}
\label{tab:auto-slide-comparison-simplified}
\centering
\setlength{\tabcolsep}{2pt}
\small
\resizebox{\linewidth}{!}{%
\begin{tabular}{l ccccc}
\toprule
\multirow{2}{*}{\textbf{Method}} &
\textbf{Content} & \textbf{Text-Fig} & \textbf{Multi} & \textbf{Output} & \textbf{User} \\
& \textbf{Struct.} & \textbf{Align.} & \textbf{Modal} & \textbf{Format} & \textbf{Editability} \\
\midrule
PPSGen~\citep{hu2013ppsgen}                         & \pmark & \xmark & \xmark & text      & \xmark \\
D2S~\citep{sun2021d2s}                              & \pmark & \cmark & \cmark & text      & \xmark \\
DOC2PPT~\citep{fu2022doc2ppt}                       & \cmark & \cmark & \cmark & PPTX/PDF  & \xmark \\
Persona-Aware D2S~\citep{mondal2024presentations}   & \cmark & \pmark & \cmark & PDF       & \xmark \\
GDP~\citep{maheshwari2024presentations}             & \cmark & \xmark & \xmark & text      & \xmark \\
DocPres~\citep{bandyopadhyay2024enhancing}          & \cmark & \pmark & \cmark & PDF       & \xmark \\
PASS~\citep{aggarwal2025pass}                       & \cmark & \pmark & \cmark & PPTX      & \xmark \\
RCPS~\citep{xi2025multi}                            & \cmark & \cmark & \cmark & PPTX/PDF  & \xmark \\
PPTAgent~\citep{zheng2025pptagent}                  & \cmark & \cmark & \cmark & PPTX/HTML & \cmark \\
Auto-Slides~\citep{yang2025auto}                    & \cmark & \cmark & \cmark & PDF       & \xmark \\
AutoPresent~\citep{ge2025autopresent}               & \xmark & \pmark & \cmark & PPTX      & \xmark \\
\textbf{SlideGen (Ours)}                                   & \cmark & \cmark & \cmark & PPTX      & \cmark \\
\bottomrule
\end{tabular}%
}
\vspace{-12pt}
\end{table}

\begin{figure*}[!t]
  \includegraphics[width=0.98\textwidth]{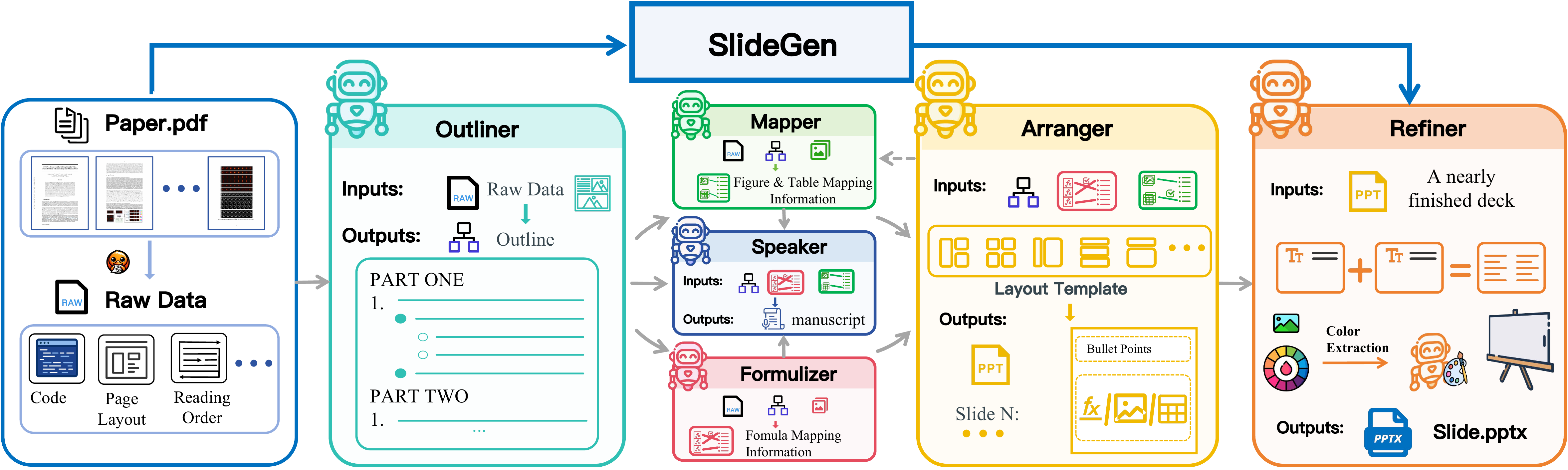}
  \caption{\textbf{Overview of SlideGen.} (1) Outliner generates the presentation narrative. (2) Mapper \& Formulizer assign figures and equations to specific slides. (3) Arranger assembles each slide based on a selected layout template. (4) Finally, Refiner adjusts the visual consistency, and Speaker adds presenter notes. The gray arrows indicate the data flow between agents.}
  \vspace{-5pt}
  \Description{}
  \label{fig:pipeline}
\end{figure*}

Despite their central role, academic slides are still created almost entirely by hand.  Preparing a high-quality presentation from a long scientific paper is time-consuming, stylistically inconsistent, and difficult to scale as research output continues to grow.  Automating this process is thus an important challenge at the intersection of long-context language understanding and layout-sensitive multimodal reasoning~\citep{fu2022doc2ppt,naveed2025comprehensive}.  Recent advances in multimodal large language models~\citep{zhou2023survey,liu2023medical,you2021self,you2021knowledge,you2022end,zhang2025prompt,zhang2024counting,liu2021auto,liu2023benchmarking,liu2025application,wei2025ai} have sparked increasing interest in paper-to-slide generation~\citep{sun2021d2s,fu2022doc2ppt,bandyopadhyay2024enhancing,xi2025multi,ge2025autopresent,xu2025textual,mondal2024presentations,shi2025presentagent,zheng2025pptagent,yang2025auto,zhang2026postergen,sun2025docagent}.  Early systems such as D2S~\citep{sun2021d2s} and Doc2PPT~\citep{fu2022doc2ppt} focused primarily on extracting salient textual content, while more recent approaches incorporate programmatic slide editing~\citep{ge2025autopresent} or iterative refinement pipelines~\citep{zheng2025pptagent,xu2025textual}.

However, a key limitation persists, as most existing methods still frame slide generation mainly as a summarization or content assembly task, rather than a multimodal design problem. 
As a result, generated slides often exhibit repetitive layouts, poor balance between text and figures, and limited narrative coherence across pages~\citep{ge2025autopresent,zheng2025pptagent}. 
In practice, these issues lead to cluttered slides, reduced readability, and visual fatigue (\cref{fig:layout_comp}). \cref{tab:auto-slide-comparison-simplified} summarizes representative systems along key functional dimensions, highlighting that while recent approaches increasingly support multimodal inputs and structured content extraction, few explicitly address visual layout design or produce fully editable presentation outputs.

Motivated by this gap, we propose \textbf{SlideGen}, a modular multi-agent framework that coordinates narrative planning, multimodal grounding, and principled layout composition for slide generation. Rather than producing slides through a single monolithic model or rigid formatting rules, SlideGen orchestrates a team of specialized vision-language agents that jointly reason about document structure and visual organization. \cref{fig:pipeline} illustrates the overall workflow of {SlideGen}.  Given a scientific paper, the system first performs global document parsing and multimodal asset extraction using Docling~\citep{livathinos2025docling} and Marker~\citep{paruchuri2025marker}, following prior document-based visual generation pipelines~\citep{pang2025paper2poster}.  While this step provides the raw ingredients for slide construction, turning them into an effective presentation requires coordinated decisions about storyline, evidence selection, and layout design.

To this end, SlideGen decomposes slide generation into a set of collaborative roles, each handled by a specialized agent. 
\textbf{Outliner} determines the high-level narrative of the presentation, allocating key ideas into a coherent sequence of slides. 
\textbf{Mapper} and \textbf{Formulizer} then align figures, tables, and equations with the claims they support, ensuring that multimodal content is used to explain rather than merely decorate. 
\textbf{Speaker} synthesizes concise presenter notes that connect slides into a smooth oral narrative.
Crucially, SlideGen treats layout composition as a dedicated stage rather than a byproduct of content placement. 
\textbf{Arranger} selects diverse templates and composes slide pages by balancing textual and visual content, avoiding the uniform and repetitive layouts common in prior systems~\citep{ge2025autopresent,zheng2025pptagent}. 
Finally, \textbf{Refiner} adjusts emphasis, merges sparse pages, and resolves layout inconsistencies to improve readability and coherence across the entire deck.

Beyond system design, evaluating slide quality presents its own challenges. 
Early work primarily relied on text-based measures such as ROUGE and perplexity~\citep{lin2004rouge,jelinek1977perplexity,sun2021d2s,fu2022doc2ppt}, while more recent protocols incorporate multimodal QA and VLM-based judges to assess content fidelity, design, and narrative coherence~\citep{pang2025paper2poster,zheng2025pptagent,shi2025presentagent}. 
However, these evaluations only partially capture visual organization and aesthetic balance, including sparsity, clutter, and fragmentation.
To address this gap, we establish a standardized evaluation protocol spanning four complementary dimensions: 
(i) \textit{visual aesthetics}, 
(ii) \textit{communication effectiveness}, 
(iii) \textit{holistic coherence}, and 
(iv) \textit{textual clarity}. 
In particular, we propose \textit{geometry-aware density (GAD)}, a layout-centric metric that quantifies visual clutter, sparsity, and fragmentation, and correlates strongly with human judgments.

Our contributions are as follows.
(i) We introduce \textbf{SlideGen}, a multi-agent framework that handles narrative planning, figure-equation grounding, and template-based layout composition to produce editable PPTX slides without reference decks.
(ii) We define the four-dimensional evaluation protocol of slide generation and propose \textit{geometry-aware density (GAD)}, a novel layout metric that quantifies occupancy and fragmentation balance with high concordance to human judgments.
(iii) We design an extensible layout library with fixed-slot templates to handle diverse slide compositions, which effectively prevents element overlap by construction.
Taken together, our work shows that decomposing slide generation into dedicated agents for narrative planning, multimodal grounding, and layout composition produces slides that are both faithful to the source paper and effective as presentations.

\vspace{-10pt}

%% file: sections/2_related_work.tex
\section{Related Work}

\subsection{Vision-Language Agents for Slides}

Early document-to-slide systems largely framed slide generation as a text generation problem. 
Representative approaches include query-based single-document summarization~\citep{sun2021d2s} and sequence-to-sequence mapping from paper sections to slide content~\citep{fu2022doc2ppt,kothawade2020deep}. 
While effective at identifying salient textual information, these methods pay limited attention to visual structure and multimodal organization, often producing slides that prioritize content selection over presentation quality.

With the emergence of large vision-language models (VLMs)~\citep{you2020towards,you2024calibrating,cao2023multi,yang2023multimodal,you2021mrd,you2020contextualized,chen2021self,liu2021aligning,chen2021adaptive,liu2022retrieve,wei2026think,gao2026neurosonic}, research on automatic slide generation has increasingly moved from single-shot prompting toward multi-stage and agent-based pipelines~\citep{openai2024gpt4technicalreport,naveed2025comprehensive}. 
A common design pattern in this line of work is to decompose slide generation into planning and grounding stages. 
For instance, \textit{DocPres}~\citep{bandyopadhyay2024enhancing} separates global summarization, outline construction, and slide-section grounding, while \textit{RCPS}~\citep{xi2025multi} assigns specialized agents to global planning, layout planning, and iterative refinement. 
Relatedly, \citep{xu2025textual} introduces a Reviewer-Refiner loop that improves layout fidelity through iterative feedback.
More recently, \textit{PaperX}~\citep{yu2026paperx} organizes content through DAG-based semantic planning, and both \textit{PreGenie}~\citep{xu-etal-2025-pregenie} and \textit{SlideBot}~\citep{xie2026slidebot} adopt multi-agent architectures for end-to-end presentation generation.
\textit{SlideTailor}~\citep{zeng2026slidetailor} takes a different angle by conditioning on user-supplied reference slides to personalize the output style, whereas our work targets fully automatic, reference-free generation.

Despite these advances~\citep{zhao2025timeseriesscientist,wen2025beyond,you2025uncovering,xiong2025quantagent,sun2026coma,chen2023bridge,min2025unihgkr,li2025visreason,wang2026let,guo2026csrv2}, most existing agent-based systems still treat visual layout as a secondary concern, often relying on post-hoc adjustments or template-driven formatting. 
As a result, generated slides may achieve reasonable content coverage but still exhibit issues such as repetitive layouts, unbalanced visual density, or weak narrative coherence across pages.

Among recent high-performing baselines, \textit{PPTAgent}~\citep{zheng2025pptagent} adopts a two-stage, edit-based pipeline that iteratively refines HTML slide layouts through self-correction. 
While effective in leveraging agent feedback, this approach relies on explicit reference layouts for editing and is prone to layout artifacts, including element overlap and text overflow. 
A common thread across these systems is that visual layout decisions are made either implicitly through VLM-generated HTML or by imitating a given reference deck, rather than through explicit layout reasoning.
No existing system jointly addresses narrative planning, multimodal asset grounding, and principled layout composition within a single coordinated pipeline.

\subsection{Evaluation Protocols and Metrics for Slides}
Evaluation methods for automatic slide generation have shifted from text-only measures to multimodal and narrative-aware protocols.
Early work primarily relied on n-gram overlap metrics such as ROUGE and language-model fluency measures such as perplexity to assess slide text quality~\citep{sun2021d2s,fu2022doc2ppt,lin2004rouge,jelinek1977perplexity}. 
While useful for evaluating textual summaries, these metrics fail to capture visual organization and presentation effectiveness.

More recent approaches extend evaluation beyond pure text fidelity. 
Several works introduce multimodal, source-grounded question answering, where questions derived from the original paper are used to assess how well generated slides support factual reasoning. 
Others employ VLM-based judges operating on slide renderings to evaluate layout design, readability, and narrative flow~\citep{pang2025paper2poster,zheng2025pptagent,shi2025presentagent}. 
VLM-as-judge protocols further assess content fidelity, design quality, and cross-slide coherence~\citep{bandyopadhyay2024enhancing,zheng2025pptagent,pang2025paper2poster,xi2025multi,shi2025presentagent}.
In the adjacent poster domain, \textit{PosterLLaVa}~\citep{yang2024posterllava} proposes geometric metrics for occlusion and balance, but these target single-page compositions and do not transfer directly to multi-slide decks.
However, these evaluations can be sensitive to prompt formulation and model choice, and they only partially reflect visual layout aesthetics such as sparsity, clutter, and fragmentation.

To address this gap, we propose \textit{geometry-aware density (GAD)}, a layout-centric metric that quantifies page occupancy, overlap, and fragmentation to assess visual organization and aesthetic balance. 
We further validate the GAD metric against human judgments, demonstrating its effectiveness as a complementary measure for evaluating design-aware slide generation systems.

%% file: sections/3_method.tex
\section{SlideGen}
\label{sec:method}

\begin{figure}[!t]
\centering
\includegraphics[width=0.98\linewidth]{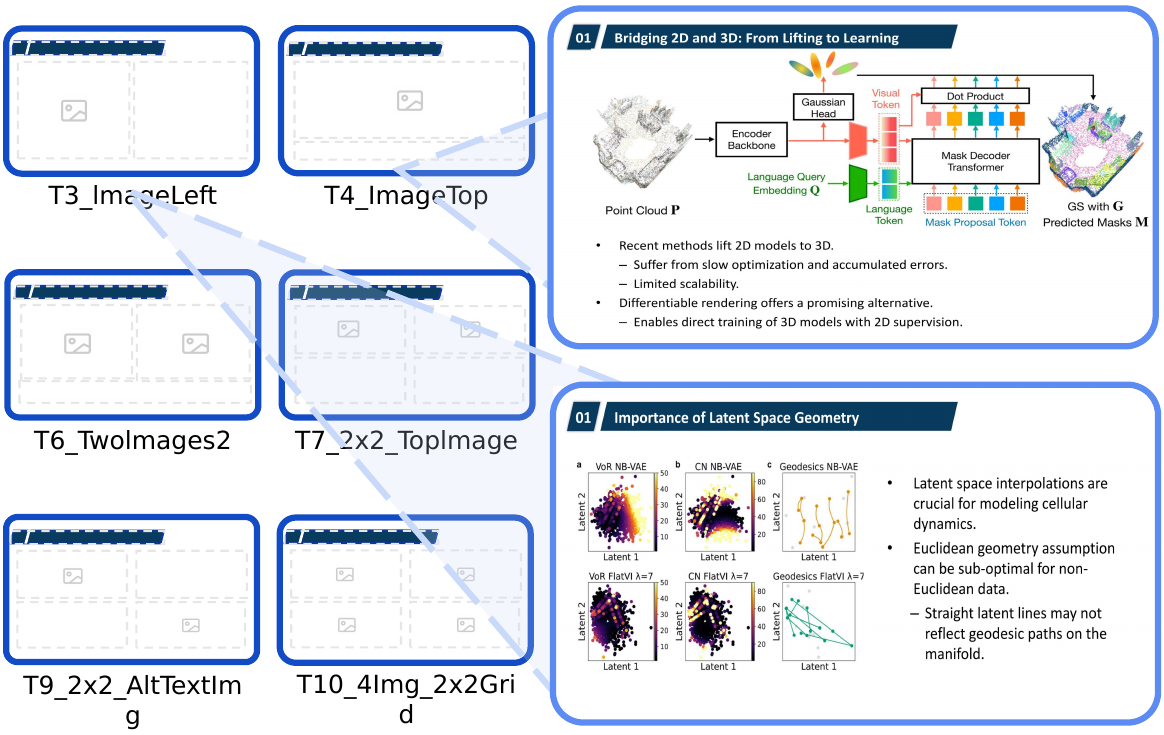}
\vspace{-5pt}
\caption{Representative templates from the library. The templates range from text-based to multi-figure and equation-centric layouts. Arranger selects the template based on the content profile of each slide. \cref{fig:template} in Appendix shows all 19 templates along with the selection rules in \cref{tab:template-rules}.}
\Description{}
\label{fig:template_library}
\vspace{-10pt}
\end{figure}

\myparagraph{Problem Formulation.}
To turn a scientific paper into slides, we need to jointly plan the narrative, assign the figures and equations to the appropriate slides, and decide on the layout of the slide.
These design choices interact in the way that the figures assigned to a slide influence the choice of the layout, and the narrative determines which figures to be combined on a slide.

We denote the paper by $\mathcal{P}$ and the generated slide deck by $\mathcal{S} = \{s_1, \dots, s_N\}$, where a paper provides section text $\mathcal{T}$, figures and tables $\mathcal{V}$, and equations $\mathcal{E}$.
For each slide $s_i$, the agents decide on the title $t_i$ and bullet points $b_i$, the figures $v_i \subseteq \mathcal{V}$ and equations $e_i \subseteq \mathcal{E}$ to be included, and assign the layout $\ell_i$ selected from the predefined template library $\mathcal{L}$ (shown in \cref{fig:template}).

\myparagraph{Overview.}
SlideGen is a multi-agent framework that transforms scientific papers into editable PPTX slide decks.
The entire multi-agent pipeline is shown in \cref{fig:pipeline}.


In detail, SlideGen consists of six agents organized into four stages, where each agent builds on the outputs of its predecessors:
(1) \textbf{Narrative Planning} (Outliner),
(2) \textbf{Multimodal Grounding} (Mapper and Formulizer),
(3) \textbf{Layout Composition} (Arranger), and
(4) \textbf{Global Refinement} (Refiner and Speaker).
The division of stages is based on the main design principle that effective slides emerge from coordinated reasoning over structure, content alignment, and visual organization, rather than isolated content generation.
The stages follow a dependency order, as narrative must be planned before figures can be grounded, and grounding must complete before layouts can be composed.
The pipeline is therefore primarily feed-forward, with one targeted exception for layout feasibility: when Arranger detects that a content assignment cannot be placed in any compatible template, it returns an infeasibility signal to Mapper, which redistributes or splits the affected visual assets across slides before layout composition resumes.

We use separate agents rather than a single end-to-end model for two critical reasons.
First and foremost, every individual agent stage can benefit from a specific prompting scheme. For example, while Outliner needs to reason over the complete context of the paper, Arranger only needs access to the content profile of the slide and the template library.
Second, the division of agents makes the pipeline auditable. A poorly generated layout can be easily traced to Arranger without re-evaluating the entire pipeline.

\myparagraph{Preprocessing.}
Before any agent runs, we parse the input PDF into section text $\mathcal{T}$ and extract figures and tables with their captions into $\mathcal{V}$.
Here equations need special handling, since raw PDF text does not preserve mathematical layout. To this end, we detect equation bounding boxes and crop them as images into $\mathcal{E}$.

\myparagraph{Outliner.}
Outliner agent determines the global narrative of the presentation. 
Given $\mathcal{P}$, it produces an ordered outline:
\begin{equation}
\mathcal{O} = (o_1, \dots, o_M),
\end{equation}
where each $o_j$ corresponds to a coherent presentation segment.
For each segment, Outliner assigns one or more slides and generates concise titles and summaries. 
Outliner may split a long topic across two slides or fold a minor one into its neighbor, depending on how much space each idea needs. The resulting outline therefore varies from paper to paper (see Appendix \cref{sec:all-example} and \cref{fig:Outliner1,fig:Outliner2,fig:Outliner3,fig:Outliner4,fig:Outliner5}).

\myparagraph{Mapper \& Formulizer.}
Given the outline $\mathcal{O}$, Mapper agent assigns visual assets $v \in \mathcal{V}$ to slides based on semantic alignment. 
Each assignment produces a tuple $(v, s_i)$ with a short justification explaining its narrative role (more details in Appendix Section~\ref{sec:all-example} and Figure~\ref{fig:Mapper1}).
Formulizer agent addresses a distinct grounding challenge. Unlike figures, equations lack captions, so their relevance to a slide must be inferred from surrounding text.
We crop each display equation as a high-resolution image to preserve the original typesetting, then match it to the outline based on its textual context.
This avoids the re-rendering artifacts that LaTeX round-tripping often introduces. 
Together, Mapper and Formulizer ensure that figures, tables, and equations are tied to explanatory context.

\myparagraph{Arranger.}
Based on the outline and the multimodal assignments from previous stages, Arranger assembles each slide by matching its content with an appropriate layout template.
For each slide $s_i$, Arranger selects a layout template $\ell_i$ from a predefined library $\mathcal{L}$, and assigns the text blocks and visual content to the fixed slots in the template.
(i) \textit{Template library.}
We use a finite template library instead of unconstrained coordinate generation, since fixed slots prevent overlaps and allow for editability in PowerPoint or WPS.
As shown in \cref{fig:template_library}, the template library $\mathcal{L}$ contains 19 templates covering different page structures, e.g., text-only, text-image pairings, multi-image grids, and formula-centric. These 19 layouts are a demonstration set rather than a hard system limit. 
Users can extend the library through the slide-master interface without the need for modifying any part of the pipeline (shown in Appendix \cref{fig:slide_master}).
(ii) \textit{Template selection.} Arranger selects a template based on the content of the slide, i.e., the number of bullet points ($b_i$) and images ($v_i$), the presence of any equations ($e_i$), and the aspect ratio of the dominant figure. Specifically, it solves:
\begin{equation}
\ell_i = \arg\max_{\ell \in \mathcal{L}} \mathrm{Compat}(\ell, b_i, v_i, e_i),
\end{equation}
where $\mathrm{Compat}$ checks (1) slot availability for the images in $v_i$, (2) aspect-ratio compatibility between the dominant figure and the main image slot (landscape for wide figures, portrait for tall ones), and (3) sufficient text space for the bullet points ($b_i$). If every candidate template violates these constraints, Arranger reports the slide as infeasible and sends the overloaded asset set back to Mapper. Mapper then either redistributes low-priority visuals to adjacent slides in the same outline segment or splits the segment into an additional slide, which prevents the system from forcing incompatible content into a single page.  The complete rule table for all 19 templates can be found in Appendix \cref{tab:template-rules}.

In practice, a slide containing a single wide and dominant figure with a corresponding short caption is often allocated to an image-top layout type (e.g., \texttt{T4}). If the dominant figure is tall or near square in shape, Arranger suggests a balanced split layout type, e.g., image-right (\texttt{T2}) or image-left (\texttt{T3}) to avoid sacrificing readability and wasting horizontal space.
By decoupling content planning from layout selection, Arranger produces a near-complete deck with varied designs. A full example is provided in Appendix \cref{sec:arranger-example}.

\myparagraph{Refiner.}
After the Arranger stage, Refiner makes t hree deck-level adjustments.
\textit{(i) Slide merging.} Consecutive text-only slides in the same section tend to look sparse on their own. Refiner merges each such pair into a two-column layout (T19), with one slide forming the left column and the other the right. \textit{(ii) Overflow handling.} For body text and bullets, Refiner first applies bounded font scaling down to readable thresholds (body text at least 16pt and sub-bullets at least 14pt). If text still exceeds its placeholder, it asks the language model to compress the content while preserving the main claim and referenced evidence. For slide titles, whose font size should remain visually consistent, Refiner applies compression directly. Visual overflow is handled earlier by Arranger's compatibility check and Mapper redistribution fallback.
\textit{(iii) Color theming.} To make the deck visually cohesive, Refiner extracts the most frequent color from the paper's figures after filtering out near-white and near-black pixels. It then darkens and saturates this color so it works as a slide accent. Details are provided in Appendix~\cref{sec:color-refine}.

\myparagraph{Speaker.}
Although bullet points summarize content and convey meanings visually, a presenter also needs notes that explain transitions and put visuals in context. To this end, Speaker produces one paragraph of notes per subsection, typically 2 to 5 sentences long. Each paragraph leverages the outline generated by Outliner and expands on the essential ideas in the subsections without referring to the bullets. 
Furthermore, the notes include the rationales for Mapper's figures and Formulizer's equation descriptions, helping the presenter understand why each visual appears on its slide. This produces a narrative layer parallel to the visual deck. A new presenter can read the notes and deliver the talk without going back to the original paper.
An example is shown in Appendix \cref{fig:speaker1}.

%% file: sections/4_experiments.tex
\section{Experiment}
  
\subsection{Dataset}
\setlength{\columnsep}{14pt}

\myparagraph{Data Source.} We curated a domain-specific dataset focused on recent advances in machine learning and natural language processing, with a particular emphasis on research diversity and quality. Our dataset consists of 200 peer-reviewed papers collected from leading AI venues between 2022 and 2025, including \textit{Oral} presentations only as designated by each conference, as shown in \cref{tab:pdfcountsmatrix}. 

\begin{table*}[!t]
  \centering
  \caption{Comparison of all methods on SlideQA (scored by three open-source and three closed-source VLMs), Perplexity, layout Density, and VLM-as-Judge. Methods are grouped by backbone; \textbf{bold} marks the best within each group}
  \label{tab:unified-results}
  \setlength{\tabcolsep}{4pt}%
  \renewcommand{\arraystretch}{1.08}%
  \resizebox{\textwidth}{!}{%
  \begin{tabular}{@{}l ccc ccc c !{\vrule width 0.4pt} c !{\vrule width 0.4pt} ccc !{\vrule width 0.4pt} cccc@{}}
    \toprule
    \multirow{2}{*}[-0.5ex]{\textbf{Model}} &
      \multicolumn{3}{c}{\textbf{Verbatim} $\uparrow$} &
      \multicolumn{3}{c}{\textbf{Interpretive} $\uparrow$} &
      \multirow{2}{*}[-0.5ex]{\shortstack{\textbf{SlideQA}\\\textbf{Avg.}}} &
      \multirow{2}{*}[-0.5ex]{\textbf{PPL}$\downarrow$} &
      \multicolumn{3}{c!{\vrule width 0.4pt}}{\textbf{Density} $\uparrow$} &
      \multicolumn{4}{c}{\textbf{VLM-as-Judge} $\uparrow$} \\
    \cmidrule(lr){2-4} \cmidrule(lr){5-7} \cmidrule(lr){10-12} \cmidrule(lr){13-16}
     & open & closed & \textbf{V-Avg.} & open & closed & \textbf{I-Avg.} &
     &  & OM & FR & \textbf{D-Avg.} &
     \textbf{Content} & \textbf{Design} & \textbf{Coherence} & \textbf{Avg.} \\
    \midrule

    \multicolumn{16}{l}{\textit{GPT-5}}\\
    \rowcolor{GroupBlue!50}
    HTML-5 &
      74.95 & 70.17 & 72.56 &
      84.24 & 87.88 & 86.06 &
      79.31 &
      189.38 &
      54.32 & 60.66 & 56.86 &
      3.54 & 4.02 & 4.09 & 3.88 \\
    \rowcolor{GroupBlue!50}
    Image-5 &
      66.99 & 53.90 & 60.45 &
      73.06 & 80.51 & 76.79 &
      68.62 &
      605.02 &
      67.96 & 79.29 & 72.49 &
      2.84 & 3.16 & 3.21 & 3.07 \\
    \rowcolor{GroupBlue!50}
    PPTAgent-5 &
      61.37 & 63.21 & 62.29 &
      75.22 & 78.55 & 76.89 &
      69.59 &
      450.12 &
      55.32 & 62.40 & 58.15 &
      3.10 & 3.25 & 3.40 & 3.25 \\
    \rowcolor{GroupBlue!50}
    PosterAgent-5 &
      70.08 & \textbf{78.98} & \textbf{74.53} &
      82.21 & 85.34 & 83.78 &
      79.15 &
      220.75 &
      62.42 & 70.28 & 65.56 &
      3.45 & 3.60 & 3.75 & 3.60 \\
    \rowcolor{GroupBlue}
    \textbf{Ours-5} &
      \textbf{75.70} & 70.05 & 72.88 &
      \textbf{84.65} & \textbf{90.36} & \textbf{87.51} &
      \textbf{80.19} &
      \textbf{48.40} &
      \textbf{69.08} & \textbf{84.56} & \textbf{75.27} &
      \textbf{4.12} & \textbf{4.30} & \textbf{4.35} & \textbf{4.26} \\

    \\[-1ex]
    \multicolumn{16}{l}{\textit{GPT-4o}}\\
    \rowcolor{GroupGreen!50}
    HTML-4o &
      60.48 & \textbf{75.58} & 68.03 &
      87.38 & 91.37 & 89.38 &
      78.70 &
      200.79 &
      41.15 & 46.38 & 43.24 &
      3.02 & 2.76 & 3.97 & 3.25 \\
    \rowcolor{GroupGreen!50}
    Image-4o &
      48.97 & 30.85 & 39.91 &
      50.11 & 70.72 & 60.42 &
      50.16 &
      793.71 &
      75.28 & 76.24 & 75.66 &
      2.39 & 3.09 & 3.50 & 2.99 \\
    \rowcolor{GroupGreen!50}
    PPTAgent-4o &
      57.99 & 52.44 & 55.22 &
      57.51 & 56.34 & 56.93 &
      56.07 &
      721.54 &
      53.27 & 56.31 & 54.49 &
      3.25 & 3.24 & 3.29 & 3.26 \\
    \rowcolor{GroupGreen!50}
    PosterAgent-4o &
      67.75 & 67.86 & 67.81 &
      72.99 & 79.89 & 76.44 &
      72.12 &
      139.67 &
      68.76 & 76.25 & 71.76 &
      3.19 & 3.48 & 4.53 & 3.73 \\
    \rowcolor{GroupGreen}
    \textbf{Ours-4o} &
      \textbf{75.89} & 71.23 & \textbf{73.56} &
      \textbf{90.60} & \textbf{93.89} & \textbf{92.25} &
      \textbf{83.90} &
      \textbf{50.59} &
      \textbf{79.66} & \textbf{82.32} & \textbf{80.99} &
      \textbf{4.01} & \textbf{4.28} & \textbf{4.66} & \textbf{4.32} \\

    \\[-1ex]
    \multicolumn{16}{l}{\textit{Qwen}}\\
    \rowcolor{pink!20}
    PosterAgent-7B &
      49.74 & 47.43 & 48.59 &
      54.30 & 56.25 & 55.28 &
      51.93 &
      450.30 &
      42.48 & 49.11 & 45.13 &
      2.41 & 2.69 & 2.84 & 2.65 \\
    \rowcolor{pink!20}
    SlideGen-7B &
      55.52 & 53.10 & 54.31 &
      60.83 & 63.16 & 62.00 &
      58.15 &
      180.50 &
      46.61 & 54.02 & 49.57 &
      2.70 & 2.95 & 3.12 & 2.92 \\
    \rowcolor{pink!20}
    PosterAgent-72B &
      58.97 & 56.32 & 57.65 &
      65.29 & 68.88 & 67.09 &
      62.37 &
      150.60 &
      49.35 & 57.42 & 52.58 &
      2.92 & 3.13 & 3.30 & 3.12 \\
    \rowcolor{pink!30}
    SlideGen-72B &
      \textbf{62.74} & \textbf{60.59} & \textbf{61.67} &
      \textbf{72.14} & \textbf{74.51} & \textbf{73.33} &
      \textbf{67.50} &
      \textbf{80.90} &
      \textbf{52.31} & \textbf{60.56} & \textbf{55.61} &
      \textbf{3.10} & \textbf{3.32} & \textbf{3.54} & \textbf{3.32} \\

    \bottomrule
  \end{tabular}
  }
\end{table*}

\subsection{Evaluation Metrics}
\label{sec:metrics}

We evaluate this Paper-to-Slide task with four complementary metrics: (i) layout quality (GAD), (ii) deck-only answerability (SlideQA), (iii) presentation quality across \textcolor{BrickRed}{Content}, \textcolor{ForestGreen}{Design} and \textcolor{Goldenrod}{Coherence} (VLM-as-Judge), and (iv) textual coherence (writing and flow).

\myparagraph{Layout Balance.}\label{sec:geom-density}
We propose the \textit{Geometry-Aware Density (GAD)} score to quantify how well a slide balances occupied and empty space. GAD combines two per-slide terms.

\textit{(i) Occupancy.} The Area Occupancy penalizes deviations from a target density $\tau$, which can be expressed by:
\begin{equation}
\mathrm{OM}_i = 1-\big|\rho_i-\tau\big|,
\end{equation}
where $\rho_i$ is the fraction of slide $i$'s canvas covered by content regions, which are computed from PPTX bounding boxes.

\textit{(ii) Fragmentation reward} peaks when the number of content elements $M_i^{\mathrm{eff}}$ (excluding items smaller than a gate $a_{\min}$) equals an optimal count $M^\star$:
\begin{equation}
\mathrm{FR}_i = \max\!\Big\{\,0,\ 1-\frac{(M_i^{\mathrm{eff}}-M^\star)^2}{\kappa}\,\Big\}.
\end{equation}

The overall deck-level GAD score then averages both terms across all $N$ slides, under the constraint of $\lambda_1+\lambda_2=1$:
\begin{equation}
\mathrm{GAD} = \frac{1}{N}\sum_{i=1}^{N}\big(\lambda_1\,\mathrm{OM}_i + \lambda_2\,\mathrm{FR}_i\big).
\end{equation}
We calibrate $(\lambda_1, \lambda_2, M^\star, \kappa)$ against human ratings via leave-one-deck-out cross-validation (Appendix~\cref{alg:lodo}). The optimum lies at $\lambda_1 \approx 0.6$ and remains stable across $[0.5, 0.7]$. The full derivation and sensitivity analysis are provided in Appendix~\cref{sec:Geometry-Aware}.

 \begin{figure*}
 \begin{subfigure}{\linewidth}
  \centering
  \includegraphics[width=\linewidth]{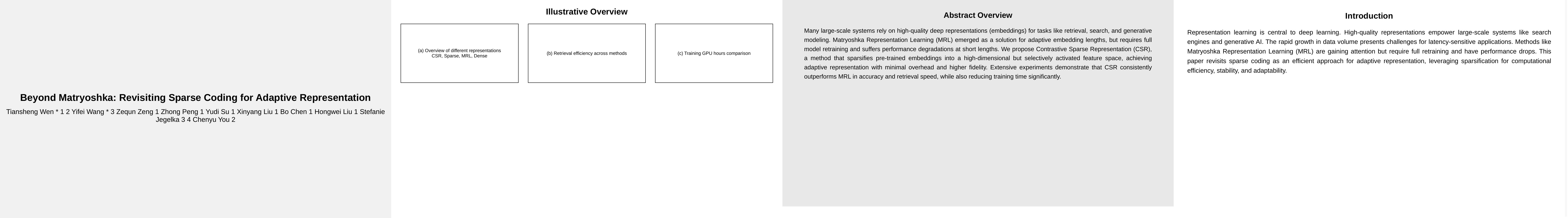}
  \vspace{-10pt}
  \caption{Example slides generated via \textbf{GPT-Image baseline (GPT-5)}. } \label{fig:gpt5_a3}
  \end{subfigure}
 \begin{subfigure}{\linewidth}
  \centering
  \includegraphics[width=\linewidth,page=1]{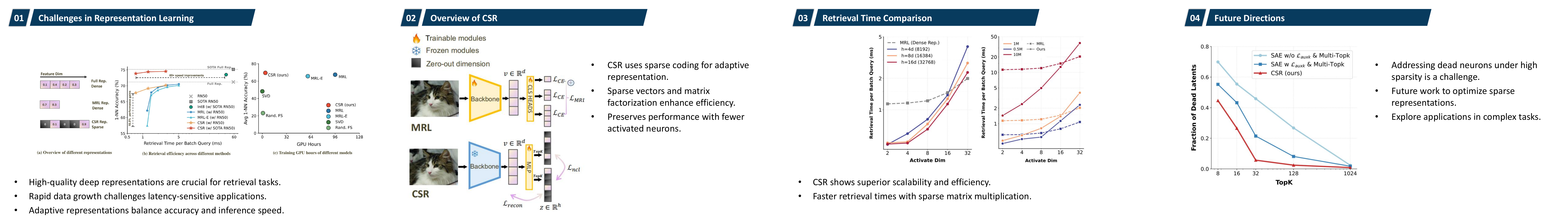}
  \vspace{-10pt}
  \caption{Example slides generated via \textbf{SlideGen (GPT-4o)}.} \label{fig:gpt5_a1}
  \end{subfigure}
  \begin{subfigure}{\linewidth}
  \centering
  \includegraphics[width=\linewidth,page=2]{figures/gpt-5_gpt-5_output_slide.pdf}
  \vspace{-10pt}
  \caption{Example slides generated via   \textbf{SlideGen (GPT-5)}. } 
  \vspace{-5pt}
  \label{fig:gpt5_a2}
  \end{subfigure}  
\vspace{-15pt}
\caption{\textbf{Comparison of generated slides from the same source paper.}
(a) Generated by GPT-Image (GPT-5). (b) Generated by SlideGen (GPT-4o). (c) Generated by SlideGen (GPT-5). Additional samples are shown in Appendix \cref{fig:example11,fig:example12,fig:example13,fig:example14,fig:example15}.}
\Description{}
\label{fig:gpt5_example}
\vspace{-15pt}
\end{figure*}

\myparagraph{Holistic Assessment.} Following PPTEVAL \citep{zheng2025pptagent}, we evaluate decks along three dimensions: \textcolor{BrickRed}{Content}, \textcolor{ForestGreen}{Design} and  \textcolor{Goldenrod}{Coherence}, using  GPT-4o as the judge. Scores range from 1 to 5 and are accompanied by brief rationales. The criteria are listed in \cref{tab:ppteval-criteria}.

\myparagraph{Communication
Effectiveness.} Since slide decks are the primary vehicle by which speakers convey knowledge and audiences learn it, we need to evaluate whether our generated presentations communicate the material, and how much they succeed in doing so. Following PaperQuiz ~\citep{pang2025paper2poster}, for each paper, we first generate a quiz of 100 questions from the paper PDF: 50 verbatim questions answerable directly from the text, covering diverse factual aspects, and 50 interpretive questions targeting higher-level comprehension.
Then the questions are  answered by six different VLM readers.

 \myparagraph{Textual Coherence.} 
We quantify textual coherence using the standard ``Perplexity'' (PPL) metric, calculated for the entire slide text under Llama-2-7b-hf. A lower PPL score indicates more predictable and coherent language, see details  in Appendix \cref{sec:metrics-ppl}.

\subsection{Baselines and Settings}
\label{sec:baselines&settings}
For a fair comparison, all methods take the same raw PDF as the only input and produce a multi-slide deck on a 16:9 canvas; the number of slides is unconstrained. The baselines we use fall into two groups. \textit{(i) End-to-end generators}: GPT-5 HTML and GPT-4o HTML produce slides as HTML+CSS, from which text is read directly. GPT-5 Image and GPT-4o Image synthesize slide images page by page; for these, we extract text with \textbf{EasyOCR} before computing PPL and SlideQA. \textit{(ii) Multi-agent workflows}: PPTAgent-4o, PPTAgent-5, PosterAgent-4o, PosterAgent-5, and PosterAgent-Qwen-2.5-VL 7B\&72B (used in \textit{slide mode}) decompose planning, drafting, and layout into iterative editing steps. Here, we exclude RCPS~\citep{xi2025multi} (no public code) and AutoPresent~\citep{ge2025autopresent} (requires extra user instructions).
We instantiate SlideGen with two backbones, GPT-4o and GPT-5, for a controlled comparison. VLM-as-Judge reads rendered images directly and does not depend on text extraction.

\begin{figure*}[!t]
\centering
\includegraphics[width=0.98\linewidth]{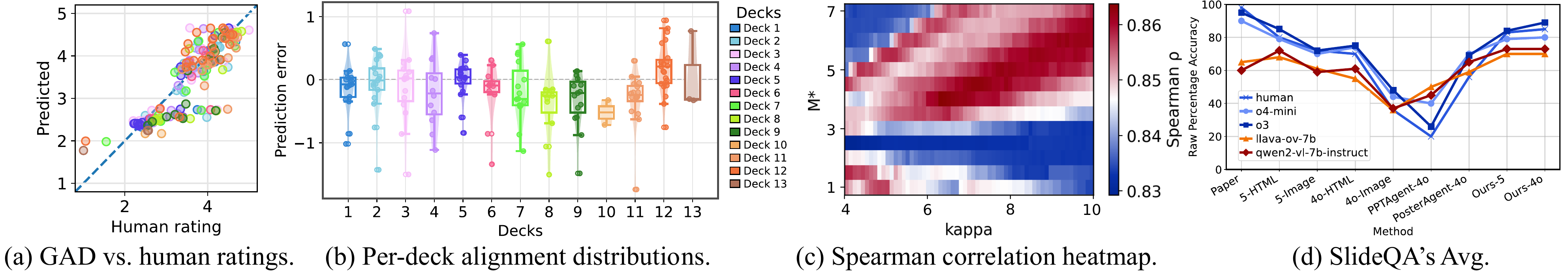}
\vspace{-5pt}
\caption{\textbf{Overview of quantitative results (Leave-One-Deck-Out and human evaluation).}
(a) \textbf{Prediction vs. human ratings.} Each point is one page. The dashed line denotes \(y = x\), which indicates perfect agreement. Metrics are computed across all test pages from 13 decks, about 750 pages in total. Here: RMSE = 0.580, Spearman's \(\rho\) = 0.820, Pearson's \(r\) = 0.811.
(b) \textbf{Per-deck alignment distributions}: page-wise prediction
errors \(\hat{y}-y\) shown as box plots.
(c) \textbf{Spearman correlation heatmap} over the parameter space \((M^*, \kappa)\) on the training folds. 
(d) \textbf{Average SlideQA scores} across different types of slides (x-axis) for readers (colored lines) on human evaluation subset.}
  \Description{}
\label{fig:all_in_one_row}
\vspace{-5pt}
\end{figure*}

\subsection{Main Results}

\myparagraph{Overall Performance.}
As shown in \cref{tab:unified-results}, SlideGen is the best overall among the compared methods.
With GPT-4o, it achieves the best SlideQA average score (83.9), the best VLM-as-Judge score (4.32), and the lowest perplexity score (50.6).
GPT-5 as base model preserves the same ranking, and extends to open-source backbones.
Here, SlideGen-72B outperforms PosterAgent-72B in all columns, and the 7B variant still leads its counterpart by 6.2 SlideQA scores while having less than half the perplexity.
Among all four backbone groups, no other method than SlideGen ranks first on all three metrics simultaneously.
PosterAgent is competitive in the individual columns, and GPT-Image excels in Density metric. 
Still, neither reaches the best scores on question answering, layout quality, and text fluency simultaneously. 
The superiority of SlideGen is due to its balanced optimization over all three criteria instead of just excelling on any single one.

\myparagraph{Metric Patterns.}
Furthermore, SlideGen also achieves the best Density scores among all methods across both backbone groups, with D-Avg 81.0 for GPT-4o and 75.3 for GPT-5. 
This result follows from the design of Arranger design that places content into fixed template slots, and therefore avoids any element overlap by construction (Section~\ref{sec:method}).
Meanwhile, the image-only baselines illustrate a notable phenomenon. 
Their pages can appear well spaced and therefore achieve high scores on Density, but they perform much weaker on SlideQA and VLM-as-Judge metrics.
In contrast to these baselines, SlideGen achieves the best Density scores due to Arranger's template-based placement while preserving content quality, as illustrated through its SlideQA and Judge scores. \cref{fig:gpt5_example} shows a side-by-side comparison of slides generated by GPT-Image and SlideGen from the same source paper.

Across all methods, the interpretive SlideQA performance is substantially above the verbatim SlideQA.
This pattern is expected, as slides distill the original text into bullets, figures, and equations, making it harder to preserve the original words than the original ideas.
More importantly, the performance gap varies from 1.7 for PPTAgent-4o to 21.4 for HTML-4o.
Since the same six VLM readers evaluate all the methods, this indicates the level to which each method preserves the original details in the source paper.
SlideGen maintains its performance gap while achieving high scores on both sub-metrics, which indicates that it condenses content selectively without discarding key details.

\begin{figure*}[!t]
\centering
\includegraphics[width=0.98\linewidth]{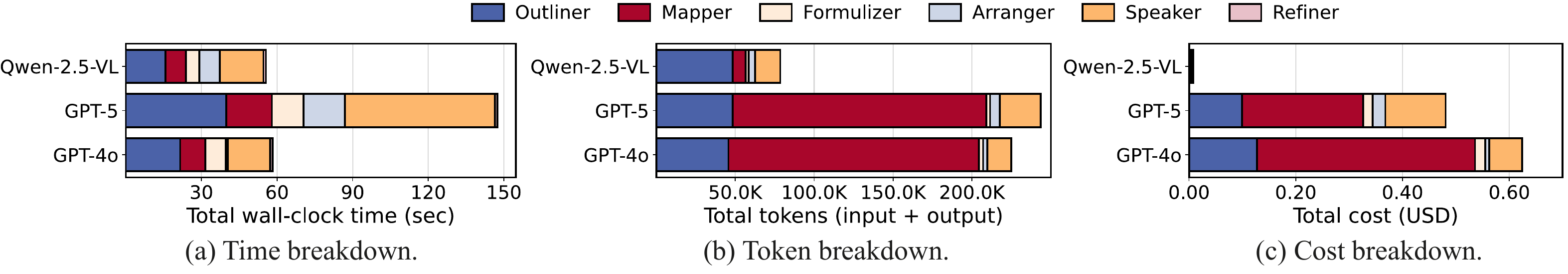}
\vspace{-8pt}
\caption{\textbf{Agent-level efficiency and cost.} Horizontal stacked bars per variant show per-agent contributions to (a) total wall-clock time in seconds, (b) total tokens (input and output), and (c) total cost in USD.}
  \Description{}
\label{fig:agent-runtime-cost}
\vspace{-10pt}
\end{figure*}

\subsection{Metric Validation}
\label{sec:gad-human}

\myparagraph{Layout Quality.}
Automatic GAD metrics show high alignment with human judgement.
For layout quality, we collected ratings from 40 raters on 13 decks using leave-one-deck-out calibration.
Refer to Section~\ref{sec:geom-density} for the full details.
The optimal hyperparameters are $M^\star=4$ and $\kappa=6.3$, which were selected at the peak Spearman correlation (\cref{fig:all_in_one_row}(c)).
The calibrated metric achieves Spearman correlation $\rho=0.820$ and Pearson correlation $r=0.811$ with human judgments (\cref{fig:all_in_one_row}(a)).
Since each test deck is held out during the fitting process, this agreement shows the exact out-of-sample generalization capability of SlideGen.
These results support the use of GAD as a layout-oriented metric to complement content-based metrics, and also further clarify the Density advantage in Table~\ref{tab:unified-results}.
Further details are provided in Appendix \cref{sec:Geometry-Aware}.

\myparagraph{Reading-Based Stability.}
\begin{table}[!t]
  \centering
  \caption{\textbf{Cross-judge robustness on the GPT-4o backbone group.}
  Both judges use the same prompts and criteria.
  Spearman $\rho$ is computed on method-level average scores.}
  \label{tab:cross-judge}
  \vspace{-5pt}
  \resizebox{\columnwidth}{!}{%
  \begin{tabular}{@{}l cccc cccc@{}}
    \toprule
    \multirow{2}{*}{\textbf{Method}} &
      \multicolumn{4}{c}{\textbf{GPT-4o Judge}} &
      \multicolumn{4}{c}{\textbf{Gemini-2.5-Flash Judge}} \\
    \cmidrule(lr){2-5} \cmidrule(lr){6-9}
     & \textbf{Cont.} & \textbf{Des.} & \textbf{Coh.} & \textbf{Avg}
     & \textbf{Cont.} & \textbf{Des.} & \textbf{Coh.} & \textbf{Avg} \\
    \midrule
    HTML-4o        & 3.02 & 2.76 & 3.97 & 3.25  & 2.71 & 2.42 & 3.88 & 3.00 \\
    Image-4o       & 2.39 & 3.09 & 3.50 & 2.99  & 2.05 & 2.74 & 3.08 & 2.62 \\
    PPTAgent-4o    & 3.25 & 3.24 & 3.29 & 3.26  & 2.88 & 2.95 & 3.01 & 2.95 \\
    PosterAgent-4o & 3.19 & 3.48 & 4.53 & 3.73  & 2.96 & 3.11 & 4.41 & 3.49 \\
    \textbf{Ours-4o} & \textbf{4.01} & \textbf{4.28} & \textbf{4.66} & \textbf{4.32}
                      & \textbf{3.72} & \textbf{4.05} & \textbf{4.73} & \textbf{4.17} \\
    \midrule
    \multicolumn{9}{c}{Spearman $\rho$ (method-level Avg) $= 0.90$} \\
    \bottomrule
  \end{tabular}}
  \vspace{-10pt}
\end{table}
We verified that the reported rankings are stable across both human raters and VLM-as-Judges.
For SlideQA, we recruited 5 PhD students to answer questions on 5 randomly selected papers, each with 8 deck variants.
Method rankings are largely preserved across the human panel and six VLM readers (\cref{fig:all_in_one_row}(d)).
For VLM-as-Judge, we tested robustness to judge selection.
Since GPT-4o serves as both the generation backbone and the default judge for this group, we test on an independent evaluator to mitigate self-evaluation bias.
We therefore re-scored all five methods in this group with Gemini-2.5-Flash using the same prompts and rubrics.
As shown in \cref{tab:cross-judge}, the two judges achieve Spearman $\rho{=}0.90$ at the method level, and SlideGen remains the top-ranked method under both judges.
Gemini assigns systematically lower absolute scores (mean offset $\approx 0.26$), yet the margin between SlideGen and the runner-up is larger under Gemini ($0.68$) than under GPT-4o ($0.59$), which suggests that GPT-4o does not favor its own outputs in this evaluation.
Altogether, these evaluations confirm that the reported rankings reflect genuine method differences rather than an artifact of any single evaluator. 

\subsection{Ablation Study}
\begin{table}[!t]
\caption{\textbf{Leave-one-agent-out ablation.}
Each variant removes one agent and replaces it with a simple rule-based fallback.
All experiments use the GPT-4o backbone on the same evaluation subset.}
\label{tab:ablation_compact}
\centering
\scriptsize
\vspace{-5pt}

\resizebox{0.98\linewidth}{!}{%
\begin{tabular}{@{}lccccc@{}}
\toprule
\textbf{Variant} &
\textbf{V-Avg} $\uparrow$ &
\textbf{I-Avg} $\uparrow$ &
\textbf{PPL} $\downarrow$ &
\textbf{D-Avg} $\uparrow$ &
\textbf{Judge} $\uparrow$ \\
\midrule
Full (SlideGen)      & \textbf{73.56} & \textbf{92.25} & \textbf{50.59} & \textbf{80.99} & \textbf{4.32} \\
\midrule
w/o Outliner         & 62.41 & 78.33 & 118.72 & 79.85 & 3.48 \\
w/o Mapper           & 68.12 & 83.90 & 63.47  & 64.73 & 3.67 \\
w/o Formulizer       & 71.03 & 89.41 & 57.82  & 77.26 & 4.12 \\
w/o Arranger         & 72.18 & 90.07 & 54.31  & 37.64 & 3.52 \\
w/o Refiner          & 72.84 & 91.12 & 61.45  & 76.80 & 4.16 \\
w/o Speaker          & 72.30 & 90.58 & 52.13  & 80.45 & 4.22 \\
\midrule
w/o Layout Library & 72.65 & 90.83 & 53.07  & 67.52 & 3.98 \\
\bottomrule
\end{tabular}}
\vspace{-10pt}
  
\end{table}

The ablation study in \cref{tab:ablation_compact} shows that SlideGen benefits from complementary stages rather than one dominant prompt.
Removing Outliner causes the largest drop in QA performance, which highlights the importance of global narrative planning.
Removing Arranger causes by far the largest drop in Density, which confirms that explicit layout composition drives spatial balance.
Removing Mapper lowers interpretive accuracy by 8.4 points, which links figure-to-slide grounding to reader understanding.
The ablations are therefore consistent with the pipeline design that Outliner mainly affects content organization, Arranger mainly affects layout quality, and Mapper mainly affects multimodal grounding.
This separation confirms that each agent addresses a distinct aspect of slide quality.
Detailed ablation setups are provided in Appendix \cref{sec:ablation_details}.

\subsection{Efficiency and Practical Cost}
\label{sec:eff_cost}
We measure wall-clock time, token count, and USD cost per agent for all 200 papers in the benchmark with the same prompts, templates, and decoding configurations on the same machine.
The overall cost is computed as:
\begin{equation}
  \text{Cost} = \tfrac{\text{input\_tokens}}{1000}\,C_{\text{in}} + \tfrac{\text{output\_tokens}}{1000}\,C_{\text{out}}
\end{equation}
\cref{fig:agent-runtime-cost} shows the distribution of these three metrics by agent.
Outliner and Arranger together account for the majority of the token cost, which is reasonable given that Outliner needs to read the full paper and Arranger needs to assess the layout for every slide.
These two agents also show the largest performance degradation when removed in the ablation study (as shown in \cref{tab:ablation_compact}), and therefore their costs align well with the contribution.

\subsection{User Study}
\label{sec:user-study-main}

To further measure how well SlideGen's output holds up from a human perspective, we conducted a user study with 30 participants who have prior experience in crafting and delivering academic presentations.
The participants included 4 undergraduates, 14 master's students, and 12 PhD students across five disciplines: Computer Science, Electrical Engineering, Data Science/Statistics, English, and Cognitive Science.
Each participant was given 10 full PPTX slide decks produced by SlideGen and competitive baselines, with method labels removed and slide deck orders shuffled.

We designed the evaluation around two complementary questions, as shown in \cref{tab:user-study}.
Question 1 focuses on the usability and preference of SlideGen's output compared to the competitive baseline.
Question 2 evaluates SlideGen's quality compared to human performance, where participants were asked to rate SlideGen on a four-point scale, from ``worse than most humans'' to ``top-tier expert quality'', which is also shown in \cref{tab:user-study}.
All the slide decks were provided in the format of PPTX, so participants could review layout, text, and speaker notes carefully before rating.

\begin{table}[!t]
\centering
\caption{\textbf{User study results.} Responses from 30 participants on two questions about SlideGen's generated slide decks.}
\label{tab:user-study}
\vspace{-5pt}
\resizebox{\columnwidth}{!}{%
\begin{tabular}{@{}llc@{}}
\toprule
\textbf{Question} & \textbf{Response Option} & \textbf{\%} \\
\midrule
\multirow{2}{*}{\begin{tabular}[l]{@{}l@{}}\textbf{Q1:} Which slide deck would\\ you choose as your presentation draft?\end{tabular}}
& \textbf{SlideGen} & \textbf{93.33} \\
& Other Baselines & 6.67 \\
\midrule
\multirow{4}{*}{\begin{tabular}[l]{@{}l@{}}\textbf{Q2:} How does SlideGen's output\\ compare to human-made slides?\end{tabular}}
& A. Worse than most humans & 0.00 \\
& B. About as good as average & 20.00 \\
& \textbf{C. Better than most humans} & \textbf{73.33} \\
& \textbf{D. Top-tier, expert quality} & \textbf{6.67} \\
\bottomrule
\end{tabular}
}
\vspace{-10pt}
\end{table}
                                                         
For Question 1, 28 out of 30 participants preferred the slide decks generated by SlideGen, a proportion far above chance (binomial test, $p < 0.001$). 
For Question 2, 73.3\% of participants rated the quality of the slide generated by SlideGen as better than most human-made slides, and 6.7\% of participants rated it at the highest expert level. 
The remaining 20\% rated it as comparable to average human work, and no one rated it worse. 
The results were consistent for different education levels and disciplines, as shown in \cref{fig:preference_by_education_major}, and the strongest endorsement were from participants in Data Science and Computer Science.
We attribute this to SlideGen's structured layout and faithful content extraction, which together produce slides that researchers can adopt as working drafts.

%% file: sections/5_conclusion.tex
\section{Conclusion and Limitations}
In this work, we propose SlideGen, a multi-agent framework that decomposes the process of paper-to-slide into four sub-tasks: narrative planning, multimodal grounding, layout composition, and global refinement. We designed six specialized vision-language agents that contribute complementary capabilities through a staged pipeline to produce editable PPTX files directly from the raw PDF data, without the need for reference slides or manual intervention. Moreover, we introduce \textbf{Geometry-Aware Density}, a layout metric that obtains Spearman correlation coefficient $\rho$ of 0.820 compared to human judgments using leave-one-deck-out calibration. On our curated 200-paper benchmark, SlideGen obtains the best performance on SlideQA, VLM-as-Judge, and perplexity across all four backbones, and 93.3\% of the participants in the user study prefer the slides generated by SlideGen to other baselines.

Despite these results, SlideGen also has several limitations. First, the quality of slide generation relies on the underlying large language model. For example, the Qwen-7B variant achieves noticeably lower performance on all metrics compared to GPT-4o and GPT-5 variants. This indicates that a certain proportion of the effectiveness depends on a strong backbone. Second, the 19 fixed templates are comprehensive for general slide structures, but cannot cover every possible arrangement, and adding domain-specific layouts still requires manual design. Future work will focus on how to enable human-in-the-loop for improving generation effectiveness, where users can iteratively adjust agent outputs, and on learned layout selection method to replace the fixed template library with a learned model that can generate diversified slide layouts.

%% file: sections/appendix.tex
\section{Data and Evaluation}\label{sec:data_eval}

\subsection{Dataset}\label{sec:datasetapp}

We include 200 peer-reviewed papers from leading AI venues between 2022 and 2025. Table~\ref{tab:pdfcountsmatrix} reports counts by venue and year. 
The selected conferences were chosen for their rigorous review process, topical breadth, including multimodal learning, generative modeling, interpretability, and inclusion of rich visual and mathematical content, making them ideal for downstream tasks such as slide generation, summarization, and modality-aware learning.

\begin{table}[!t]
  \caption{Dataset composition by venue and year (orals only).}
  \label{tab:pdfcountsmatrix}
  \centering
  
\resizebox{0.75\linewidth}{!}{
  \begin{tabular}{@{}lcccc c@{}}
    \toprule
    \textbf{Venue} & \textbf{2022} & \textbf{2023} & \textbf{2024} & \textbf{2025} & \textbf{Total} \\
    \midrule
    ICLR    & 17 & 31 & 29 & 23 & 100 \\
    ICML    & -- & 16 & 24 & 30 & 70 \\
    NeurIPS & -- & 10 & 20 & -- & 30 \\
    \midrule
    Total   & 17 & 57 & 73 & 53 & 200 \\
    \bottomrule
  \end{tabular}
  }
\end{table}

\begin{table*}[t]
\caption{\textbf{Comprehensive comparison of automatic slide generation systems.}
The table summarizes key capabilities of existing approaches.
\textbf{SlideGen} is the first unified framework that fulfills all major functional criteria, integrating complete content planning, layout reasoning, and visual refinement.
\textbf{Columns:}
\textit{Content Struct.} -- whether the system constructs a slide-level outline;
\textit{Text--Fig Align.} -- alignment of figures and tables with corresponding text;
\textit{Multimodal} -- support for inputs beyond text;
\textit{Output Format} -- type of generated presentation file;
\textit{Pref. Eval.} -- automatic evaluation via trained preference models (e.g., PREVAL, PPTEval);
\textit{Iter. Visual Opt.} -- post-render refinement of visual layout;
\textit{Fine Layout Ctrl.} -- explicit element placement using coordinates or bounding boxes;
\textit{Aesthetic Priors} -- design priors learned from expert-authored slides;
\textit{User Editability} -- support for easily modifying or extending the output (e.g., iterative user input loops or configurable parameters).
\textbf{Legend:} \cmark\;= supported;\; \pmark\;= partly supported;\; \xmark\;= not supported.
Table~\ref{tab:col-legend} shows the per-column scoring legend.}
\label{tab:auto-slide-comparison}

\centering
\setlength{\tabcolsep}{5pt}
\small

\resizebox{\textwidth}{!}{%
\begin{tabular}{@{}l*{9}{c}@{}}
\toprule
\textbf{Framework} &
\shortstack{\textbf{Content}\\\textbf{Struct.}} &
\shortstack{\textbf{Text--Fig}\\\textbf{Align.}} &
\shortstack{\textbf{Multi}\\\textbf{Modal}} &
\shortstack{\textbf{Output}\\\textbf{Format}} &
\shortstack{\textbf{Pref.}\\\textbf{Eval.}} &
\shortstack{\textbf{Iter.}\\\textbf{Visual Opt.}} &
\shortstack{\textbf{Fine}\\\textbf{Layout Ctrl.}} &
\shortstack{\textbf{Aesthetic}\\\textbf{Priors}} &
\shortstack{\textbf{User}\\\textbf{Editability}} \\
\midrule
PPSGen (2013)~\citep{hu2013ppsgen}                         & \pmark & \xmark & \xmark & text      & \xmark & \xmark & \xmark & \xmark & \xmark \\
D2S (2021)~\citep{sun2021d2s}                              & \pmark & \cmark & \cmark & text      & \xmark & \xmark & \xmark & \xmark & \xmark \\
DOC2PPT (2022)~\citep{fu2022doc2ppt}                       & \cmark & \cmark & \cmark & PPTX/PDF  & \xmark & \xmark & \cmark & \cmark & \xmark \\
Persona-Aware D2S (2024)~\citep{mondal2024presentations}   & \cmark & \pmark & \cmark & PDF       & \xmark & \xmark & \xmark & \xmark & \xmark \\
GDP (2024)~\citep{maheshwari2024presentations}             & \cmark & \xmark & \xmark & text      & \xmark & \xmark & \xmark & \xmark & \xmark \\
DocPres (2024)~\citep{bandyopadhyay2024enhancing}          & \cmark & \pmark & \cmark & PDF       & \xmark & \xmark & \pmark & \xmark & \xmark \\
PASS (2025)~\citep{aggarwal2025pass}                       & \cmark & \pmark & \cmark & PPTX      & \xmark & \xmark & \xmark & \xmark & \xmark \\
RCPS (2025)~\citep{xi2025multi}                            & \cmark & \cmark & \cmark & PPTX/PDF  & \cmark & \cmark & \cmark & \pmark & \xmark \\
PPTAgent (2025)~\citep{zheng2025pptagent}                  & \cmark & \cmark & \cmark & PPTX/HTML & \cmark & \cmark & \pmark & \pmark & \cmark \\
Auto-Slides (2025)~\citep{yang2025auto}                    & \cmark & \cmark & \cmark & PDF       & \xmark & \cmark & \pmark & \pmark & \xmark \\
AutoPresent (2025)~\citep{ge2025autopresent}               & \xmark & \pmark & \cmark & PPTX      & \cmark & \cmark & \cmark & \pmark & \xmark \\
SlideTailor (2026)~\citep{zeng2026slidetailor} & \cmark & \pmark & \cmark & PPTX & \xmark & \xmark & \pmark & \xmark & \pmark \\
SlideCraft (2025)~\citep{rao2025slidecraft}   & \cmark & \xmark & \cmark & PPTX & \xmark & \pmark & \pmark & \xmark & \cmark \\

\textbf{SlideGen (Ours)}                                   & \cmark & \cmark & \cmark & PPTX      & \cmark & \cmark & \cmark & \cmark & \cmark \\
\bottomrule
\end{tabular}%
}

\end{table*}

\begin{table*}[t]
  \caption{Per-column scoring legend for \cmark/~\pmark~/~\xmark.}
  \label{tab:col-legend}
  \centering
  \small
  \resizebox{\textwidth}{!}{
  \begin{tabular}{@{}l l l l@{}}
    \toprule
    \textbf{Column} & \textbf{\cmark\ (supported)} & \textbf{\pmark\ (partly)} & \textbf{\xmark\ (not)} \\
    \midrule
    \textbf{Content Struct.} & Clear slide-level outline (sections → slides → key points). & Topic list only. & No outline. \\
    \textbf{Text--Fig Align.} & Precise pairing with explicit loss or post-process. & Heuristic or example-based placement. & Not handled. \\
    \textbf{Multi Modal} & Text + images (optionally formulas/audio). & Partial multimodality. & Text only. \\
    \textbf{Pref. Eval.} & Trained preference model (e.g., PREVAL/PPTEval). & LLM heuristic scoring only. & ROUGE or human-only. \\
    \textbf{Iter. Visual Opt.} & Multi-round render-critique-revise. & Single-pass minor refine. & One-shot generation. \\
    \textbf{Fine Layout Ctrl.} & BBox/coordinates or constraints. & Coarse slots/templates. & None. \\
    \textbf{Aesthetic Priors} & Learns from high-quality slides (imitation/distillation). & Weak/indirect prior. & None. \\
    \textbf{User Editability} & Supports pre-generation modifications (e.g., feedback loops, customizable parameters). & -- & No user intervention before final output. \\
    \bottomrule
  \end{tabular}}
\end{table*}

\subsection{Additional Robustness Checks}

Table~\ref{tab:camera-ready-checks} summarizes the additional checks used to address reviewer concerns about cross-domain generalization, slide-count fairness, and layout robustness.

\begin{figure*}[!t]
\centering
\begin{subfigure}[t]{0.45\linewidth}
    \centering
    \includegraphics[width=\linewidth]{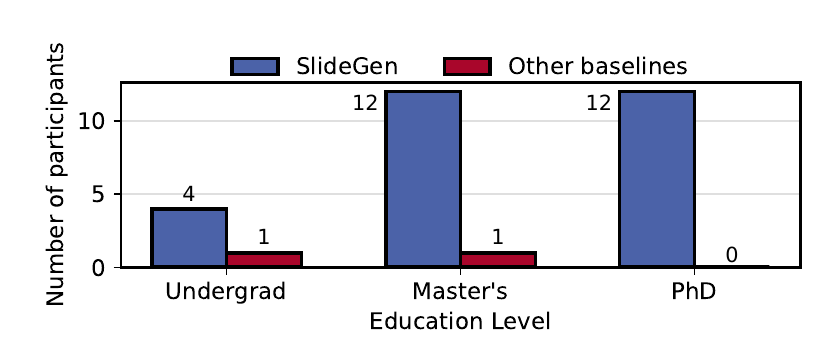}
    \vspace{-17pt}
    \Description{}
    \label{fig:preference_by_education}
\end{subfigure}
\begin{subfigure}[t]{0.45\linewidth}
    \centering
    \includegraphics[width=\linewidth]{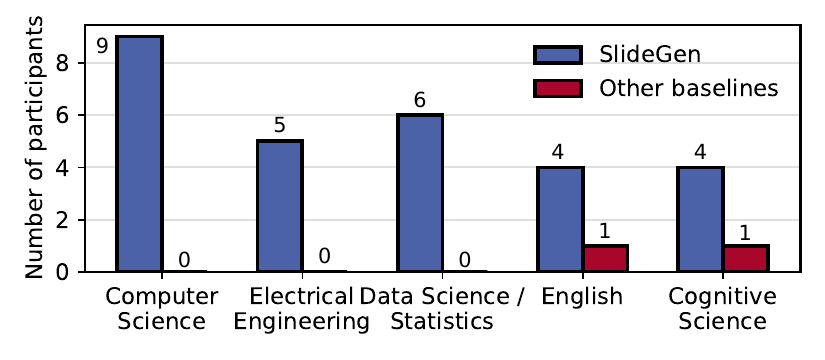}
    \vspace{-17pt}
\label{fig:preference_by_major}
\end{subfigure}
\vspace{-7pt}
\caption{Preference distributions for the slide deck generation task: (a) by education level; (b) by major.}
\label{fig:preference_by_education_major}
\vspace{-5pt}
\end{figure*}

\begin{table*}[!t]
\centering
\caption{\textbf{Additional   checks.} We report cross-domain stress-test results, slide-count decomposition, and overflow ablation to address reviewer concerns about generalization, evaluation fairness, and layout robustness.}
\label{tab:camera-ready-checks}
\scriptsize
\begin{minipage}[t]{0.31\textwidth}
\centering
\textbf{(a) Cross-domain stress test using GPT-5.}
\vspace{2pt}

\setlength{\tabcolsep}{3pt}
\resizebox{\linewidth}{!}{%
\begin{tabular}{lccc}
\toprule
\textbf{Domain} & \textbf{\#Papers} & \textbf{SlideQA} & \textbf{Judge} \\
\midrule
Biomedical  & 5  & 77.6 & 4.12 \\
Mathematics & 5  & 72.8 & 3.94 \\
\midrule
\textbf{Avg.} & 10 & 75.2 & 4.03 \\
\bottomrule
\end{tabular}}
\end{minipage}
\hfill
\begin{minipage}[t]{0.31\textwidth}
\centering
\textbf{(b) Slide-count decomposition.}
\vspace{2pt}

\setlength{\tabcolsep}{3pt}
\resizebox{\linewidth}{!}{%
\begin{tabular}{lcccc}
\toprule
\textbf{Method} & \textbf{\#Slides} & \textbf{\#Nav} & \textbf{\#Content} & \textbf{SlideQA} \\
\midrule
PPTAgent & 12.0 & 1.0 & 11.0 & 69.59 \\
Ours     & 17.9 & 6.2 & 11.7 & 80.19 \\
\bottomrule
\end{tabular}}
\end{minipage}
\hfill
\begin{minipage}[t]{0.31\textwidth}
\centering
\textbf{(c) Overflow fallback ablation.}
\vspace{2pt}

\setlength{\tabcolsep}{3pt}
\resizebox{\linewidth}{!}{%
\begin{tabular}{lc}
\toprule
\textbf{Setting} & \textbf{Overflow Rate} \\
\midrule
Original pipeline              & 5.3\% \\
+ bounded font scaling         & 0.9\% \\
+ LLM-based text compression   & 0.0\% \\
\bottomrule
\end{tabular}}
\end{minipage}
\vspace{-5pt}
\end{table*}

\subsection{Notation}
 \label{sec:notation} 
A deck consists of $N$ slides $\{s_i\}_{i=1}^{N}$.
Each slide has a role $r_i\in\{\texttt{title},\texttt{agenda},\texttt{content},\texttt{thanks}\}$.
For \texttt{content} slides we record an optional \emph{section} label $\sigma_i\in\Sigma$ and
\emph{subsection} label $\sigma'_i\in\Sigma'$.
We denote the pattern identifier by $\pi_i\in\Pi$ (e.g., \texttt{T1\_TextOnly}, \texttt{T4\_ImageTop}).

We consider a fixed slide layout: slide $s_1$ is the \texttt{title} page, slide $s_2$ is the \texttt{agenda} page, slides $s_3,\ldots,s_{N-1}$ are \texttt{content} pages, and the last slide $s_N$ is the \texttt{thanks} page. The content pages list \emph{section dividers} (``PART~1, PART~2, \dots''); these are the
\emph{agenda items}. Let
$
\mathcal{A}=[a_1,\ldots,a_m]
$
be the ordered list of top-level bullets on $s_2$.

Each slide carries a hierarchical string bullet list $B_i$, where each content box $b$ is defined as a pair $(u_{i,k}, \mathcal{S}_{i,k})$, and $u_{i,k}$ is the $k$-th top-level bullet.

Image, table, and formula assets on slide $i$ are denoted by the finite sets
$\mathcal{I}_i$ for image filenames, $\mathcal{T}_i$ for table filenames, and $\mathcal{F}_i$ for LaTeX strings, respectively. Optional speaker notes are written $n_i$. Let the slide area be $1$. For each region $b\in\mathcal{B}_i$, we denote its normalized width and height by $w_b,h_b$. The occupied area is the union area $\rho_i\in[0,1]$ of all non-background regions.

\subsection{Geometry-Aware Density }
\label{sec:Geometry-Aware}
This metric evaluates layout density with two components:
(i) area occupancy relative to a target $\tau$;
(ii) a concave quadratic preference over the effective number of content boxes, peaking at $M^\star$. 
 
\myparagraph{Why a downward-opening scoring function?} Overly monolithic slides look blocky and lack hierarchy, while excessive partitioning introduces noise and jumpy reading. A downward-opening scoring function over the effective region count captures the optimal range: it peaks near the preferred count $M^\star$ and decreases smoothly in both directions, penalizing slides with too few regions (sparse layouts) or too many (cluttered layouts) without imposing hard thresholds. The width $\kappa$ controls tolerance around $M^\star$, and the area gate $a_{\min}$ prevents gaming with tiny micro-regions. Combined with the occupancy term $1-|\rho_i-\tau|$, this yields an interpretable and reproducible measure that rewards layouts which are neither sparse nor cluttered. We also count only non-trivial regions via a minimum area gate $a_{\min}>0$:
\begin{equation}
M_i^{\mathrm{eff}}=\sum_{b\in\mathcal{B}_i}\mathbf{1}\!\left[A(b)\ge a_{\min}\right].
\end{equation}
Define a downward-opening quadratic fragmentation reward with maximum at $M^\star$:
\begin{equation}
R_i^{\mathrm{frag}}=\max\!\Big\{\,0,\ 1-\frac{(M_i^{\mathrm{eff}}-M^\star)^2}{\kappa}\,\Big\}\in[0,1].
\end{equation}
\begin{equation}
\mathrm{OM}_i \;\triangleq\; 1-\big|\rho_i-\tau\big|,\qquad
\mathrm{FR}_i \;\triangleq\; R_i^{\mathrm{frag}}.
\end{equation}

\begin{equation}
s_i^{\mathrm{geom}}
= \lambda_1\,\mathrm{OM}_i + \lambda_2\,\mathrm{FR}_i,
\qquad \lambda_1+\lambda_2=1,
\end{equation}
\begin{equation}
\mathrm{GAD} \;\triangleq\; \mathrm{DENSITY}^{\mathrm{geom}}
= \frac{1}{N}\sum_{i=1}^{N}\big(\lambda_1\,\mathrm{OM}_i + \lambda_2\,\mathrm{FR}_i\big).
\end{equation}

We set $a_{\min}=0.04,\;\tau=0.55$. The weights $\lambda_1$ and $\lambda_2$ are determined based on the values of $b_1$ and $b_2$, which are introduced later in the text. Specifically, we set:
\begin{equation}
  \lambda_1 = \frac{b_1}{b_1 + b_2}, \quad \lambda_2 = \frac{b_2}{b_1 + b_2}.
\end{equation}

\subsubsection{Sensitivity of GAD to $(\lambda_1,\lambda_2)$}
\label{sec:appendix-lambda-sensitivity}

To assess the sensitivity of GAD to the weighting between
occupancy matching and fragmentation reward, we sweep
$\lambda_1$ from~$0$ to~$1$ in steps of~$0.02$
(with $\lambda_2 = 1 - \lambda_1$), keeping
$(M^\star, \kappa)$ fixed at their calibrated values.
For each $\lambda_1$ we form the per-slide composite
$s_i^{\mathrm{geom}}
  = \lambda_1\,\mathrm{OM}_i + \lambda_2\,\mathrm{FR}_i$,
fit an affine mapping
$\hat{y} = a + b\,s^{\mathrm{geom}}$ by least squares,
and evaluate Spearman~$\rho$, Pearson~$r$, and RMSE under
the leave-one-deck-out (LODO) protocol described in
Section~\ref{sec:gad-human}.

\begin{figure}[t]
  \centering
  \includegraphics[width=\linewidth]{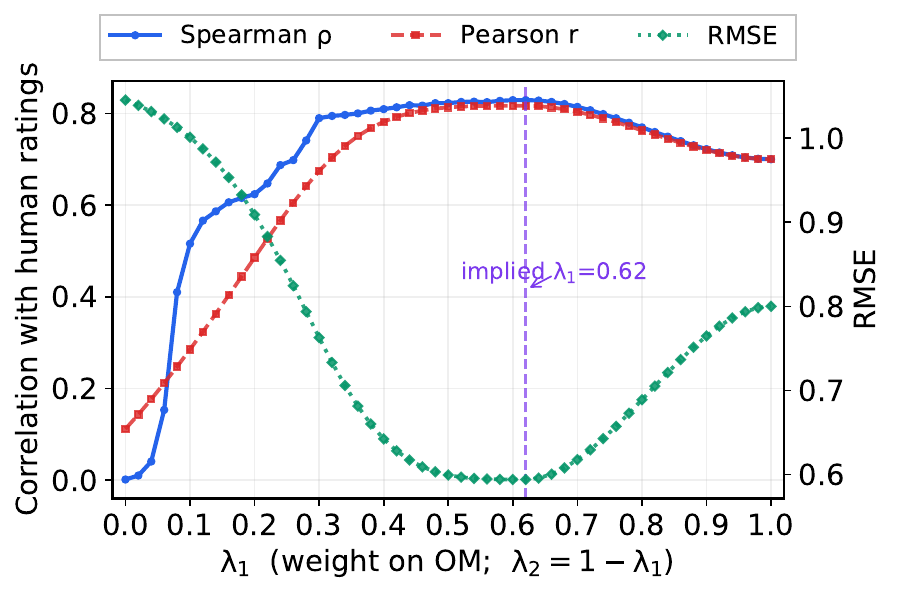}
  \caption{Sensitivity of GAD to $\lambda_1$.
    Left axis: Spearman~$\rho$ and Pearson~$r$;
    right axis: RMSE.
    The optimum lies near $\lambda_1 \approx 0.62$,
    and all metrics remain stable for
    $\lambda_1 \in [0.5,\, 0.7]$.}
  \Description{}
  \label{fig:lambda-sensitivity}
\end{figure}

As shown in Figure~\ref{fig:lambda-sensitivity},
all three metrics peak near $\lambda_1 \approx 0.62$
(Spearman $\rho = 0.820$, Pearson $r = 0.811$,
RMSE $= 0.584$) and degrade toward both extremes,
confirming that neither OM nor FR alone is sufficient
and both components contribute to predicting human-rated
layout aesthetics.
The score is robust within a broad range: for
$\lambda_1 \in [0.5,\, 0.7]$, Spearman~$\rho$ stays
above~$0.82$ and RMSE stays below~$0.63$,
indicating low sensitivity to moderate perturbations
of $\lambda_1$ around its optimum.

\begin{algorithm*} 
\caption{LODO training and prediction with linear regression mapping (OM/FR $\rightarrow$ human score)}
\label{alg:lodo}
\DontPrintSemicolon
\KwIn{Dataset $\mathcal{D}=\{(\mathrm{deck}_i,\mathrm{page}_i,y_i,\rho_i,M^{\mathrm{eff}}_i)\}_{i=1}^N$; fixed $a_{\min}, \tau$; grid $M^\star\in[m_{\min},m_{\max}]$, $\kappa\in[\kappa_{\min},\kappa_{\max}]$ with step $\Delta\kappa$}
\KwOut{Per-fold params $\{M^\star_d,\kappa_d,a_d,b_{1,d},b_{2,d}\}$; predictions $\{\hat y^{\mathrm{raw}},\hat y^{[1,5]}\}$}
Initialize prediction list $\mathcal{Y}\gets\emptyset$ and parameter table $\Theta\gets\emptyset$\;
\For{each deck $d$}{
  $\mathcal{D}_{\mathrm{train}} \gets \{\,i:\ \mathrm{deck}_i\neq d\,\}$,\quad $\mathcal{D}_{\mathrm{val}} \gets \{\,i:\ \mathrm{deck}_i=d\,\}$ \tcp*{leave-one-deck-out}
  $(M^\star_d,\kappa_d,a_d,b_{1,d},b_{2,d}) \gets \textsc{SelectAndFit}\!\left(\mathcal{D}_{\mathrm{train}},\ \tau,\ [m_{\min},m_{\max}],\ [\kappa_{\min},\kappa_{\max}],\ \Delta\kappa\right)$\;
  \For{each $i \in \mathcal{D}_{\mathrm{val}}$}{
    $\mathrm{OM}_i \gets 1 - \left|\rho_i - \tau\right|$\;
    $\mathrm{FR}_i \gets \max\!\left(0,\ 1 - \frac{(M^{\mathrm{eff}}_i - M^\star_d)^2}{\kappa_d}\right)$\;
    $\hat y^{\mathrm{raw}}_i \gets a_d + b_{1,d}\,\mathrm{OM}_i + b_{2,d}\,\mathrm{FR}_i$\;
    $\hat y^{[1,5]}_i \gets \mathrm{clip}(\hat y^{\mathrm{raw}}_i,\,1,\,5)$\;
    Append $(\mathrm{deck}_i,\mathrm{page}_i, y_i, \hat y^{\mathrm{raw}}_i, \hat y^{[1,5]}_i)$ to $\mathcal{Y}$\;
  }
  Record $(d,\ M^\star_d,\ \kappa_d,\ a_d,\ b_{1,d},\ b_{2,d})$ into $\Theta$\;
}
Compute overall Pearson/Spearman using $\hat y^{\mathrm{raw}}$ and RMSE using $\hat y^{[1,5]}$\;
\Return $\Theta$ and $\mathcal{Y}$\;
\end{algorithm*}

\begin{algorithm} 
\caption{SelectAndFit} 
\label{alg:selectfit}
\DontPrintSemicolon

\KwIn{train\_idx, tau, m\_min, m\_max, kappa\_min, kappa\_max, delta\_kappa}
\KwOut{best\_M, best\_kappa, a, b1, b2}

Extract y[i], rho[i], Meff[i] for i in train\_idx

\ForEach{i in train\_idx}{
  OM[i] $\leftarrow$ 1 - abs(rho[i] - tau)\;
}

best\_key $\leftarrow$ (-INF, -INF, +INF)\;
(best\_M, best\_kappa, a, b1, b2) $\leftarrow$ (0, 0, 0, 0, 0)\;

\For{Mstar $\leftarrow$ m\_min \textbf{to} m\_max}{
  \For{$\kappa \leftarrow \kappa_{\min}$; $\kappa \le \kappa_{\max}$; $\kappa \gets \kappa + \Delta\kappa$}{
    \ForEach{i in train\_idx}{
      FR[i] $\leftarrow$ max(0, 1 - ((Meff[i] - Mstar)$^2$)/(kappa))\;
    }
 
    (a, b1, b2) $\leftarrow$ \texttt{LinearLeastSquares}(y, [1, OM, FR])\;
 
    \For{i in train\_idx}{
      y\_raw[i]  $\leftarrow$ a + b1*OM[i] + b2*FR[i]\;
      y\_clip[i] $\leftarrow$ \texttt{clip}(y\_raw[i], 1, 5)\;
    }

    \tcp{Evaluate}
    pearson  $\leftarrow$ \texttt{Pearson}(y, y\_raw)\;
    spearman $\leftarrow$ \texttt{Spearman}(y, y\_raw)\;
    rmse     $\leftarrow$ \texttt{RMSE}(y, y\_clip)\;

    key $\leftarrow$ (pearson, spearman, -rmse)\;

    \If{key $>$ best\_key}{
      best\_key $\leftarrow$ key\;
      (best\_M, best\_kappa, a, b1, b2) $\leftarrow$ (Mstar, kappa, a, b1, b2)\;
    }
  }
}
\Return (best\_M, best\_kappa, a, b1, b2)\;
\end{algorithm}

\subsubsection{Training Method}
\label{subsec:GADmethod}

\myparagraph{Data and preprocessing.}
We prepared a total of 53 slide decks, with 40 used for training and 13 reserved for testing. For each deck $d$ and page $i$ we parse a JSON file that provides the page size, and the sizes, positions, and raw text content of content boxes $\mathcal{B}_i$.
Human judgments were obtained from $R=40$ recruited raters. Raters score each slide on a 1-5 scale (to one decimal place, in 0.1 increments): \textbf{1 }= extremely cluttered or extremely sparse; \textbf{3} = broadly acceptable but not ideal; \textbf{5} = very clean with appropriate information density. To remove differences in how strict or easy each rater scores, we perform per-rater $z$-score normalization:
 
\begin{equation}
\begin{aligned}
z^{(r)}_{d,i} &= \frac{s^{(r)}_{d,i}-\mu_r}{\sigma_r+\varepsilon}, \\
\mu_r &= \mathrm{mean}_{d,i}\!\left[s^{(r)}_{d,i}\right], \\
\sigma_r &= \mathrm{std}_{d,i}\!\left[s^{(r)}_{d,i}\right].
\end{aligned}
\end{equation}

\myparagraph{Density score and human score mapping.}
Instead of a manually set weighted sum, we learn an affine mapping from geometry score to the human scale:
\begin{equation}
\hat y_{d,i}= a + b_1\,\mathrm{OM}_{d,i} + b_2\,\mathrm{FR}_{d,i},
\end{equation}
with $(a,b_1,b_2)$ fit by least squares on training pages.  

\myparagraph{Grid search over $(M^\star,\kappa)$.}
We select $(M^\star,\kappa)$ by a grid
$M^\star\in\{m_{\min},\dots,m_{\max}\}$ and $\kappa\in\{\kappa_{\min},\kappa_{\min}+\Delta,\dots,\kappa_{\max}\}$, as shown in Figure~\ref{fig:all_in_one_row}(c) and detailed in Algorithm~\ref{alg:selectfit}.
For each grid point we recompute $\mathrm{FR}$, refit $(a,b_1,b_2)$ by least squares on the training split, and evaluate:
(i) Pearson’s $r$ on $\hat y$;
(ii) Spearman’s $\rho$ on $\hat y$;
(iii) RMSE on $\mathrm{clip}(\hat y,1,5)$.

\myparagraph{Cross-deck evaluation.}
\begin{figure}[t]
  \centering
  \includegraphics[width=0.95\linewidth]{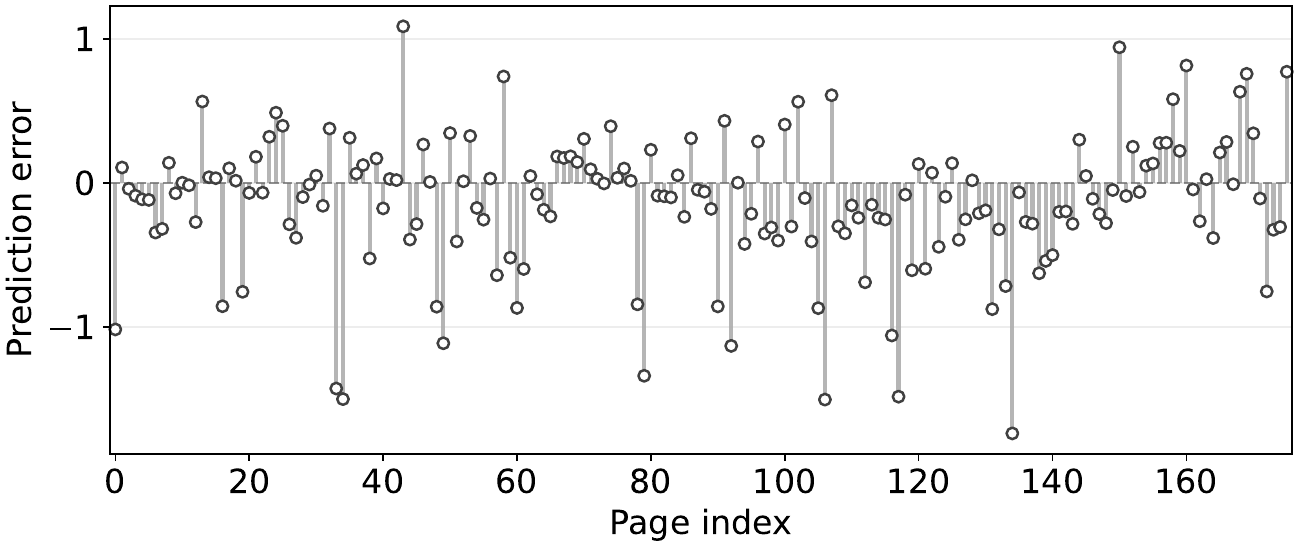}
  \caption{Per-page errors $(\hat y - y)$ under Leave-One-Deck-Out; dashed line at 0 indicates unbiased prediction.}
  \Description{}
  \label{fig:per_page_error}
\end{figure}
To assess generalization to unseen decks, we adopt Leave-One-Deck-Out (LODO) cross-validation, which holds out one entire deck at a time so that training and test pages never share presentation style; the full procedure is given in Algorithm~\ref{alg:lodo}. For each held-out deck $d^{\mathrm{val}}$:
(i) train on $\mathcal{D}\setminus\{d^{\mathrm{val}}\}$ by running the grid search to select $(M^\star,\kappa)$ and fitting $(a,b_1,b_2)$ by least squares;
(ii) validate on $d^{\mathrm{val}}$ by computing $(\mathrm{OM},\mathrm{FR})$ from the parsed layout and predicting $\hat y$ using the learned parameters.

We concatenate predictions across folds and report global Pearson and Spearman on un-clipped $\hat y$, and RMSE on clipped $\hat y$. Figure~\ref{fig:per_page_error} shows the per-page prediction errors across all folds.
 
\myparagraph{Implementation notes.}
We keep $a_{\min}$ and $\tau$ fixed; only $(M^\star,\kappa)$ are selected per fold by the grid, while $(a,b_1,b_2)$ are re-fit by least squares.
Per-rater $z$-score aggregation reduces rater bias and stabilizes the target scale.
Section pages (e.g., ``PART 01'') are filtered before feature extraction and learning.

\begin{table*}
  \caption{PPTEVAL dimensions and criteria (1--5 scale), adapted from \citep{zheng2025pptagent}.}
  \label{tab:ppteval-criteria}
  \centering
  \small
 
  \begin{tabular}{@{}>{\bfseries}l >{\raggedright\arraybackslash}p{0.78\linewidth}@{}}
    \toprule
    Dimension & \textbf{Criteria} \\
    \midrule
    \textcolor{BrickRed}{Content} &
    Text is concise and grammatically sound; key points are supported by relevant images. \\
    \textcolor{ForestGreen}{Design} &
    Harmonious colors and proper layout ensure readability; visual elements enhance appeal without clutter. \\
    \textcolor{Goldenrod}{Coherence} &
    Structure progresses logically and includes essential background information across the deck. \\
    \bottomrule
  \end{tabular}
\end{table*}
 
\begin{figure*}[!t]
\centering
\includegraphics[width=0.9\linewidth]{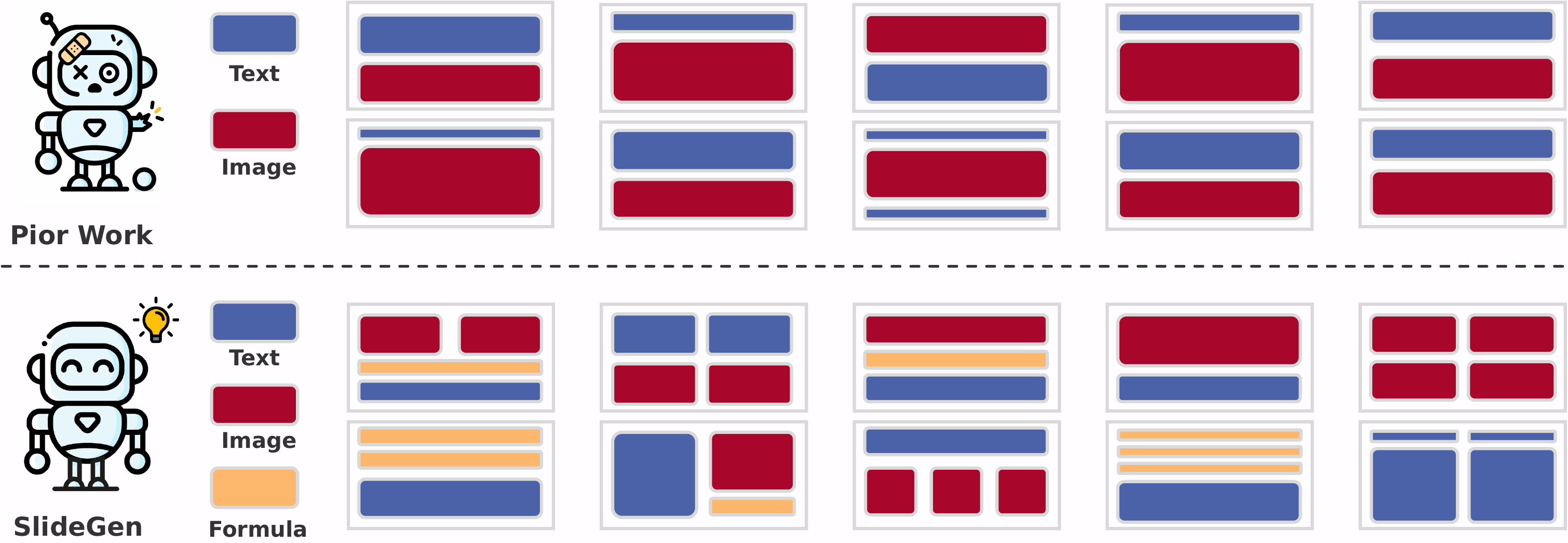}
\vspace{-2mm}
\caption{
\textbf{Comparison of generated slides with block abstractions.}
Each slide is shown as colored blocks, revealing that prior methods largely converge to similar vertical layouts, while \textbf{SlideGen} produces more varied and visually structured designs.
}
\Description{}
\label{fig:layout_comp}
\end{figure*}

\subsection{SlideQA Protocol}
\label{sec:metrics-slideqa}

\begin{figure*}[tb]
    \centering
\includegraphics[width=0.83\linewidth,page=12]{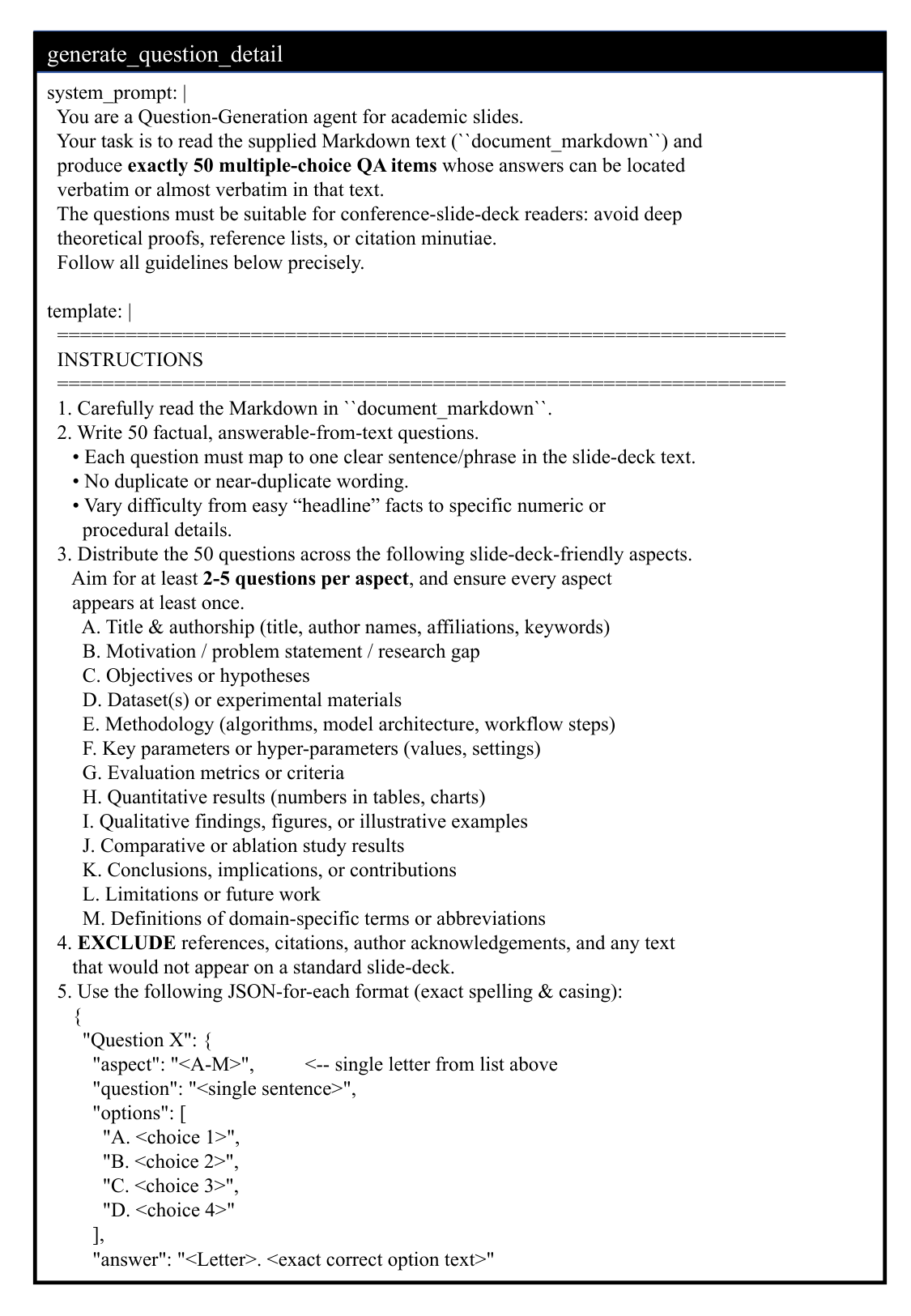}
\caption{The complete prompt used to construct both verbatim and interpretive questions, page 1 of 4.}
    \Description{}
    \label{fig:qa1}
\end{figure*}

\begin{figure*}[t]
    \centering
\includegraphics[width=0.83\linewidth,page=13]{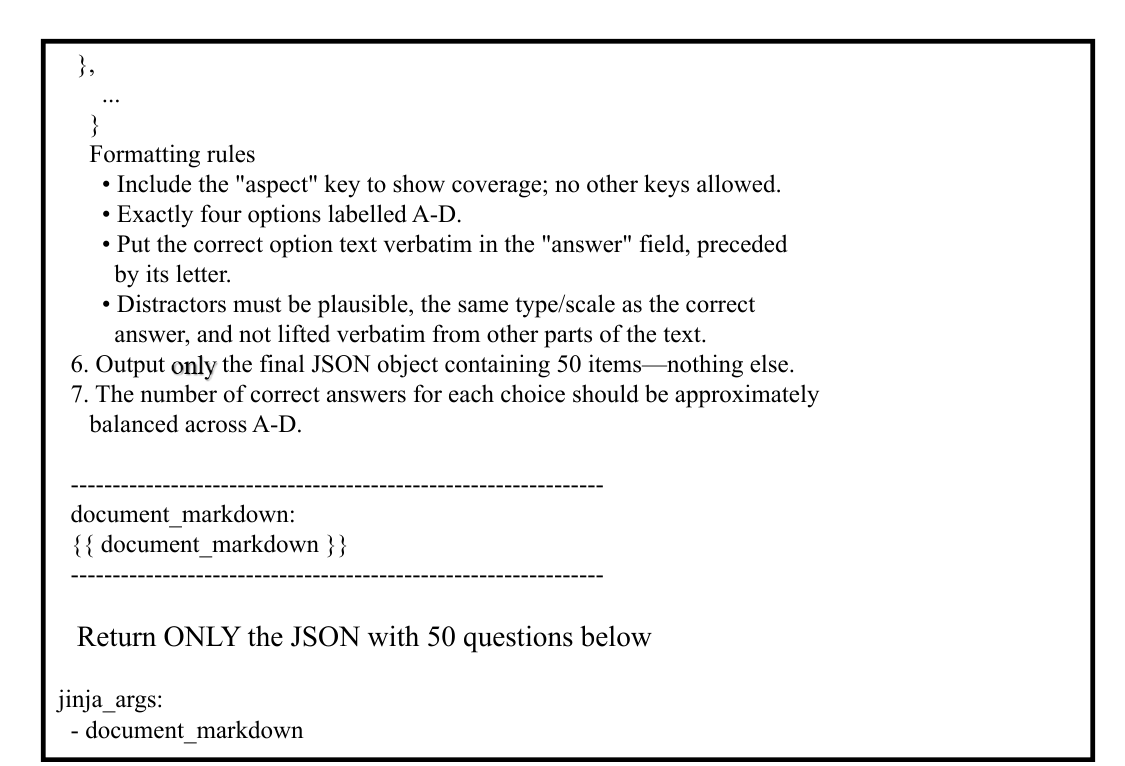}
    \caption{The complete prompt used to construct both verbatim and interpretive questions, page 2 of 4.}
    \Description{}
    \label{fig:qa2}
\end{figure*}

\begin{figure*}[!t]
    \centering
  \includegraphics[width=0.83\linewidth,page=14]{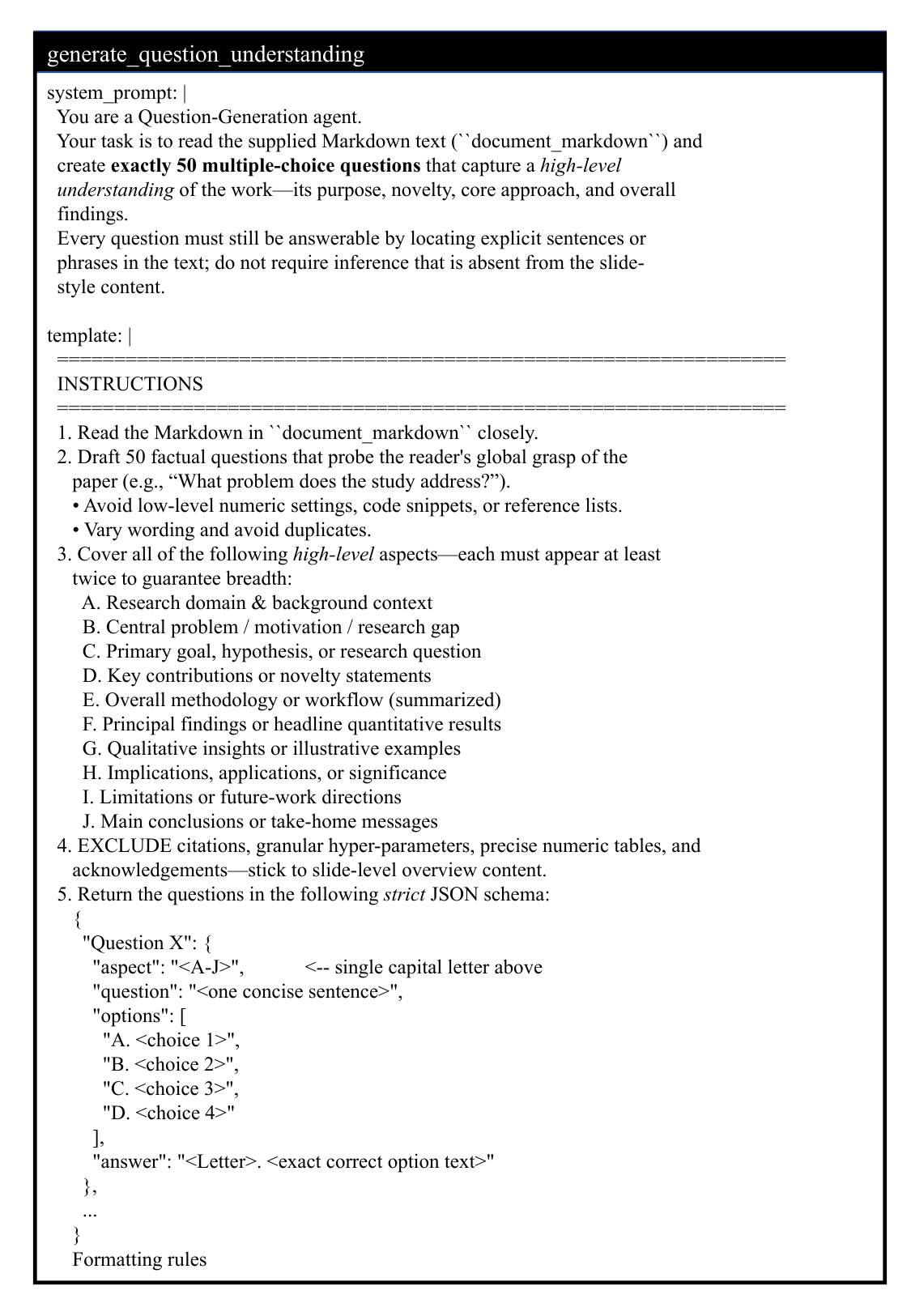}
    \caption{The complete prompt used to construct both verbatim and interpretive questions, page 3 of 4.}
    \Description{}
    \label{fig:qa3}
\end{figure*}

\begin{figure*}[!t]
    \centering\includegraphics[width=0.83\linewidth,page=15]{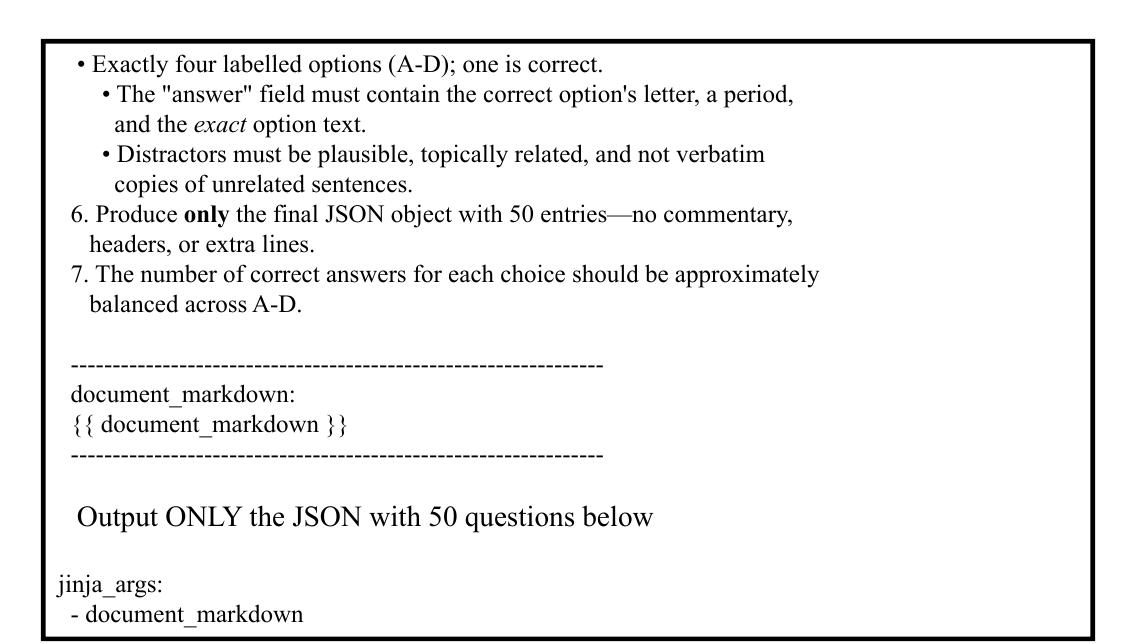}
    \caption{The complete prompt used to construct both verbatim and interpretive questions, page 4 of 4.}
    \Description{}
    \label{fig:qa4}
\end{figure*}

The protocol of SlideQA is as follows:
\textbf{(i) Question curation:}
For each source paper, we follow a deck-reader communication setup \citep{pang2025paper2poster} and employ GPT-4o as a question-generation model to produce $|\mathcal{Q}_{\mathrm{eval}}|=100$ multiple-choice questions per paper.
We construct two disjoint subsets:
$\mathcal{Q}_{\mathrm{verb}}$ with $|\mathcal{Q}_{\mathrm{verb}}|=50$ \emph{verbatim} questions directly answerable from the paper text, spanning 13 content aspects; and
$\mathcal{Q}_{\mathrm{int}}$ with $|\mathcal{Q}_{\mathrm{int}}|=50$ \emph{interpretive} questions targeting high-level comprehension across 10 conceptual dimensions.
We set $\mathcal{Q}_{\mathrm{eval}}=\mathcal{Q}_{\mathrm{verb}}\cup\mathcal{Q}_{\mathrm{int}}$ and $\mathcal{Q}_{\mathrm{verb}}\cap\mathcal{Q}_{\mathrm{int}}=\varnothing$.
\textbf{(ii) Respondents:}
Each image is presented to $M=6$ vision-language models, a mix of open- and closed-source systems, including three closed-source models: GPT-4o-mini, GPT-4o, and GPT-o3, and three open-source models: LLaVA-OV-7B, Qwen2.5-VL-7B-Instruct, and Phi-4-multimodal-instruct, to simulate reader standards from casual to expert \citep{pang2025paper2poster}.
Closed-source models generally achieve higher accuracy than open-source ones, so including both groups covers a range of reader ability levels. To make the reading setting both fair and realistic, we provide each model with the full slide deck, including both the rendered slides and the speaker notes produced by our Speaker Agent. Models must answer all questions based solely on this combined content. We report accuracy rate as our evaluation metric. The exact question-generation prompt is shown in Figures~\ref{fig:qa1}, \ref{fig:qa2}, \ref{fig:qa3}, \ref{fig:qa4}.

\myparagraph{Definition.}
Let $r_{q,m}\in\{0,1\}$ denote the correctness of model $m\in\{1,\ldots,M\}$ on question $q\in\mathcal{Q}_{\mathrm{eval}}$.
Define the per-question averaged correctness
\begin{equation}
\bar r_q \;=\; \frac{1}{M}\sum_{m=1}^M r_{q,m}.
\end{equation}
The SlideQA accuracy is then
\begin{equation}
s_R \;=\; \frac{1}{|\mathcal{Q}_{\mathrm{eval}}|}\sum_{q\in\mathcal{Q}_{\mathrm{eval}}} \bar r_q,
\label{eq:SlideQA-app}
\end{equation}
which averages correctness across both questions and models.
Subset scores restrict the sum in \eqref{eq:SlideQA-app} to $\mathcal{Q}_{\mathrm{verb}}$ and $\mathcal{Q}_{\mathrm{int}}$:
\begin{equation}
s_R^{\mathrm{verb}} \;=\; \frac{1}{|\mathcal{Q}_{\mathrm{verb}}|}\sum_{q\in\mathcal{Q}_{\mathrm{verb}}} \bar r_q,
\end{equation}

\begin{equation}
s_R^{\mathrm{int}} \;=\; \frac{1}{|\mathcal{Q}_{\mathrm{int}}|}\sum_{q\in\mathcal{Q}_{\mathrm{int}}} \bar r_q.
\end{equation}

\myparagraph{Rationale.}
This protocol simulates how readers glean information from slides: questions come from the paper, but answers must be inferred solely from the slides.

\subsection{Perplexity (PPL)}
\label{sec:metrics-ppl}
\myparagraph{What it measures.}
It quantifies the average next-token uncertainty of a language model over the deck text. Lower values indicate more fluent and predictable text. We compute this metric using the Llama-2-7b-hf language model.

\myparagraph{Definition.}
Let \(T(\cdot)\) be a fixed tokenizer and let
\[
x_{1:L} \;=\; T\!\Big(\mathrm{flat}(B_1)\ \Vert\ \cdots\ \Vert\ \mathrm{flat}(B_N)\Big)
\]
be the token sequence obtained by concatenating all slide texts. The full-sequence perplexity is
\begin{equation}
\operatorname{PPL} \;=\; \exp\!\left(-\frac{1}{L}\sum_{t=1}^{L}\log p_\theta\!\big(x_t \,\middle|\, x_{<t}\big)\right),
\end{equation}
where \(\log\) denotes the natural logarithm. Lower \(\operatorname{PPL}\) means higher predicted likelihood per token; \(\operatorname{PPL}=1\) corresponds to perfectly predictable text.

\setlength{\abovedisplayskip}{6pt}
\setlength{\belowdisplayskip}{6pt}
\setlength{\abovecaptionskip}{4pt}
\setlength{\belowcaptionskip}{4pt}

\section{Additional Analysis}
\label{sec:additional_analysis}

\begin{figure}[t]
\centering
\includegraphics[width=\linewidth]{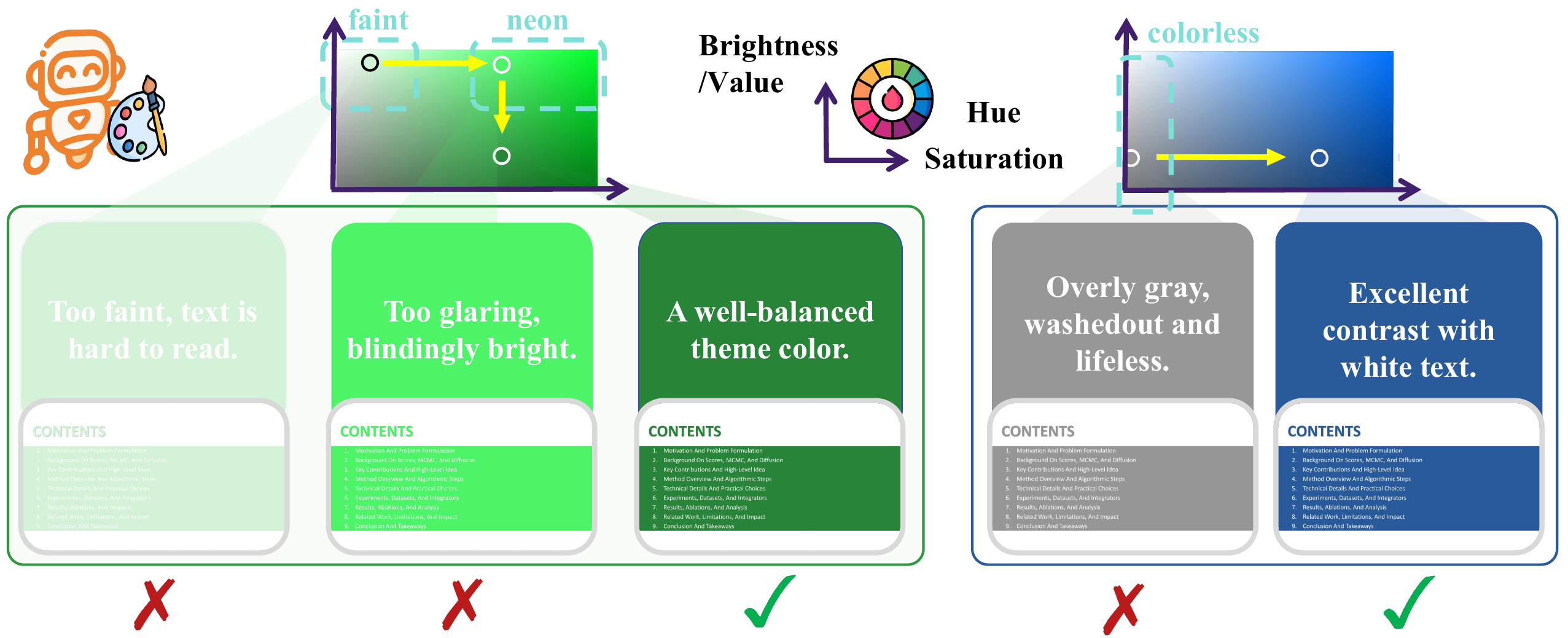}
\caption{\textbf{Color adjustment method on two fixed-hue planes for Refiner.} Examples on the left and right illustrate failure cases and the final readable and high-contrast choice.}
\vspace{-5pt}
\Description{}
\label{fig:color4}
\end{figure}

\subsection{Color Refinement Details}
\label{sec:color-refine}

In practice, the raw base color is often too faint for a presentation theme, as shown in the left panel of Figure~\ref{fig:color4}, which can hurt text readability. Refiner therefore refines the color on a fixed-hue HSV plane using a simple rule: first move right to make the color more vivid, but not garish, then move down to make it appropriately dark for a presentation theme. If the color becomes slightly too dark, we apply a small safety lift. When the base color is near gray and the hue is unstable, we lock the hue to a deep blue palette anchor and raise saturation. The right panel of Figure~\ref{fig:color4} illustrates the effect.

\subsection{Ablation Replacement Details}\label{sec:ablation_details}
To conduct a fair leave-one-agent-out ablation, we remove one agent at a time and replace it with a simple rule-based fallback that produces the same intermediate output format. All other modules, prompts, and the layout library remain unchanged. Each fallback is kept as simple as possible so that any performance drop can be clearly attributed to the removed agent rather than to format errors.

\textit{w/o Outliner.}
Instead of using Outliner agent, we directly convert each section heading in the paper into a slide title following the original document order. For each section, we extract the first three sentences as bullet points and create one slide. If the extracted text exceeds 120 tokens, we split it into two slides. No topic-level restructuring or merging is performed.

\textit{w/o Mapper.}
Instead of using Mapper agent, we assign each figure or table to the slide whose corresponding section first mentions it. If no mention is found, the asset is placed on the last content slide. No cross-slide reuse or relevance reasoning is applied.

\textit{w/o Formulizer.}
We skip equation extraction and placement entirely. No formula images or \LaTeX{} renderings appear on any slide. Textual references to equations in the bullet points are kept, but without the accompanying visual formulas.

\textit{w/o Arranger.}
We replace Arranger agent with a fixed template assignment rule. Slides with at least one image use \texttt{T3\_ImageLeft} with only the first image placed in the image slot. Slides without any image use \texttt{T1\_TextOnly}. Font sizes, spacing, and margins all use library defaults with no adaptive adjustment.

\textit{w/o Refiner.}
We skip all post-processing steps. No slide merging, no theme color extraction, and no visual refinement are applied. The deck produced by Arranger is directly used as the final output.

\textit{w/o Speaker.}
We skip speaker note generation entirely. The notes field of every slide is left empty. The placement explanations from Mapper and Formulizer that would normally appear in the notes are also removed. Since Speaker operates on the narrative layer rather than the visual deck itself, its removal has minimal impact on slide-only metrics such as GAD. The modest drop in VLM-as-Judge reflects reduced cross-slide narrative coherence rather than degraded visual or content quality, consistent with Speaker's design as a presentation-delivery component.

\myparagraph{Agent-Level Ablation.}
Table~\ref{tab:ablation_compact} reports the leave-one-agent-out results. We highlight three key findings.
(1) \textbf{Outliner is the most critical agent.} Removing it causes the largest drops across all metrics, confirming that global structure planning is the foundation for content coverage, language quality, and downstream agent performance.
(2) \textbf{Arranger and the layout library drive visual quality.} Removing Arranger leads to the most severe D-Avg degradation among all variants. Retaining Arranger but restricting it to a single template partially recovers the score, showing that both the agent's layout reasoning and the diversity of the layout library contribute to visual quality.
(3) \textbf{Each agent addresses a distinct dimension.} Speaker removal has almost no effect on density but slightly lowers the Judge score, while Mapper removal primarily hurts density and interpretive accuracy. Formulizer shows the mildest overall impact, though it remains important for equation-heavy papers. These complementary patterns confirm that the six agents are not redundant but target different quality dimensions.
 
\subsection{Prompt Structure Sensitivity}\label{sec:packaging-sensitivity}
\begin{figure}[t]
  \centering
  \includegraphics[width=0.6\linewidth]{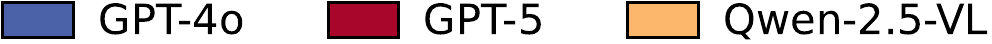}
  \vspace{2pt}

  \begin{subfigure}[t]{0.48\linewidth}
    \centering
    \includegraphics[width=\linewidth]{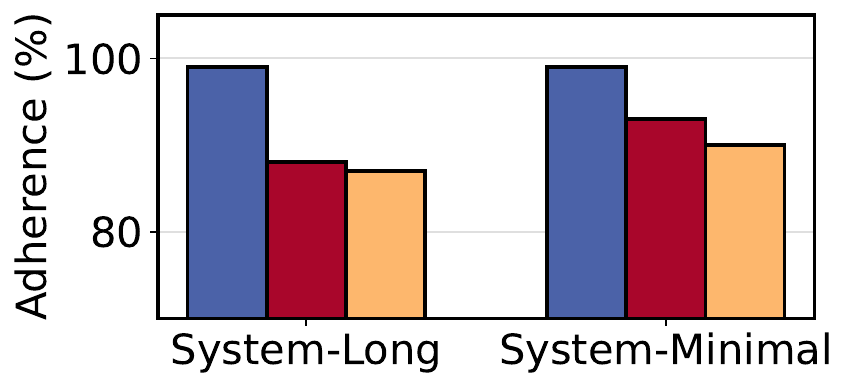}
    \vspace{-14pt}
    \caption{Speaker: Output Alignment.}
    \Description{}
    \label{fig:pack-speaker-adherence}
  \end{subfigure}\hfill
  \begin{subfigure}[t]{0.48\linewidth}
    \centering
    \includegraphics[width=\linewidth]{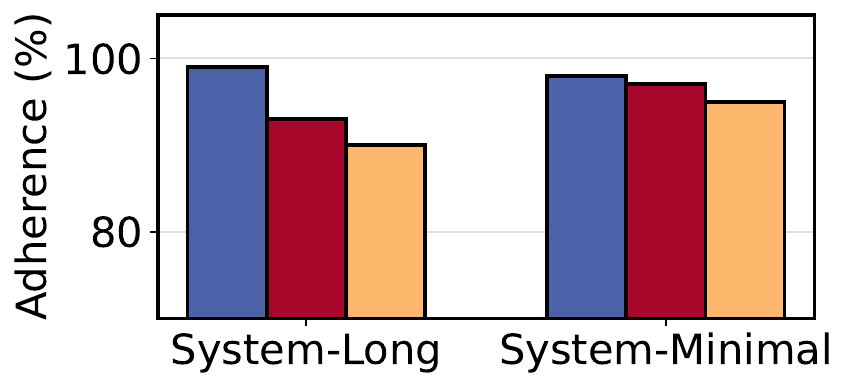}
    \vspace{-14pt}
    \caption{Arranger: Output Alignment.}
    \Description{}
    \label{fig:pack-arranger-adherence}
  \end{subfigure}


  \begin{subfigure}[t]{0.48\linewidth}
    \centering
    \includegraphics[width=\linewidth]{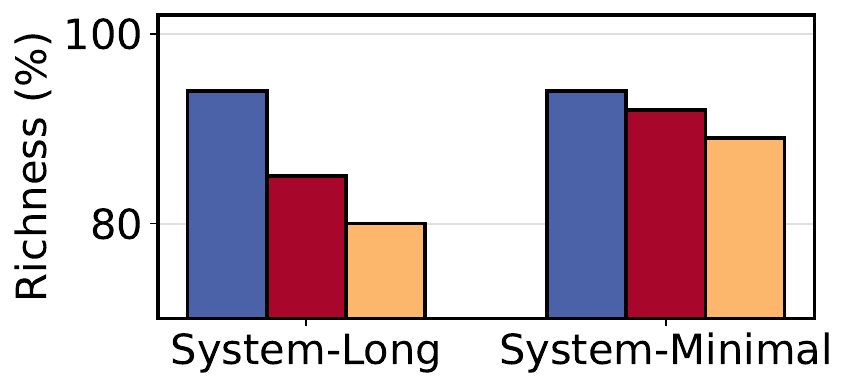}
    \vspace{-14pt}
    \caption{Speaker: Content Richness.}
    \Description{}
    \label{fig:pack-speaker-collapse}
  \end{subfigure}\hfill
  \begin{subfigure}[t]{0.48\linewidth}
    \centering
    \includegraphics[width=\linewidth]{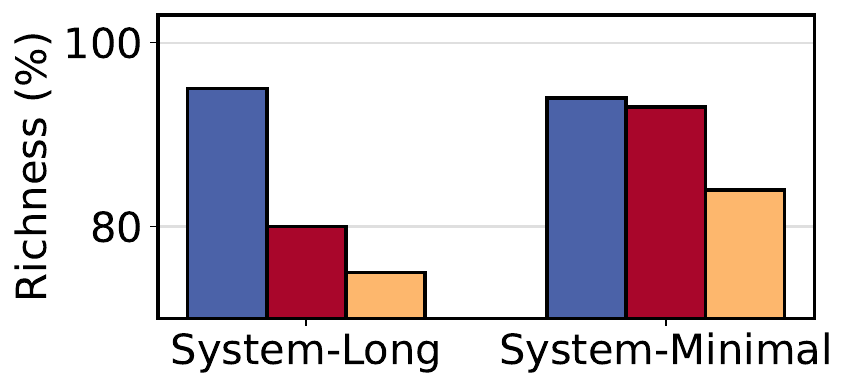}
    \vspace{-14pt}
    \caption{Arranger: Content Richness.}
    \Description{}
    \label{fig:pack-arranger-collapse}
  \end{subfigure}

  \caption{\textbf{Prompt structure sensitivity across backbones.} The top row shows Output Alignment (percentage of format-compliant outputs) for Speaker and Arranger under two prompt structures. The bottom row shows Content Richness (normalized by model-specific baselines) for the same tasks.}
  \label{fig:packaging-sensitivity-2x2}
\end{figure}

Prompt organization can influence output quality independently. We investigate whether separating formatting instructions from the system prompt into a dedicated template block affects format compliance and content detail, and whether this sensitivity varies across backbones (GPT-4o, GPT-5, and Qwen2.5-VL-7B).

\myparagraph{Experimental Setup.}
We evaluate two types of prompt structures: (i) \textbf{System-Long}: The system prompt includes both the high-level task description and detailed instructions for output formatting, including format constraints (such as JSON schema) and example outputs. (ii) \textbf{System-Minimal + Template-Based Output}: The system prompt is limited to a single sentence outlining the task's objective, with detailed formatting instructions, JSON schema, and example outputs moved into a separate template block. 

The experiment involves two tasks within the SlideGen pipeline: Speaker and Arranger. In each case, the system prompt is altered according to the two prompt structures, while the input paper and model settings remain fixed.

\myparagraph{Metrics.}
We assess the performance of each model with the following key metrics: (i) \textbf{Output Alignment (↑)}: The proportion of outputs that strictly adhere to the required format constraints (such as the JSON schema). Higher output alignment indicates better compliance with the specified output structure. (ii) \textbf{Content Richness (↑)}: This metric measures the richness of the generated content by counting the number of JSON "lines" (bullet points and fields). We normalize the richness by using model-specific baselines: For GPT-5, 600 lines = 100; for GPT-4o and Qwen, 250 lines = 100. Higher content richness indicates a more detailed output, with more sections and subsections generated by the model.

\myparagraph{Results.}
 Figure~\ref{fig:packaging-sensitivity-2x2} presents the experimental results, showing how the two prompt structures (System-Long vs. System-Minimal and Template-Based Output) affect the performance of GPT-4o, GPT-5, and Qwen across Speaker and Arranger tasks in Content Richness.

\myparagraph{Analysis.}
Based on the results, we find that switching from System-Long to System-Minimal + Template-Based Output substantially increases GPT-5’s content richness, producing outputs with more sections and greater structural depth.
In contrast, GPT-4o achieves consistently high content richness under both prompt structures. This stability suggests that GPT-4o is less sensitive to prompt organization and produces detailed outputs even with the simpler System-Long format.
Qwen2.5-VL-7B, while showing some improvement with the System-Minimal + Template-Based Output structure, still lags behind both GPT-4o and GPT-5 in content richness.

\myparagraph{Conclusion.}
For GPT-5, keeping the system prompt minimal and moving formatting instructions into the template block substantially improves content richness. GPT-4o is largely insensitive to prompt organization, while Qwen benefits modestly. These results suggest that prompt structure deserves careful tuning.

\subsection{Dataset Scope and Generalizability}
\label{appendix:dataset_scope}

\myparagraph{Dataset Scope.}
Our dataset consists of research papers from top-tier AI conferences. We chose this domain for two reasons: first, our expertise in AI enables reliable evaluation of slide quality; second, scientific papers share a common structure (introduction, methods, results) across disciplines, so the methodology generalizes beyond CS to fields such as medicine, physics, and biology.

\myparagraph{Generalizability.}
Our core framework can extend to other document types, including business reports and legal documents. The primary adaptation is expanding the layout library with domain-appropriate templates (e.g., image-centric layouts for art portfolios or data-heavy layouts for financial reports). Because template creation is independent of the pipeline, designers can add new templates without modifying any agent or algorithm.

\section{Abbreviations}
\label{sec:Abbreviations}
We provide a reference for the abbreviations of models used in this paper, as shown in Table~\ref{tab:model-abbr}.
\begin{table}
\centering
\caption{Reference for model abbreviations in this paper.}
\setlength{\tabcolsep}{10pt}
\resizebox{\linewidth}{!}{%
\begin{tabular}{ll}
\toprule
\textbf{Abbreviation} & \textbf{Full Name} \\
\midrule
\textbf{4o-mini}      & GPT-4o-mini \\
\textbf{4o}           & GPT-4o \\
\textbf{o3}           & GPT-o3 \\
\textbf{llava-ov-7b}  & LLaVA-OneVision-Qwen2-7b-ov-hf~\citep{li2024llavaonevisioneasyvisualtask} \\
\textbf{Qwen2.5-VL-7B}  & Qwen2.5-VL-7B-Instruct~\citep{Qwen2VL,Qwen-VL} \\
\textbf{Qwen2.5-VL-72B}  & Qwen2.5-VL-72B-Instruct~\citep{Qwen2VL,Qwen-VL,qwen2.5-VL} \\
\textbf{Phi-4-MM}     & Phi-4-multimodal-instruct~\citep{abouelenin2025phi} \\
\bottomrule
\end{tabular}
}
\vspace{-5pt}
\label{tab:model-abbr}
\end{table}

\section{Samples Generated by Our Pipeline}
\label{sec:slidegen_sample}
\begin{figure*}[t]
  \centering
  \includegraphics[width=0.90\linewidth,page=1]{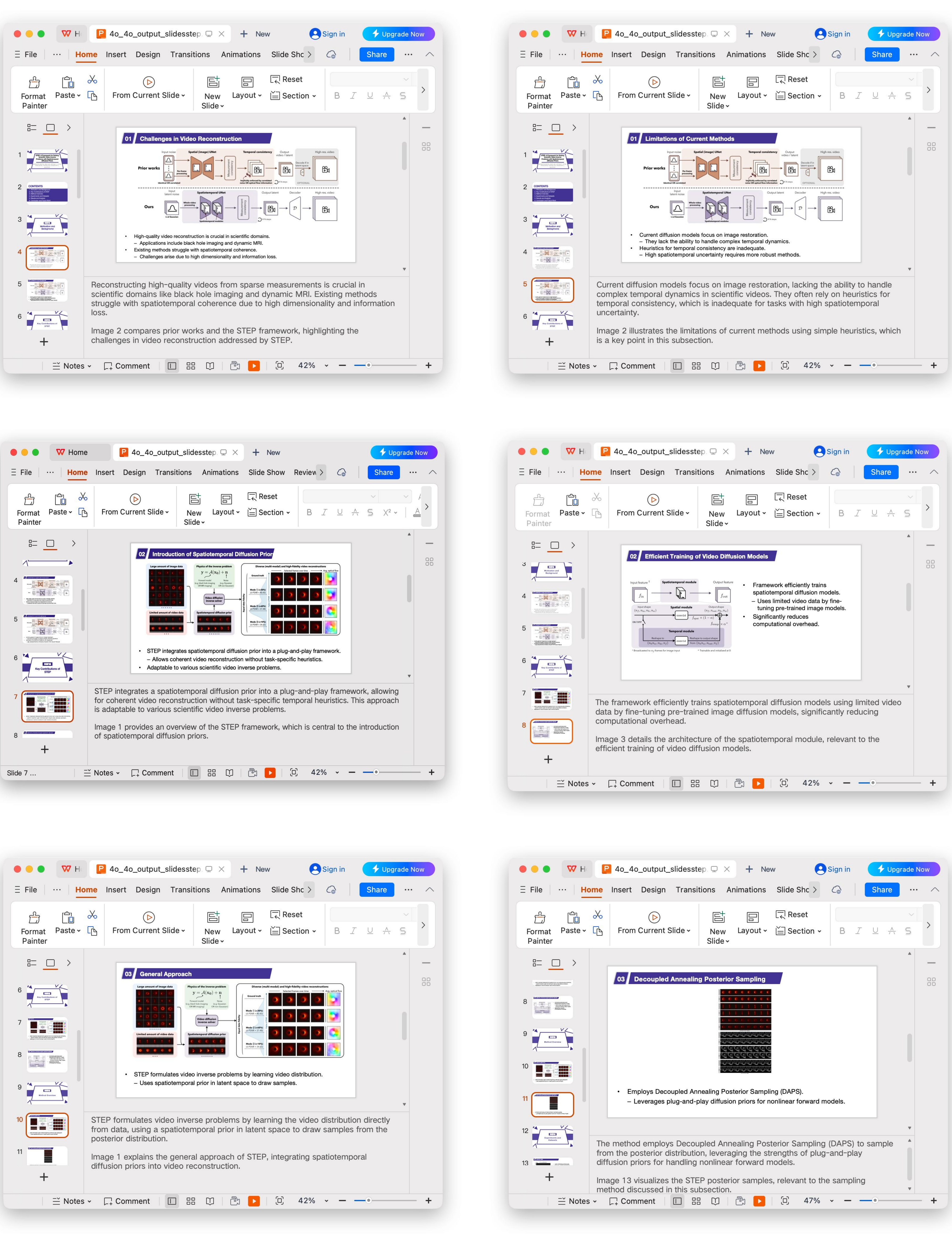}
  \caption{SlideGen output with extracted color palette and speaker notes (page 1 of 2).}
  \Description{}
  \label{fig:note_a}
\end{figure*}

\begin{figure*}[t]
  \centering
  \includegraphics[width=0.90\linewidth,page=1]{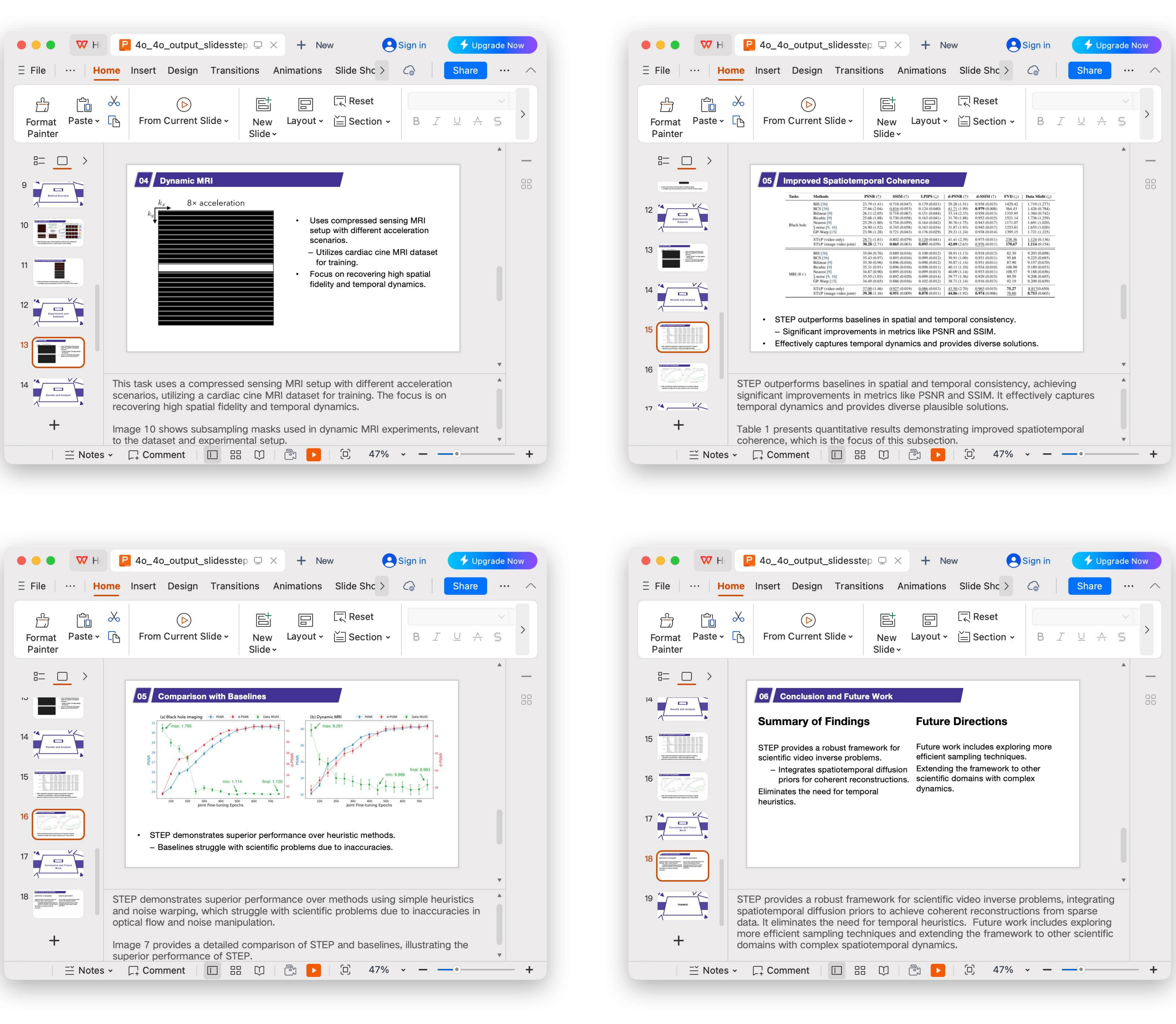}
  \caption{SlideGen output with extracted color palette and speaker notes (page 2 of 2).}
  \Description{}
  \label{fig:note_b}
\end{figure*}

\begin{figure*}[!t]
  \centering
  \includegraphics[width=0.84\linewidth,page=1]{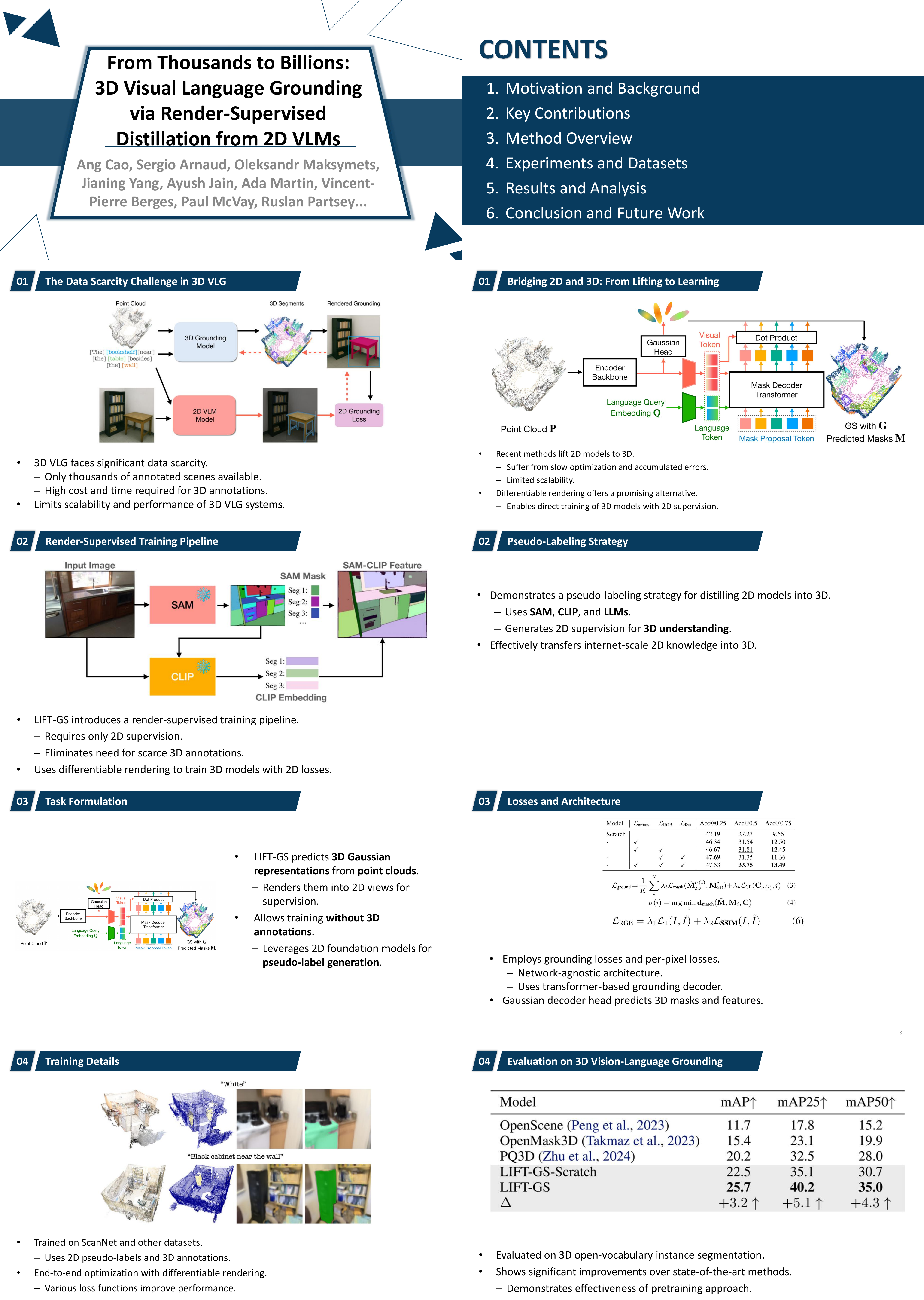}
  \caption{SlideGen output using GPT-4o with the default blue theme (page 1 of 2).}
  \Description{}
  \label{fig:example11}
\end{figure*}

\begin{figure*} [!t]
    \centering
    \includegraphics[width=\linewidth,page=2]{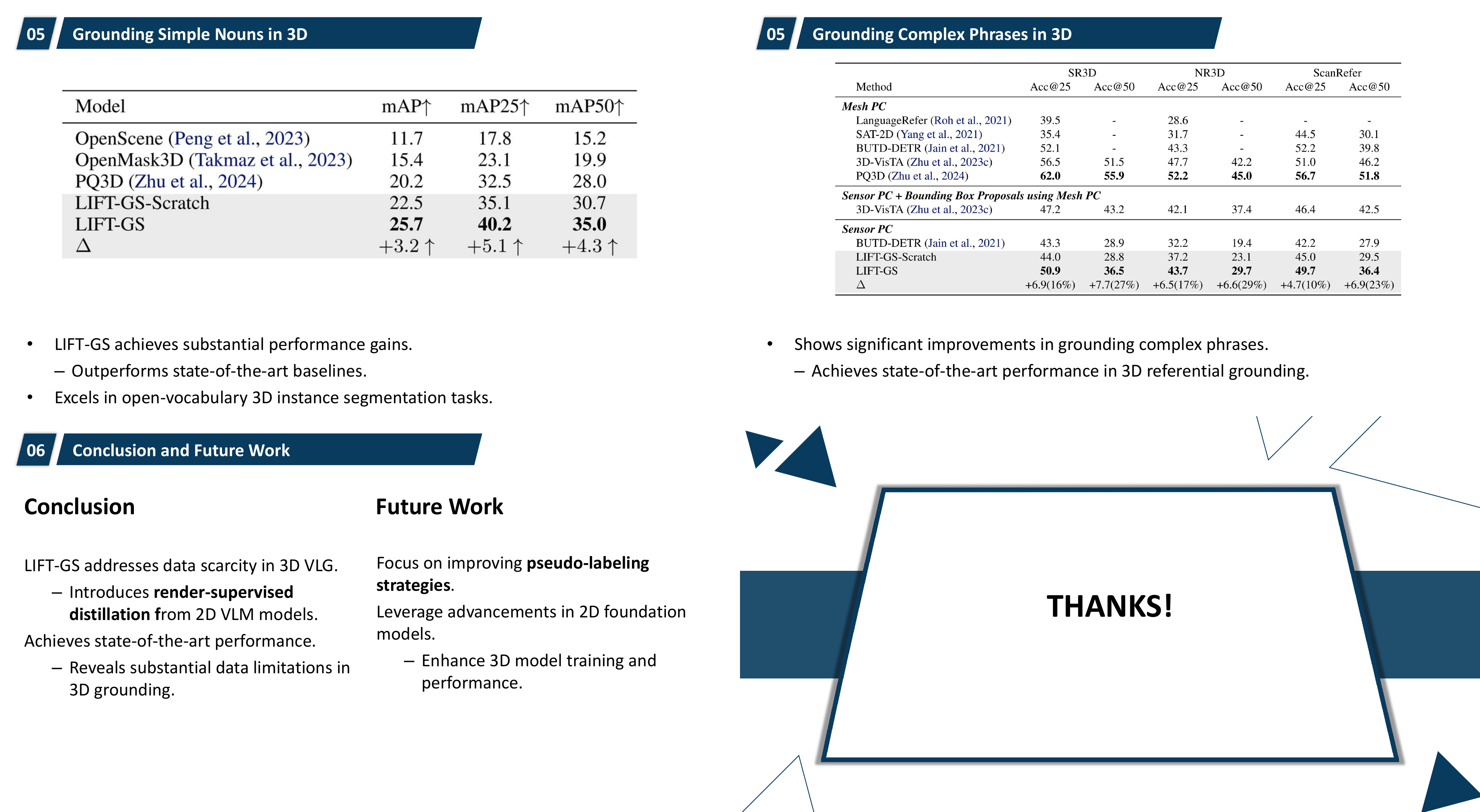}
    \caption{SlideGen output using GPT-4o with the default blue theme (page 2 of 2).}
    \Description{}
    \label{fig:example12}
\end{figure*}

\begin{figure*} [!t]
    \centering
    \includegraphics[width=0.84\linewidth,page=1]{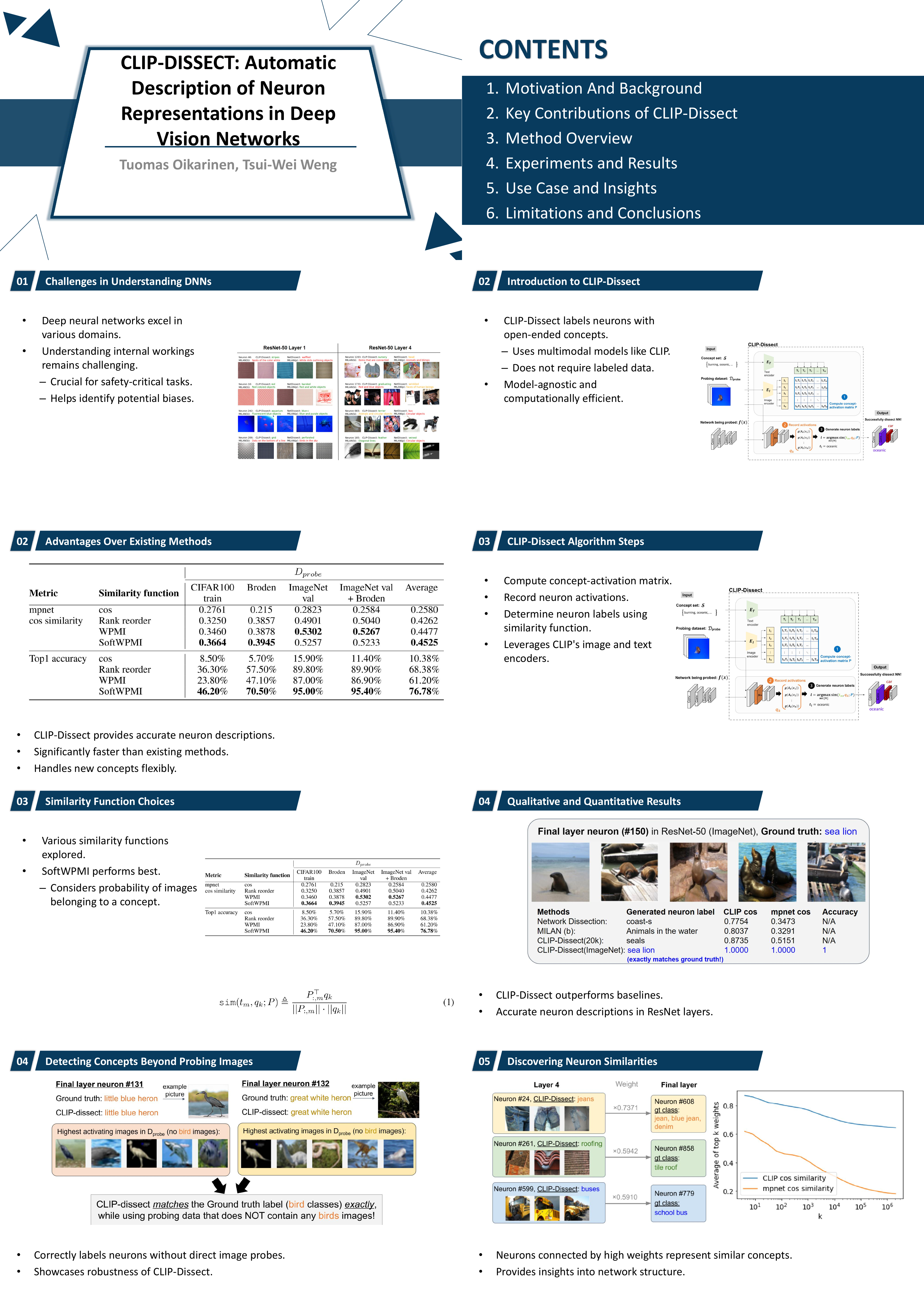}
    \caption{SlideGen output using GPT-4o with the default blue theme.}
    \Description{}
    \label{fig:example13}
\end{figure*}

\begin{figure*} 
    \centering
    \includegraphics[width=0.84\linewidth,page=1]{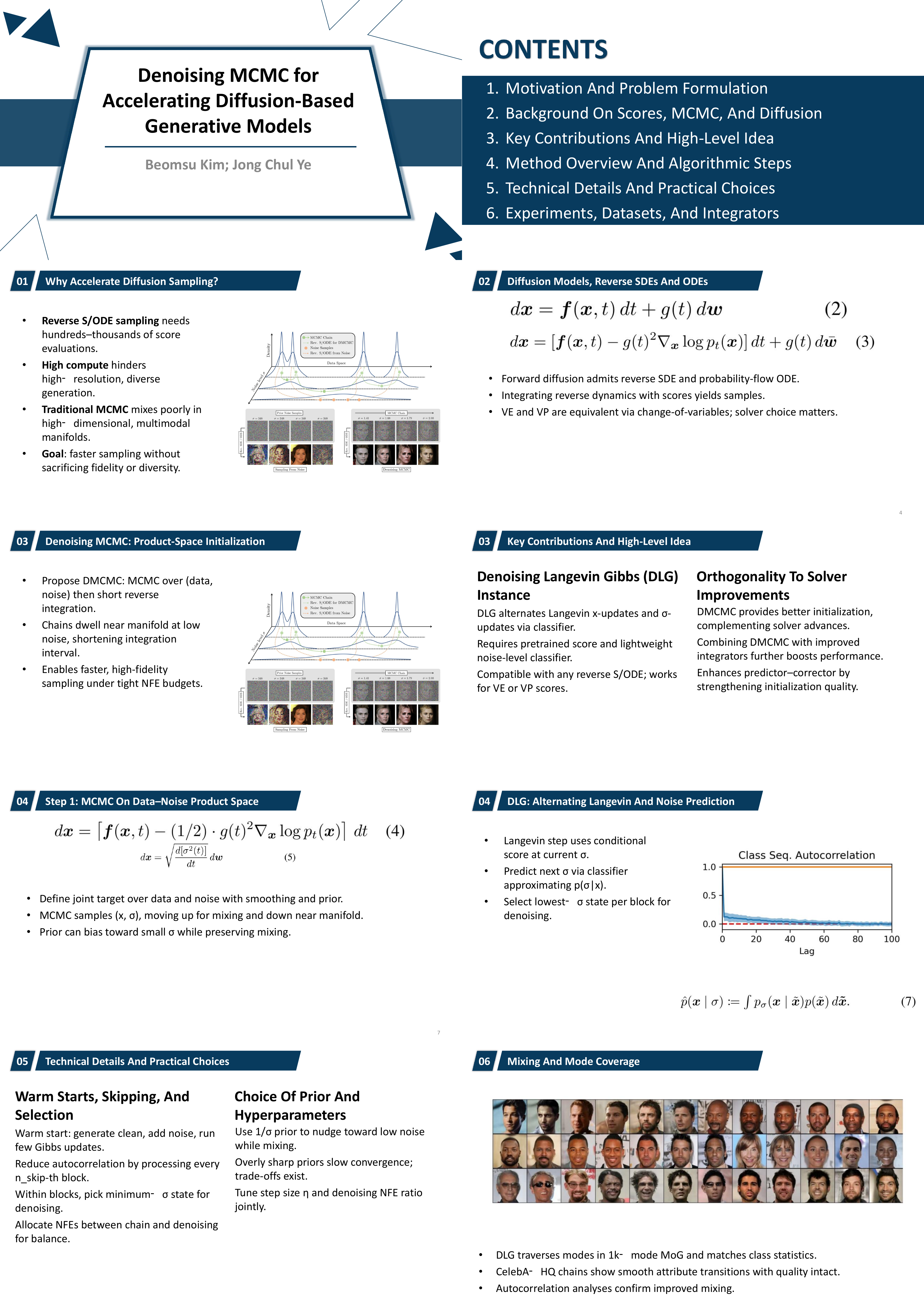}
    \caption{SlideGen output using GPT-5 with the default blue theme (page 1 of 2).}
    \Description{}
    \label{fig:example14}
\end{figure*}

\begin{figure*}[t]
    \centering
    \includegraphics[width=0.86\linewidth,page=2]{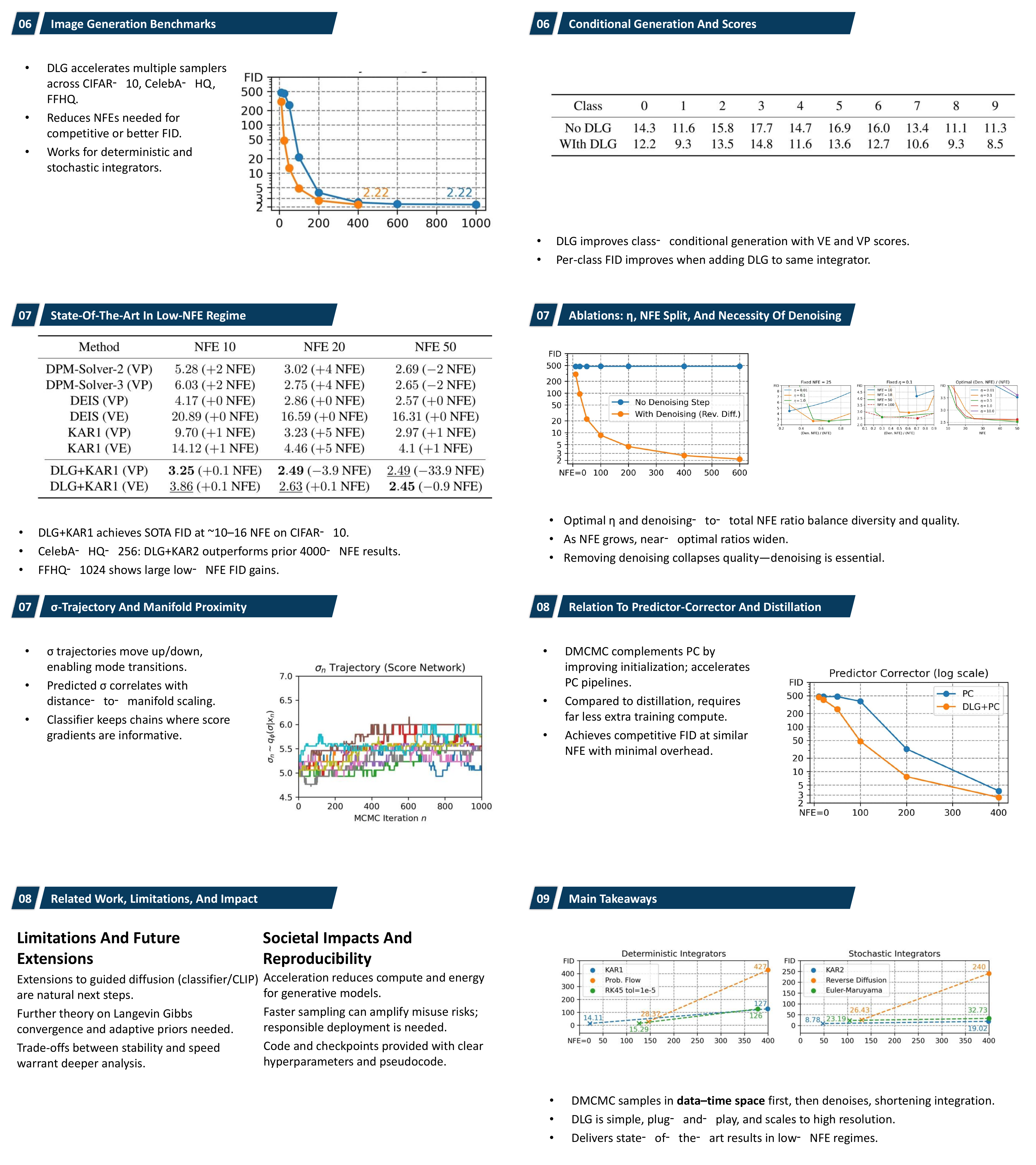}
    \caption{SlideGen output using GPT-5 with the default blue theme (page 2 of 2).}
    \Description{}
    \label{fig:example15}
\end{figure*}

\begin{figure*}[t]
    \centering
    \includegraphics[width=\linewidth,page=1]{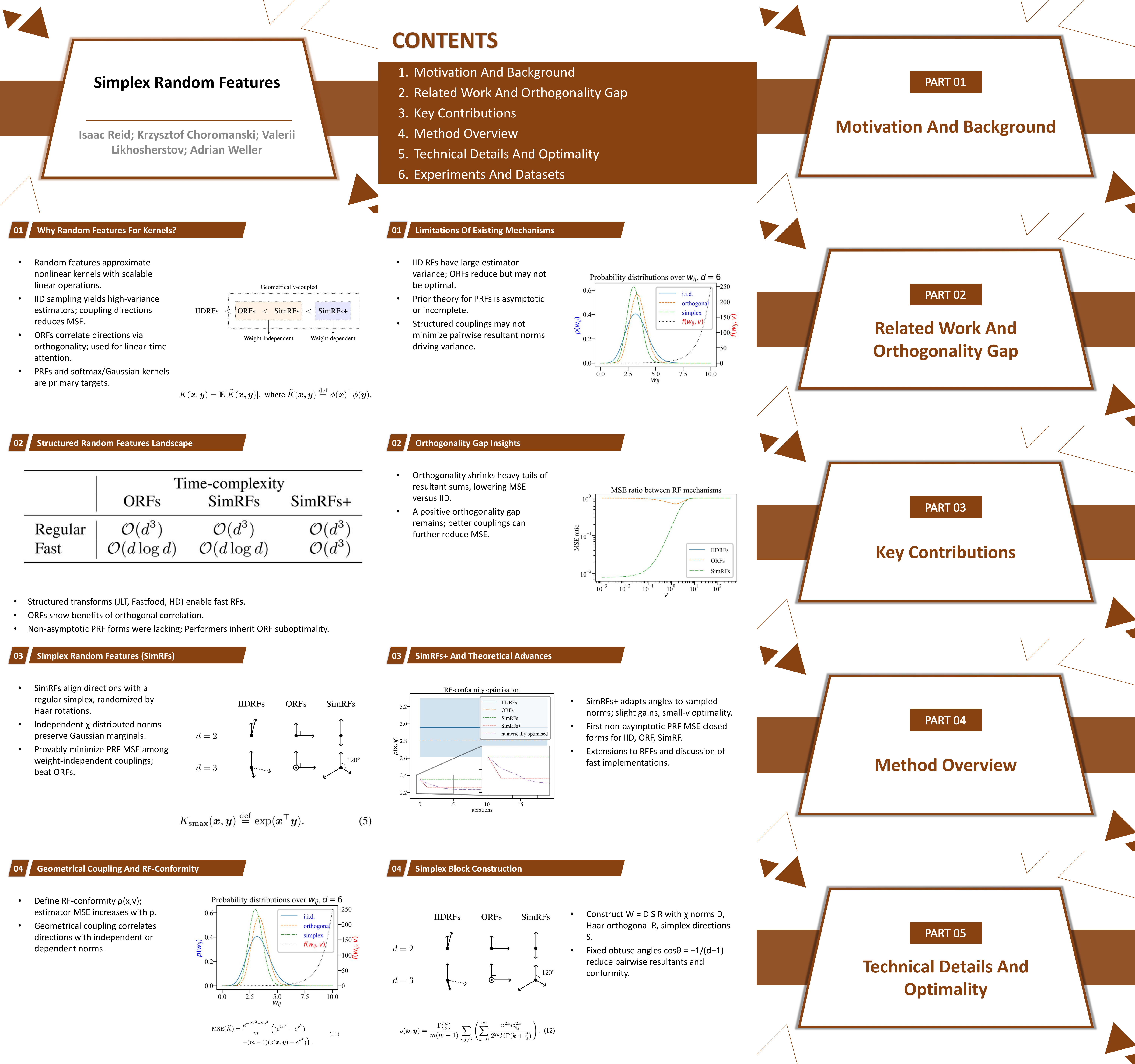}
    \caption{SlideGen output with colors extracted and refined by Refiner (GPT-5, page 1 of 2).}
    \Description{}
    \label{fig:color1}
\end{figure*}

\begin{figure*}[t]
    \centering
    \includegraphics[width=\linewidth,page=2]{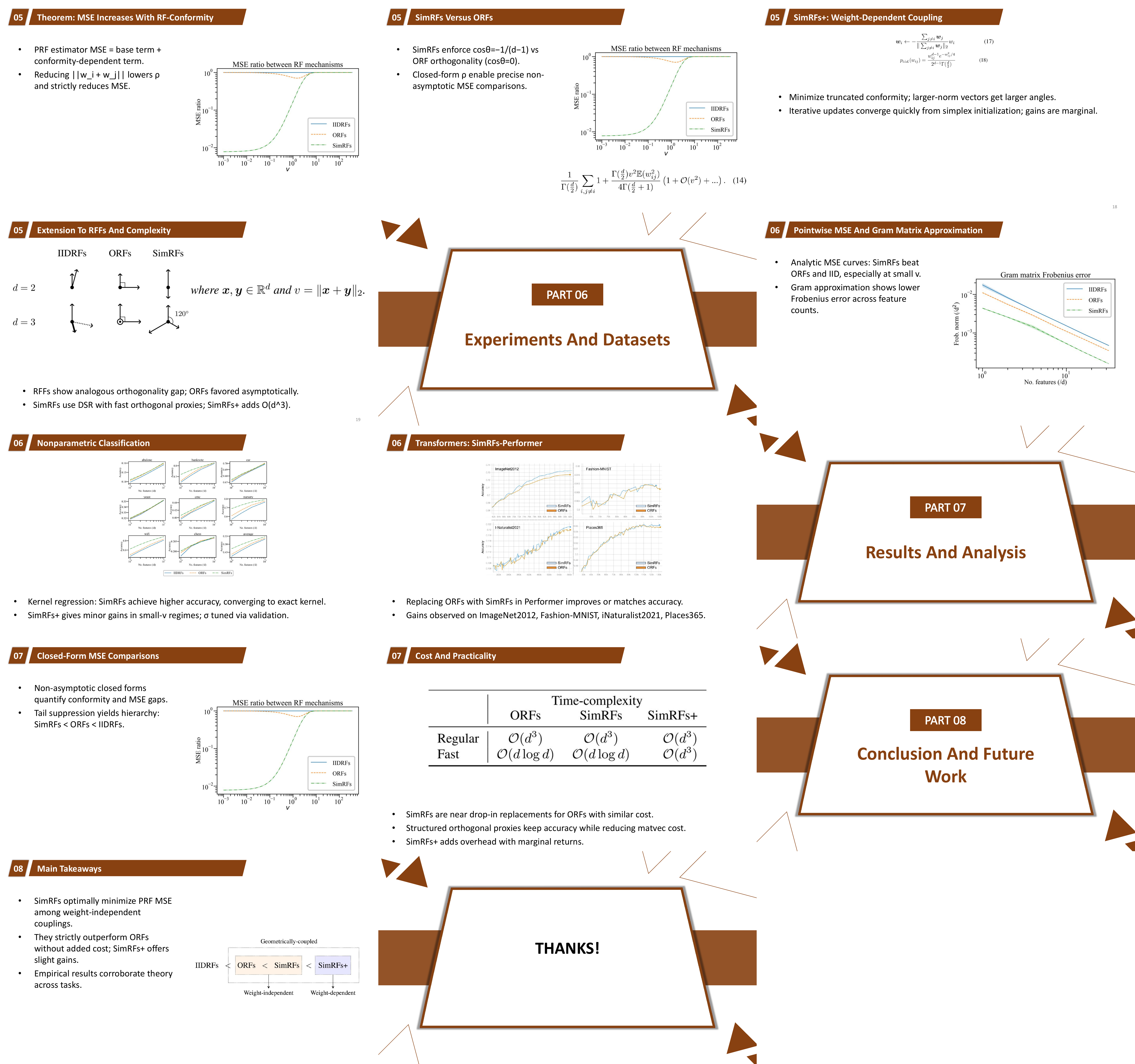}
    \caption{SlideGen output with colors extracted and refined by Refiner (GPT-5, page 2 of 2).}
    \Description{}
    \label{fig:color11}
\end{figure*}

\begin{figure*}[t]
    \centering
    \includegraphics[width=0.9\linewidth]{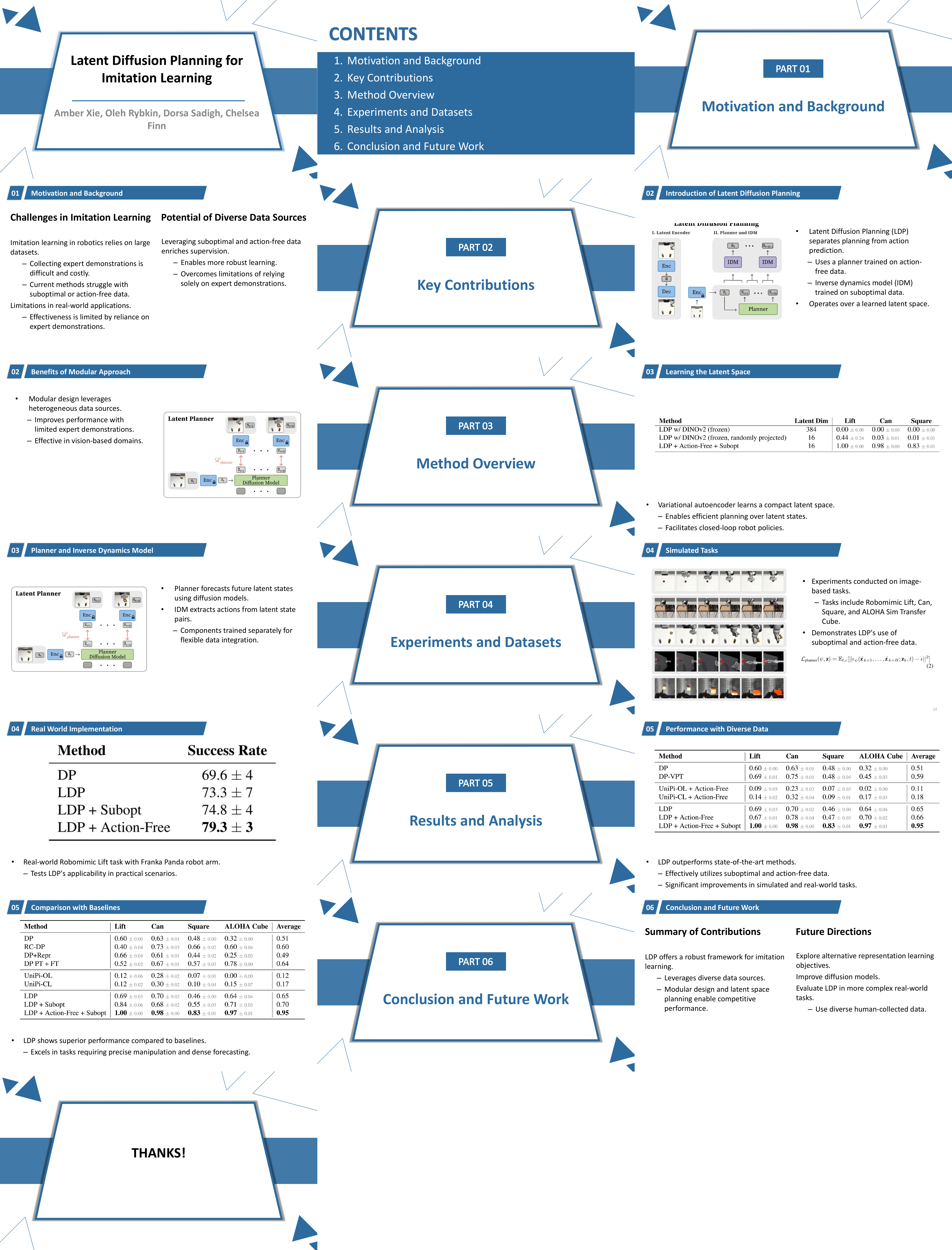}
    \caption{SlideGen output with colors extracted and refined by Refiner (GPT-4o).}
    \Description{}
    \label{fig:color2}
\end{figure*}

\begin{figure*}[t]
    \centering
    \includegraphics[width=0.9\linewidth]{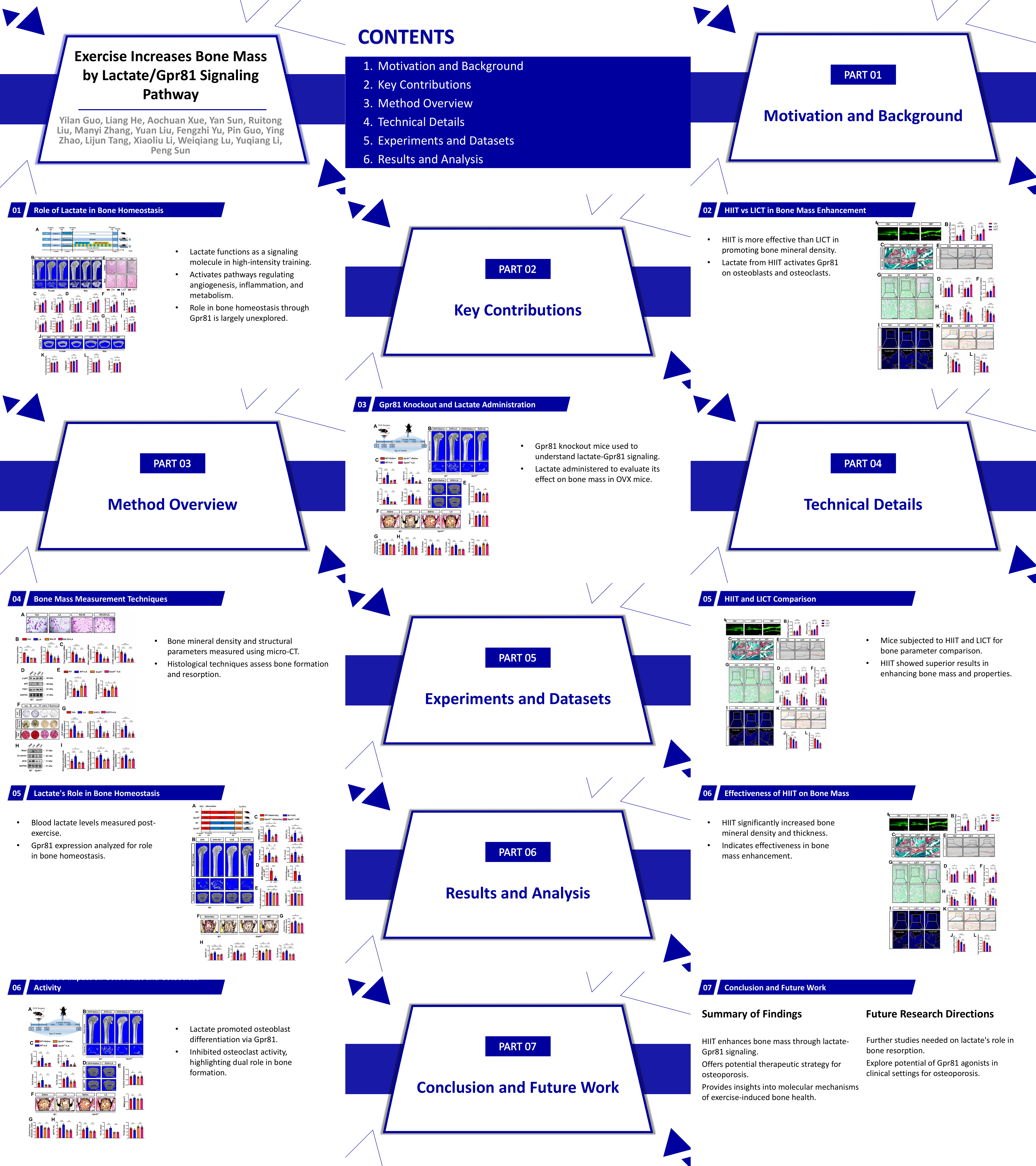}
    \caption{SlideGen output on a biomedical-domain paper~\citep{guo2025exercise}, demonstrating cross-domain generalization.}
    \Description{}
    \label{fig:medical}
\end{figure*}

We present representative slide decks generated by SlideGen to illustrate the visual diversity and content fidelity of the pipeline.
Figures~\ref{fig:example11},~\ref{fig:example12},~\ref{fig:example13},~\ref{fig:example14}, and~\ref{fig:example15} show decks produced with GPT-4o and GPT-5 using the default deep blue theme.
Figures~\ref{fig:color1},~\ref{fig:color11}, and~\ref{fig:color2} show decks where the Refiner extracts and adapts colors from the source paper, without manual color selection.
We also include a biomedical-domain paper (Figure~\ref{fig:medical}) to demonstrate cross-domain generalization beyond CS.
Finally, Figures~\ref{fig:note_a} and~\ref{fig:note_b} present a deck using the paper’s extracted color palette together with generated speaker notes.

\section{Samples Generated by Baselines}
\label{sec:baseline_sample}
\begin{figure*}[t]
    \centering
    \includegraphics[width=0.88\linewidth,page=1]{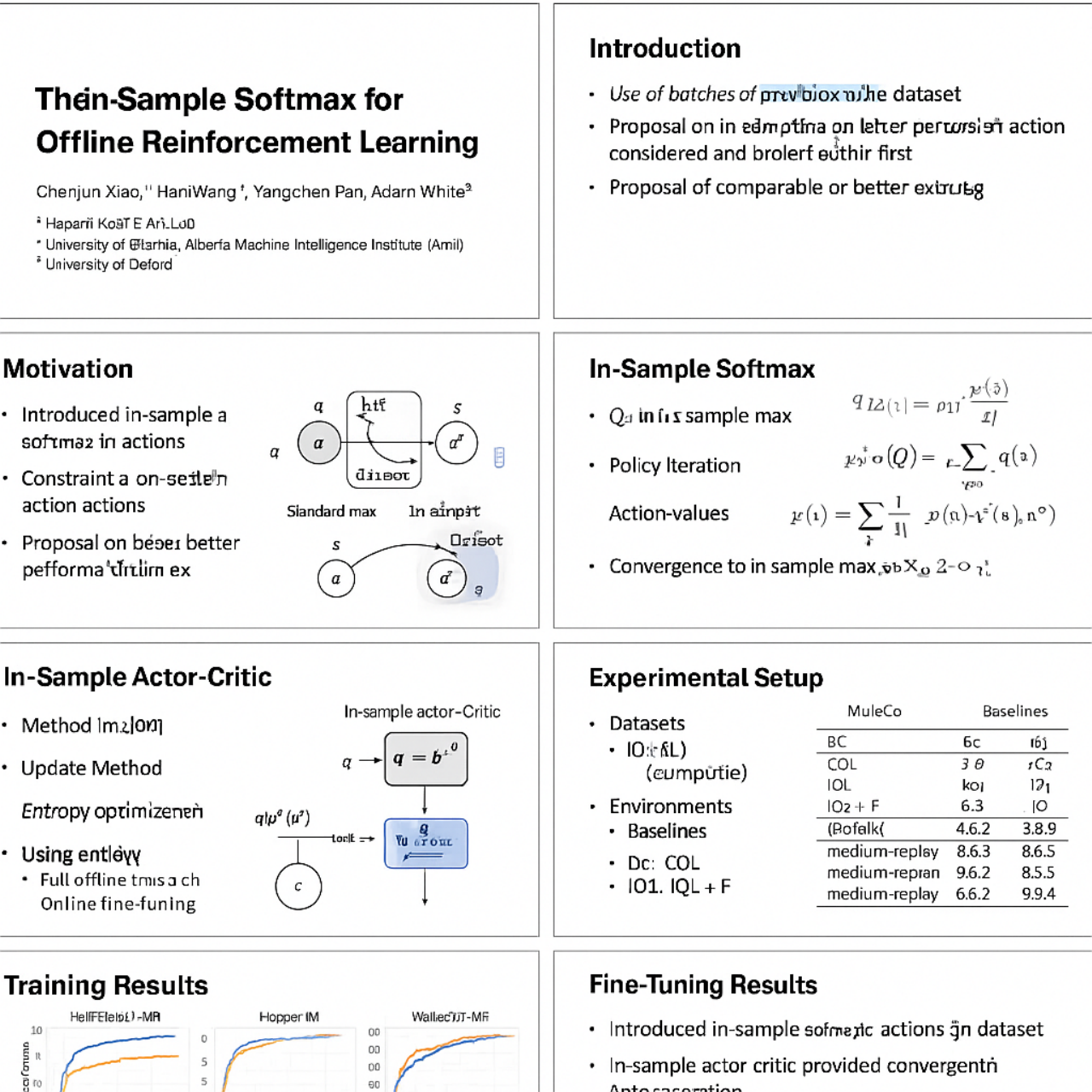}
    \caption{Baseline output generated by GPT-4o.}
    \Description{}
    \label{fig:baseline_4o_sample}
\end{figure*}

\begin{figure*}[!tb]
    \centering
    \includegraphics[width=\linewidth,page=1]{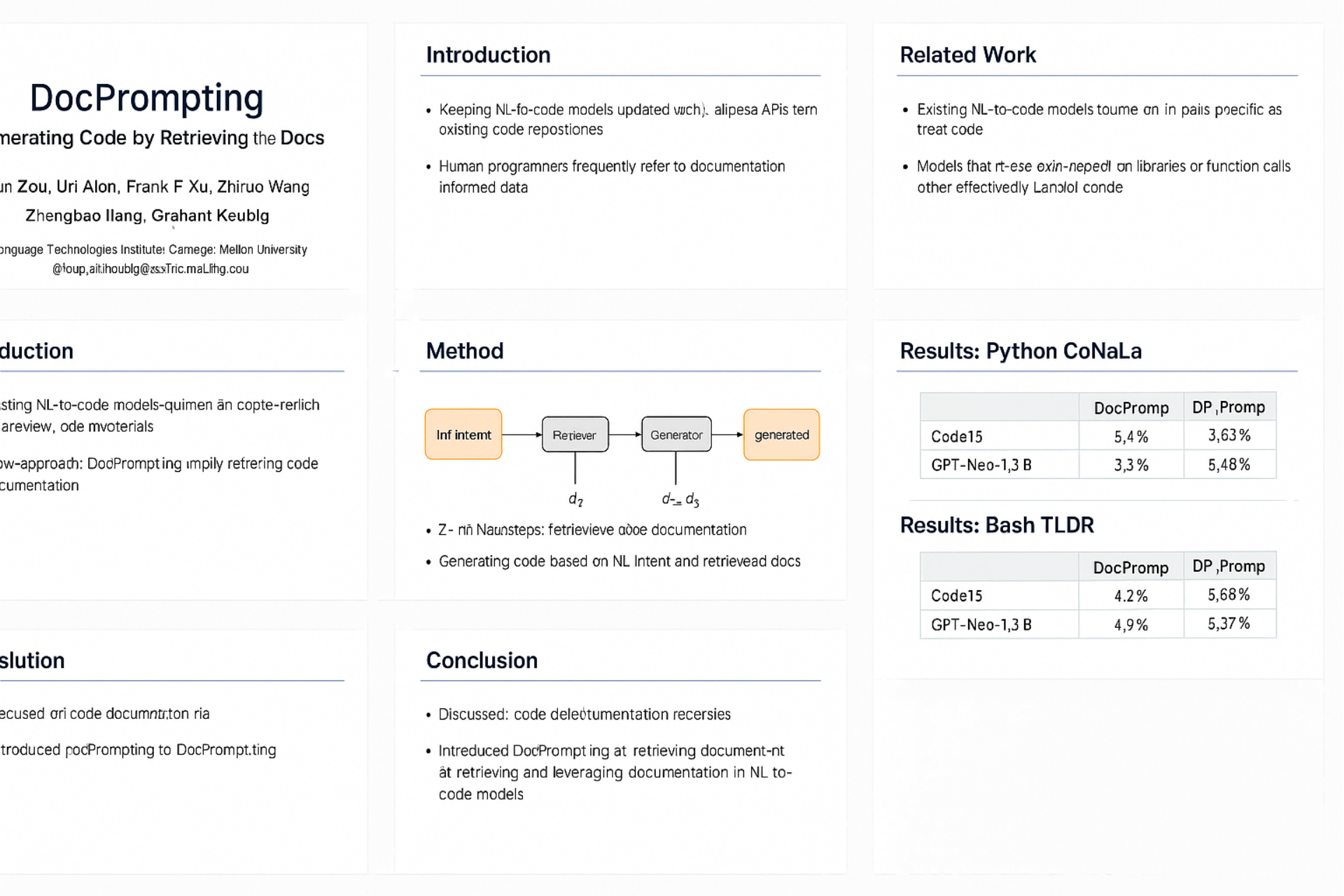}
    \caption{Baseline output generated by GPT-5.}
    \Description{}
    \label{fig:baseline_5_sample}
\end{figure*}

\begin{figure*}[t]
    \centering
    \includegraphics[width=0.86\linewidth,page=1]{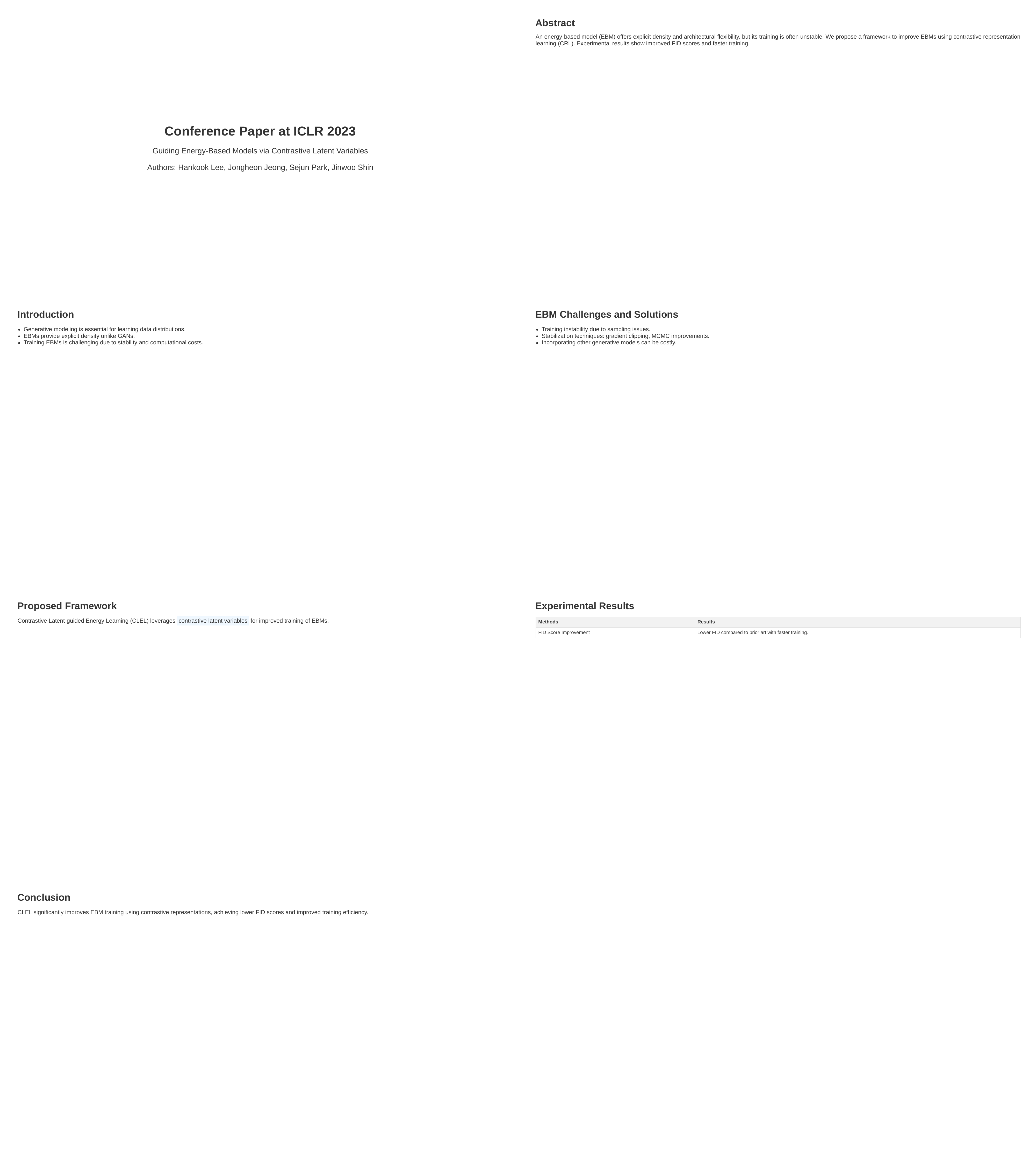}
    \caption{Baseline output generated by GPT-4o-HTML.}
    \Description{}
    \label{fig:baseline_4ohtml_sample}
\end{figure*}

\begin{figure*}[t]
    \centering
    \includegraphics[width=0.84\linewidth,page=1]{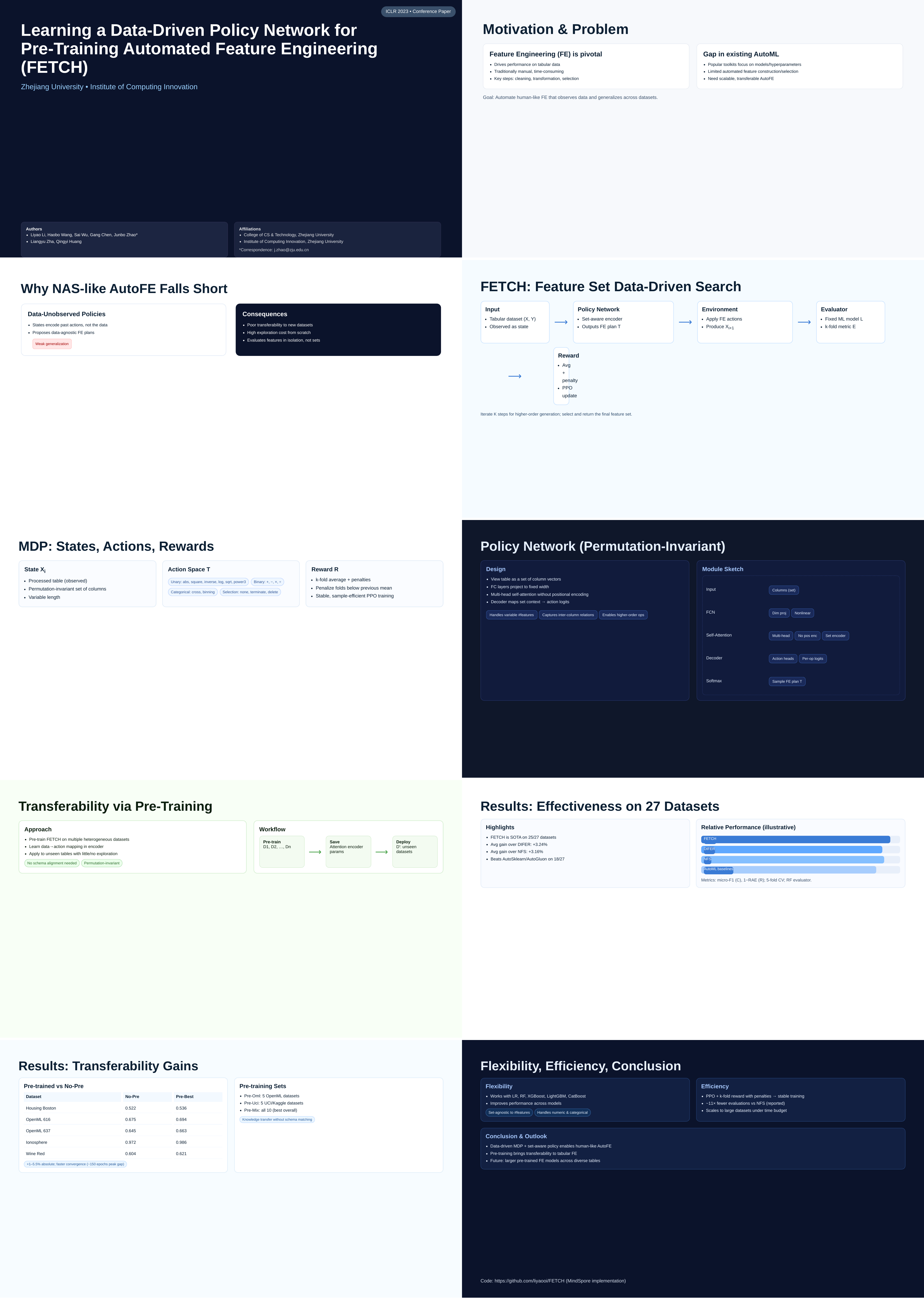}
    \caption{Baseline output generated by GPT-5-HTML.}
    \Description{}
    \label{fig:baseline_5html_sample}
\end{figure*}

For comparison, we include slide decks produced by the four baseline methods evaluated in our experiments. Each baseline takes the same input paper and generates slides without explicit layout planning or template selection. \cref{fig:baseline_4o_sample,fig:baseline_5_sample,fig:baseline_4ohtml_sample,fig:baseline_5html_sample} show the outputs of GPT-4o, GPT-5, GPT-4o-HTML, and GPT-5-HTML, respectively.

\section{Layout Library}\label{sec:Template_Library}
\begin{figure*}
\centering
\includegraphics[width=0.8\linewidth]{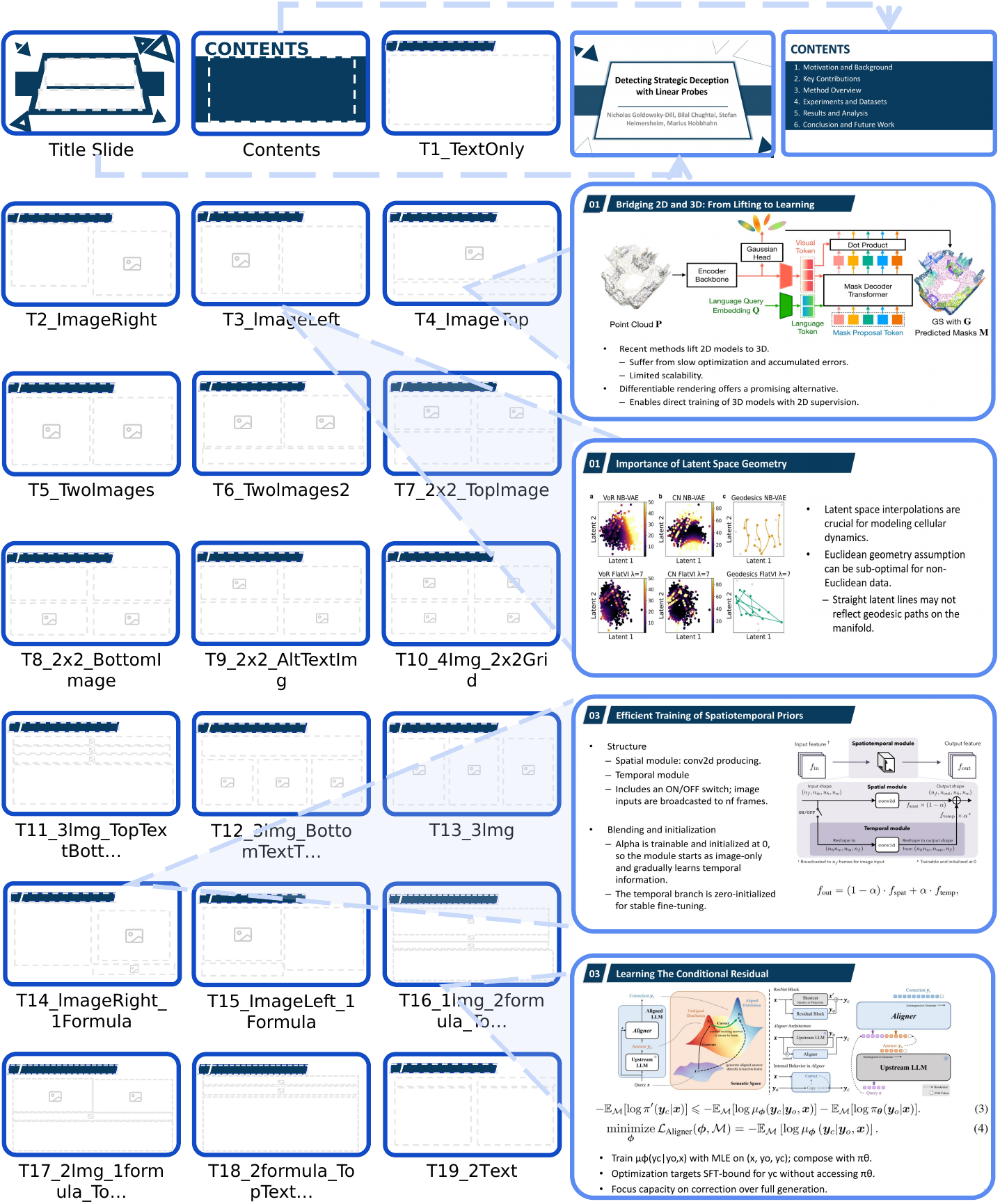}
\caption{The complete slide layout library used by Arranger. Each template addresses a typical presentation structure (e.g., text-only, image-left, two-column). Throughout the paper, we adopt 16:9 as the default deck aspect ratio, while users are free to modify the layout library’s size and aspect ratio. }
  \Description{}
\label{fig:template}
\end{figure*}

\begin{figure*}[!t]
\centering
\includegraphics[width=\linewidth]{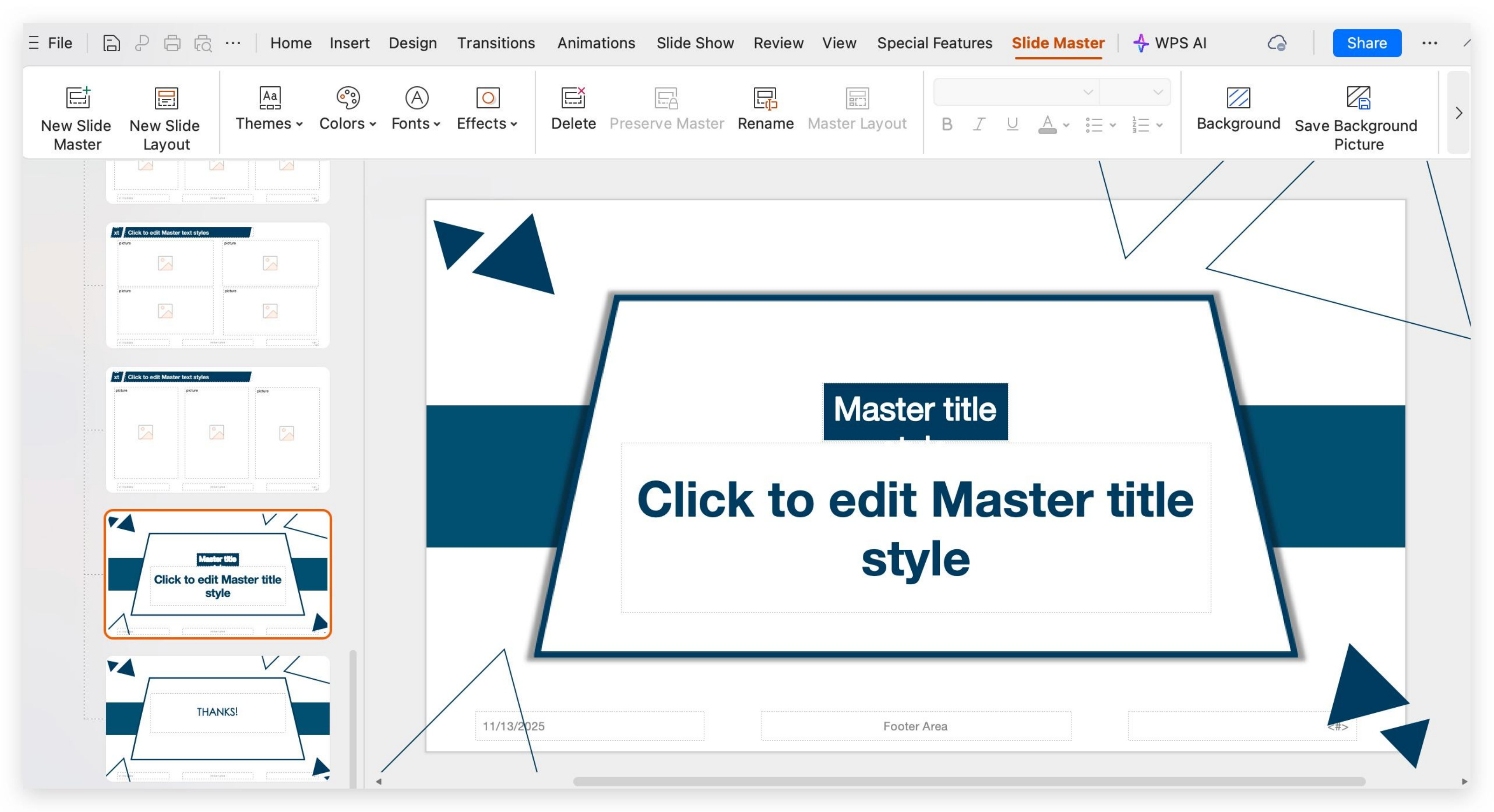}
\caption{
\textbf{Slide Master interface in WPS.}
This screenshot illustrates the Slide Master view in WPS, where users can customize templates for SlideGen by defining fonts, colors, backgrounds, logos, and other design elements. Such flexibility enables users to create personalized slide designs for various themes and purposes.
}
\Description{}
\label{fig:slide_master}
\end{figure*}

The layout library a fixed set of 19 templates covering common slide structures: text-only, single- and multi-image compositions, formula-centric layouts, and two-column merges. Figure~\ref{fig:template} shows the complete library, and Table~\ref{tab:template-rules} summarizes the selection rules that map each slide's content profile (number of images, presence of formulas, aspect ratio) to the most suitable template. Users can extend the library through the Slide Master interface (Figure~\ref{fig:slide_master}) without modifying the pipeline.
\begin{table*}[t]
\centering
\caption{Summary of slide templates and their selection rules used by Arranger.}
\setlength{\tabcolsep}{6pt}
\renewcommand{\arraystretch}{1.3} 
\begin{tabular}{
  >{\raggedright\arraybackslash}p{0.32\textwidth}
  >{\raggedright\arraybackslash}p{0.62\textwidth}
}
\toprule
\textbf{Template} & \textbf{Selection Rule} \\
\midrule

\rowcolor{blue!6}
\textbf{T1\_TextOnly} 
& Use when the subsection contains no images or tables; suitable for pure-text slides or conceptual summaries. \\

\textbf{T2\_ImageRight} 
& One tall/square image (aspect $\leq$ 1.0) + up to four bullets. Image placed on the right, text on the left. \\

\rowcolor{blue!6}
\textbf{T3\_ImageLeft} 
& Mirror of T2. One tall/square image placed on the left, text on the right. \\

\textbf{T4\_ImageTop} 
& One wide image/table (aspect $\geq$ 1.6) or any visual that spans nearly the slide width; image on top, text below. \\

\rowcolor{blue!6}
\textbf{T5\_TwoImages} 
& Exactly two images side-by-side, with no text. Suitable for visual comparison. \\

\textbf{T5\_TwoImages2} 
& Two side-by-side images on top, with a text block underneath. \\

\rowcolor{blue!6}
\textbf{T7\_2x2\_TopImage} 
& 2×2 layout: top two blocks are images, bottom two are text. For two visuals + explanatory bullets. \\

\textbf{T8\_2x2\_BottomImage} 
& 2×2 layout: top two blocks are text, bottom two are images. Use when the narrative is text-heavy. \\

\rowcolor{blue!6}
\textbf{T9\_2x2\_AltTextImg} 
& Alternating 2×2 layout (top-left \& bottom-right = images; top-right \& bottom-left = text). \\

\textbf{T10\_4Img\_2x2Grid} 
& Exactly four images in a 2×2 grid, with no text. Best for dataset galleries or qualitative comparisons. \\

\rowcolor{blue!6}
\textbf{T11\_3Img\_TopTextBottom} 
& Three images in one row across the top with a text block below. \\

\textbf{T12\_3Img\_BottomTextTop} 
& Text block on top followed by three images in a row. \\

\rowcolor{blue!6}
\textbf{T13\_3Img} 
& Title on top with three evenly spaced images below. \\

\textbf{T14\_ImageRight\_1Formula} 
& Right column contains one image/table (top) + one formula (bottom); text on the left. For paired visual + equation slides. \\

\rowcolor{blue!6}
\textbf{T15\_ImageLeft\_1Formula} 
& Mirror of T14: image/table + formula on the left column; text on the right. \\

\textbf{T16\_1Img\_2formula\_TopTextBottom} 
& Top rows contain one image/table + two formulas; text block at the bottom. \\

\rowcolor{blue!6}
\textbf{T17\_2Img\_1formula\_TopTextBottom} 
& Two images on the top row, one formula in the middle, text at the bottom. \\

\textbf{T18\_2formula\_TopTextBottom} 
& Two formulas stacked at the top with a text block below. Use when both formulas belong on the same slide. \\

\rowcolor{blue!6}
\textbf{T19\_TextMerge} 
&  Merge of two T1 slides into a single two-column layout: one T1 forms the left column, the other forms the right column. \\
\bottomrule
\end{tabular}
\label{tab:template-rules}
\end{table*}

\section{Prompts}
\label{prompts}
\begin{figure*}[!tb]
    \centering
    \includegraphics[width=0.83\linewidth,page=1]{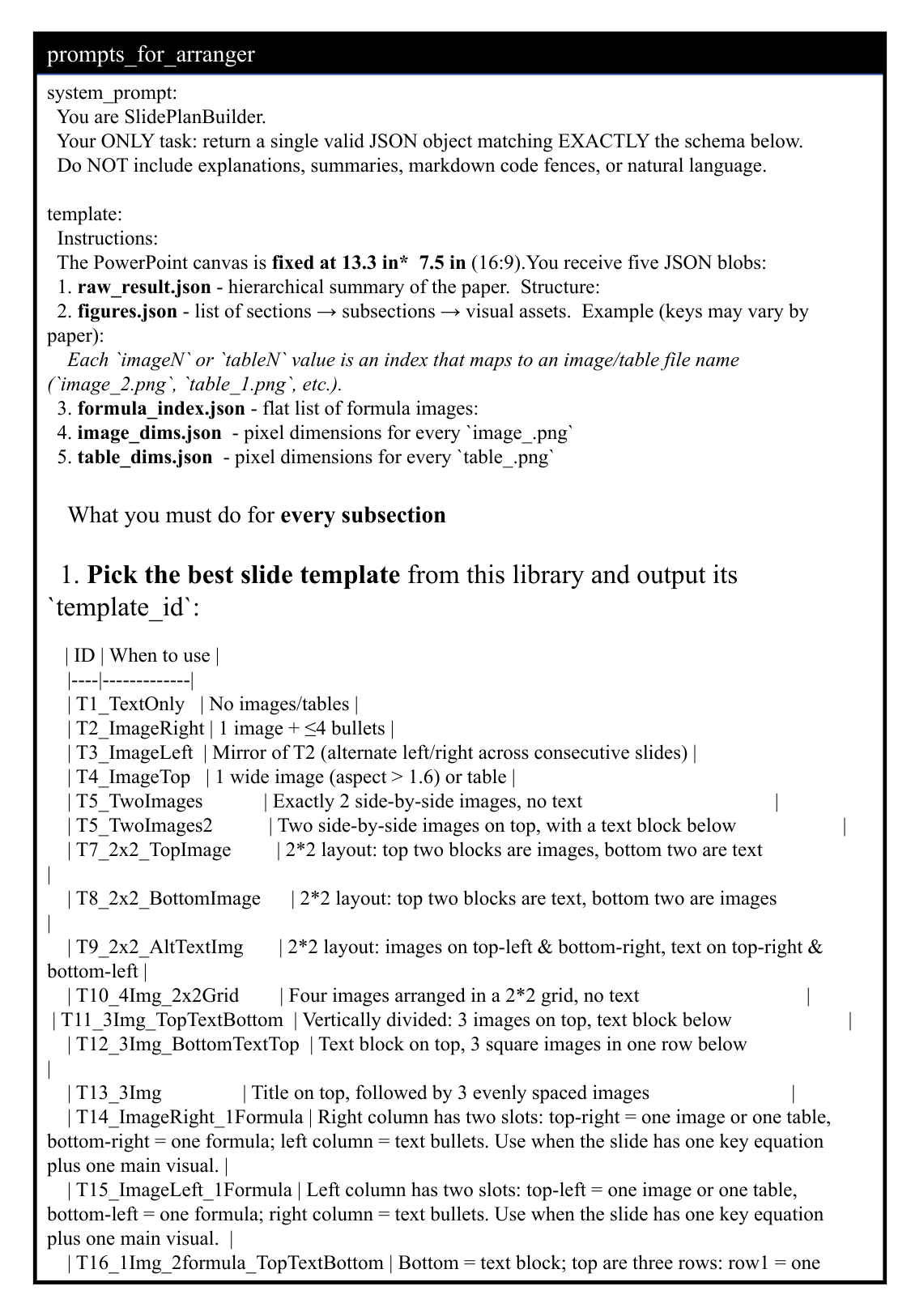}
    \caption{Prompt for Arranger.}
    \Description{}
    \label{fig:layout1}
\end{figure*}

\begin{figure*}[!tb]
    \centering
    \includegraphics[width=0.83\linewidth,page=2]{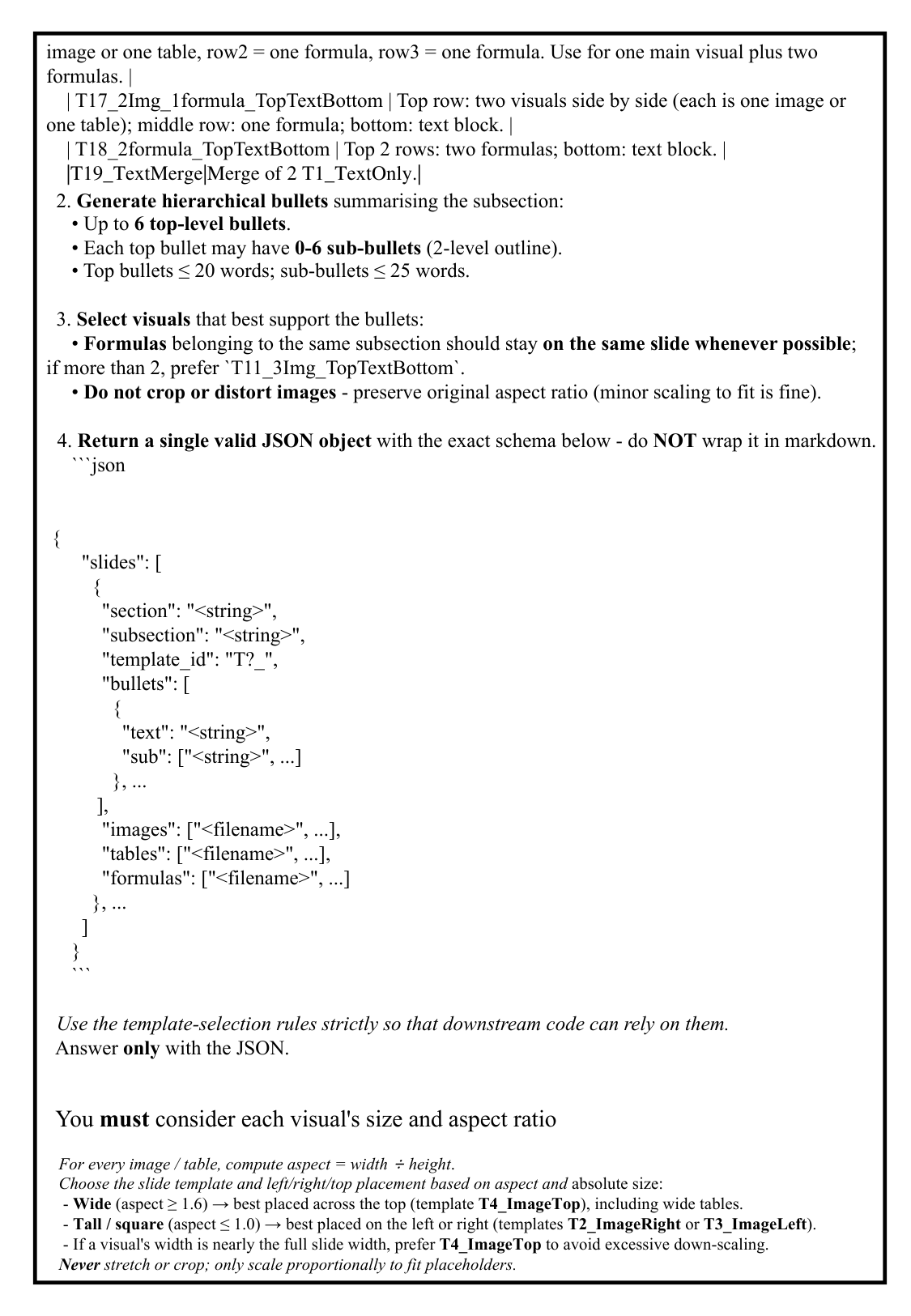}
    \caption{Prompt for Arranger.}
    \Description{}
    \label{fig:layout2}
\end{figure*}

\begin{figure*}[!tb]
    \centering
    \includegraphics[width=0.83\linewidth,page=3]{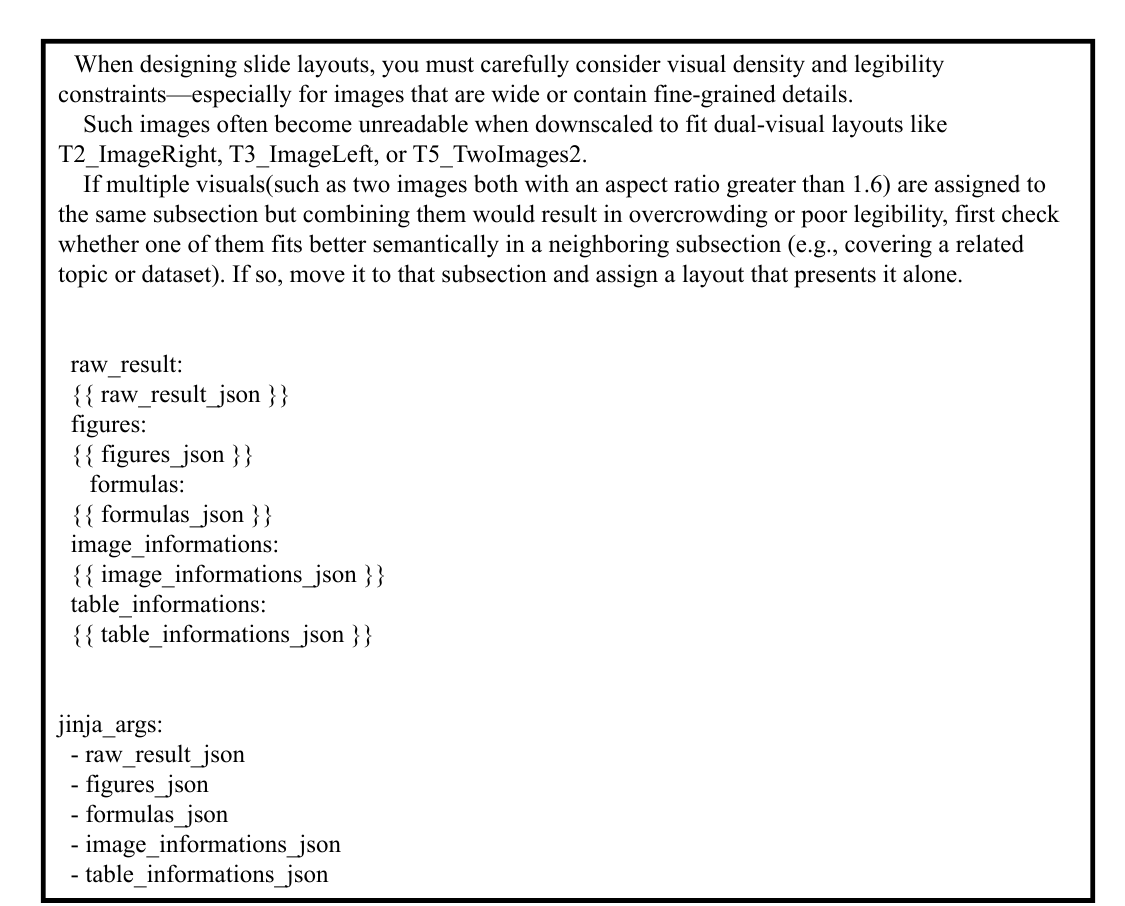}
    \caption{Prompt for Arranger.}
    \Description{}
    \label{fig:layout3}
\end{figure*}

\begin{figure*}[!tb]
    \centering
    \includegraphics[width=0.83\linewidth,page=4]{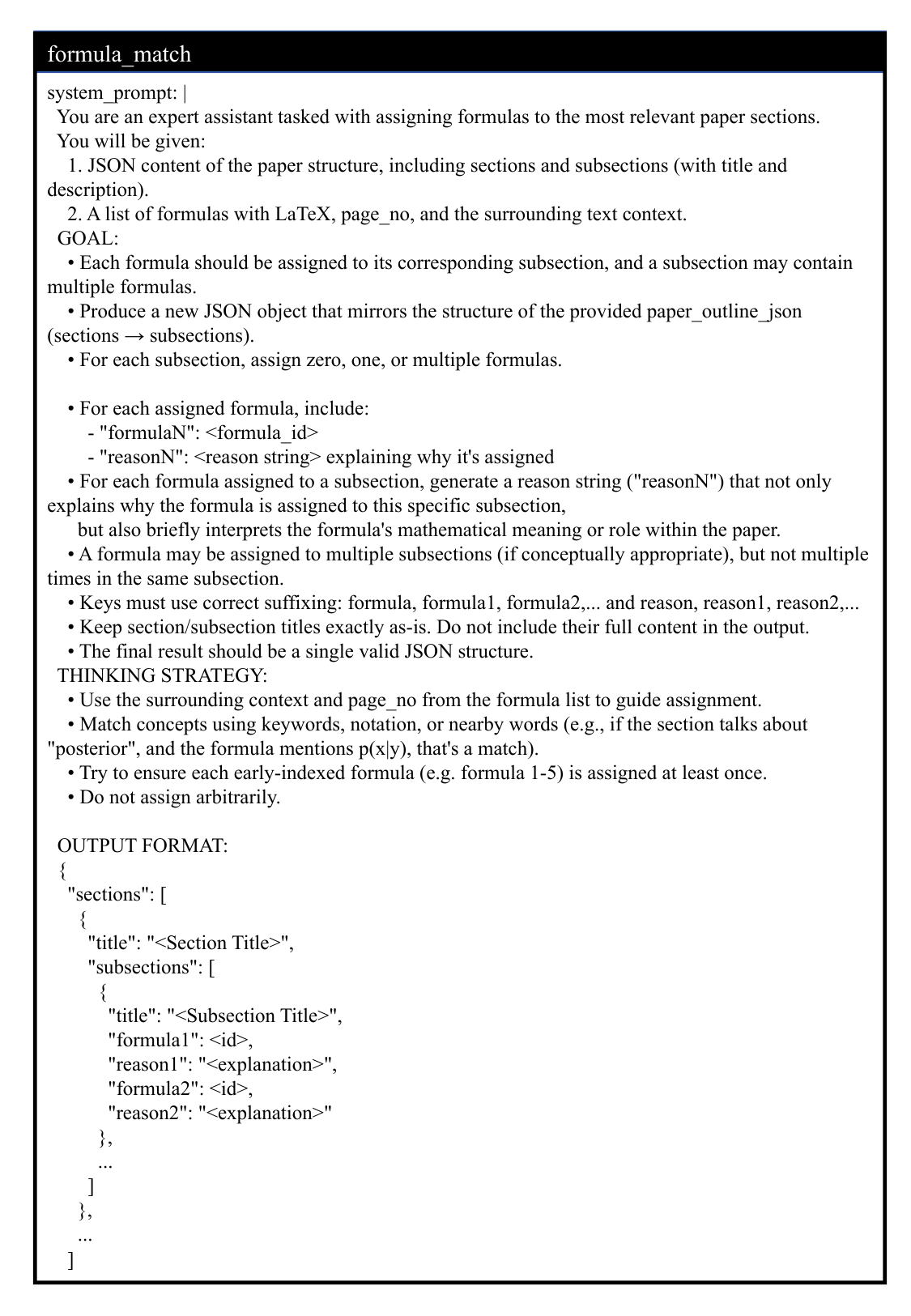}
    \caption{Prompt for Formulizer.}
    \Description{}
    \label{fig:layout4}
\end{figure*}

\begin{figure*}[!tb]
    \centering
    \includegraphics[width=0.83\linewidth,page=5]{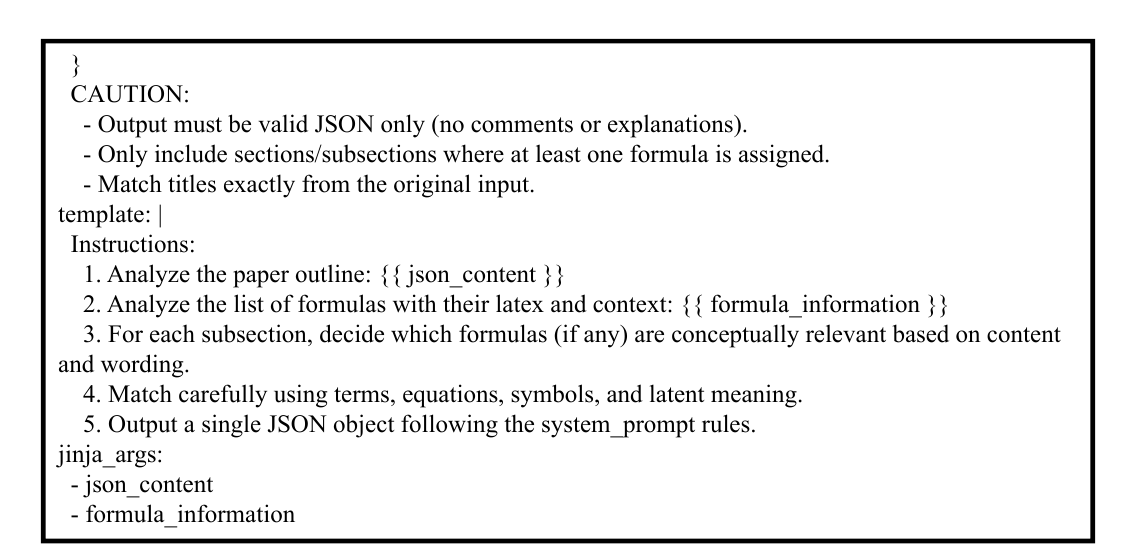}
    \caption{Prompt for Formulizer.}
    \Description{}
    \label{fig:layout5}
\end{figure*}

\begin{figure*}[!tb]
    \centering
    \includegraphics[width=0.83\linewidth,page=9]{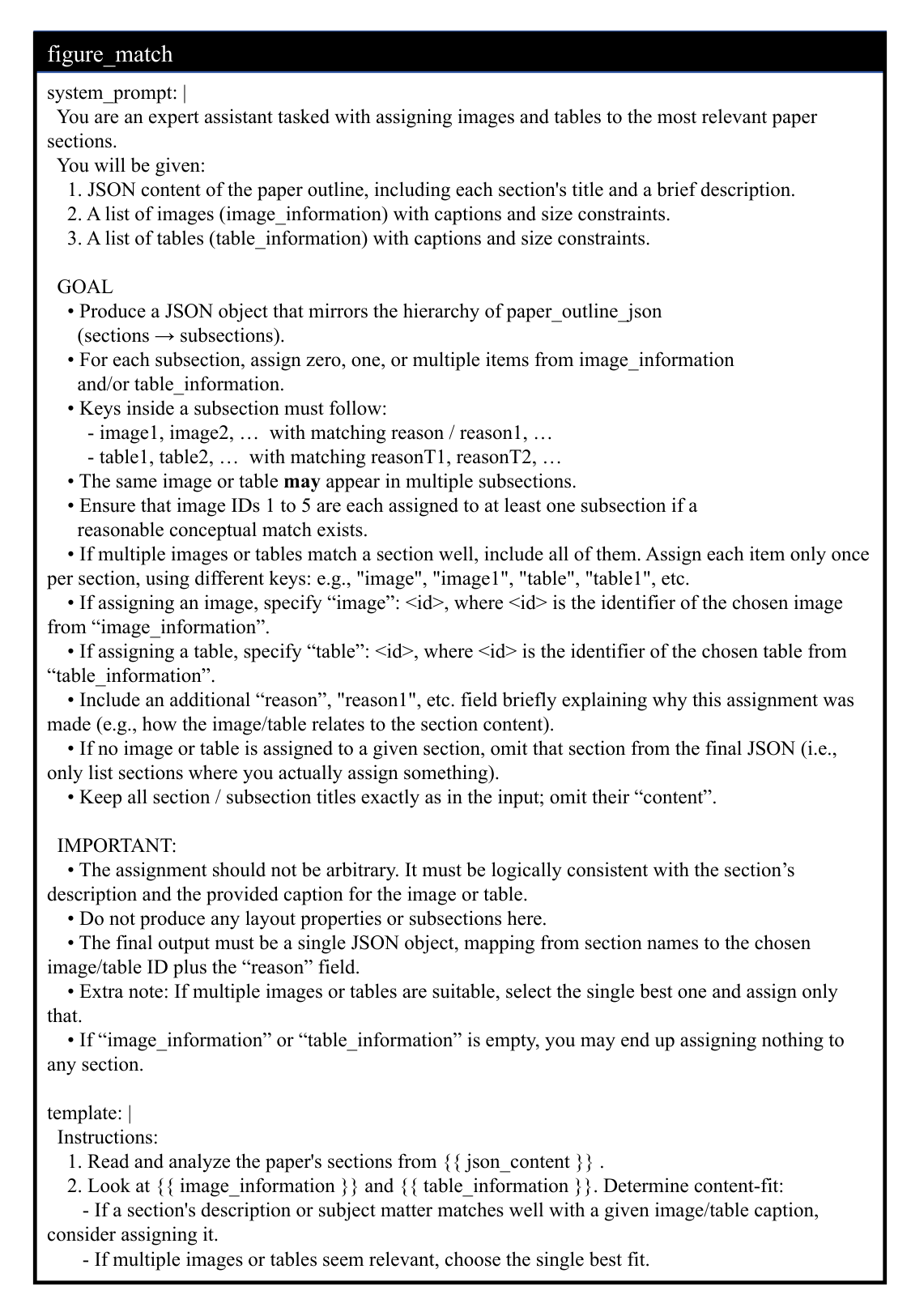}
    \caption{Prompt for Mapper.}
    \Description{}
    \label{fig:layout9}
\end{figure*}

\begin{figure*}[!tb]
    \centering
    \includegraphics[width=0.83\linewidth,page=10]{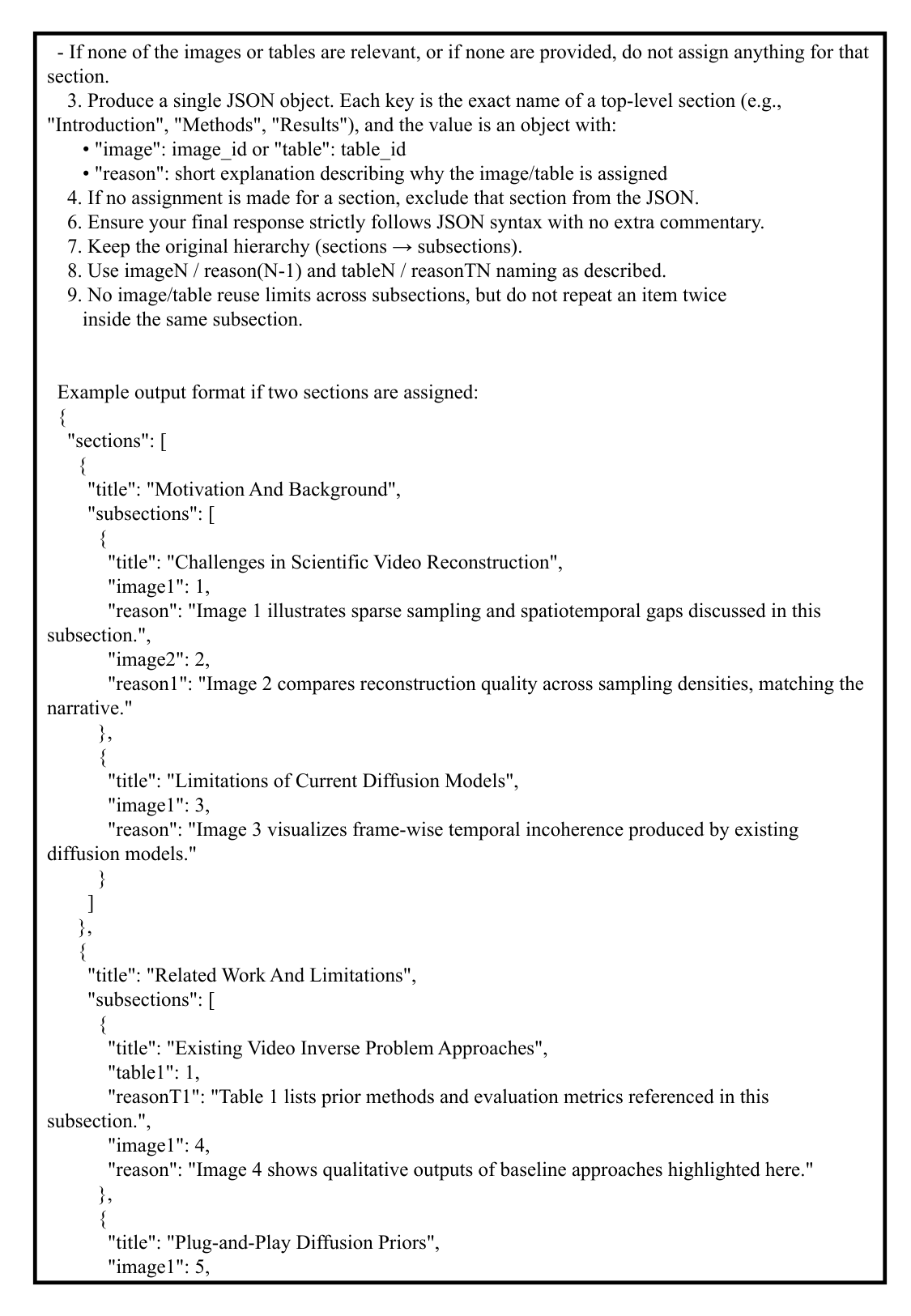}
    \caption{Prompt for Mapper.}
    \Description{}
    \label{fig:layout10}
\end{figure*}

\begin{figure*}[!tb]
    \centering
    \includegraphics[width=0.83\linewidth,page=11]{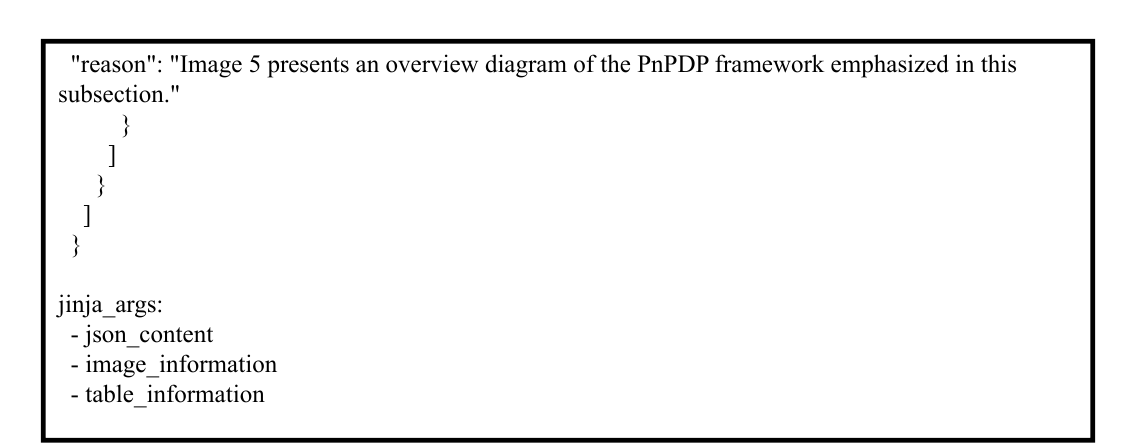}
    \caption{Prompt for Mapper.}
    \Description{}
    \label{fig:layout11}
\end{figure*}

\begin{figure*}[!tb]
    \centering
    \includegraphics[width=0.83\linewidth,page=34]{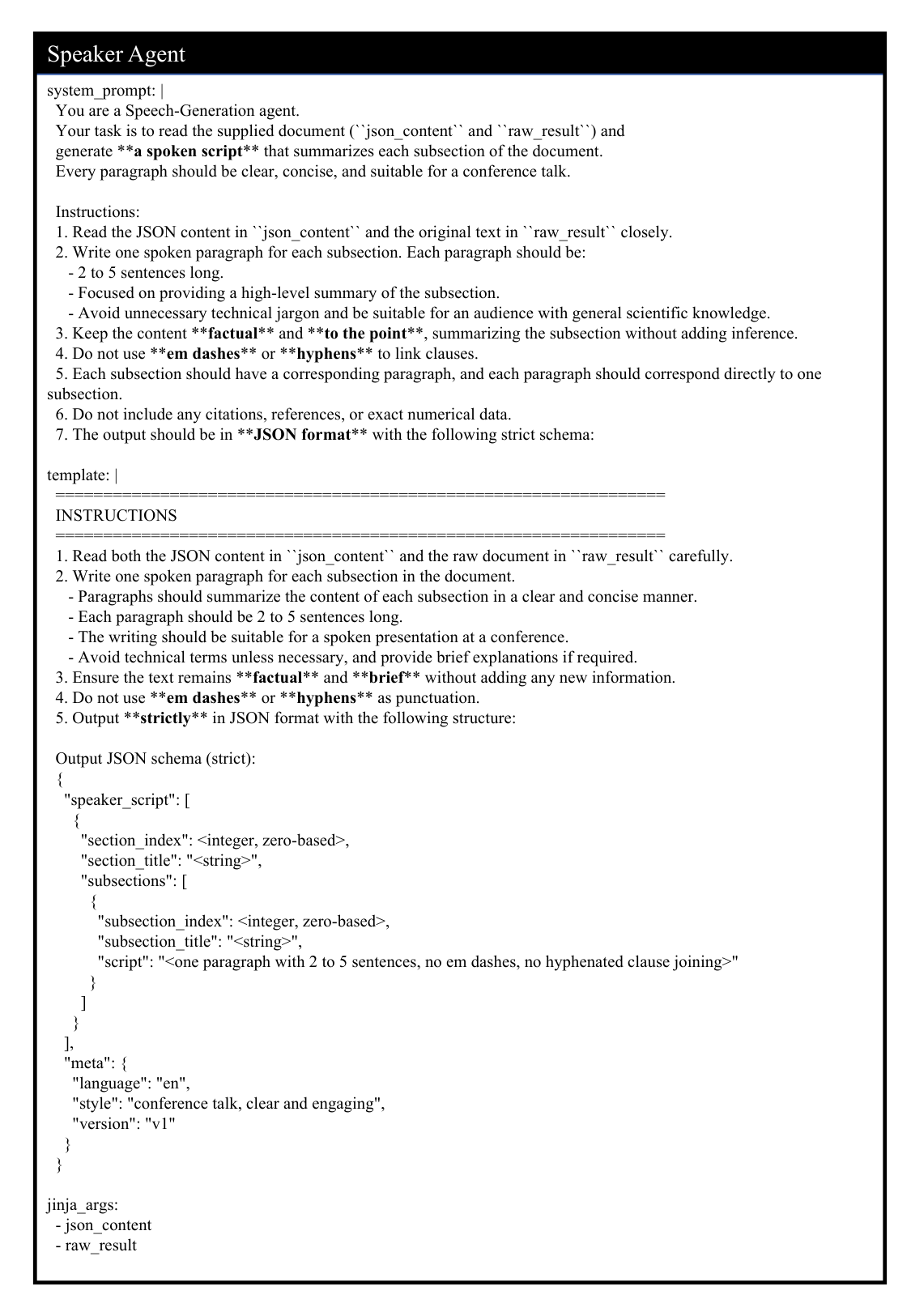}
    \caption{Prompt for Speaker.}
    \Description{}
    \label{fig:layout12}
\end{figure*}

\begin{figure*}[!tb]
    \centering
    \includegraphics[width=0.83\linewidth,page=1]{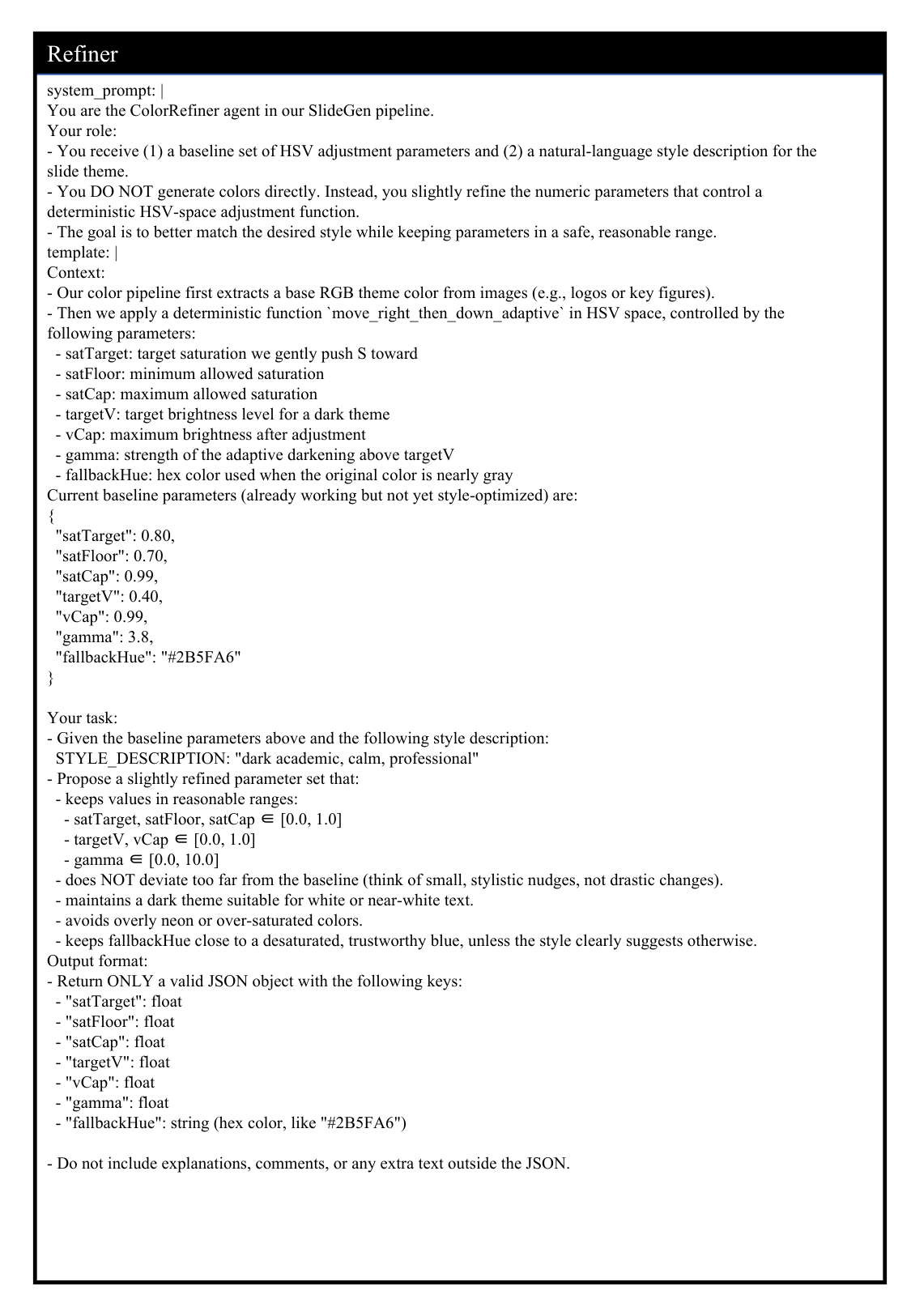}
    \caption{Prompt for Refiner.}
    \Description{}
    \label{fig:layout13}
\end{figure*}

We include the full prompts used by each agent for reproducibility. Each prompt specifies the task objective, the expected input/output format (typically a JSON schema), and any constraints or few-shot examples provided to the model. Figures~\ref{fig:layout1},~\ref{fig:layout2}, and~\ref{fig:layout3} show the Arranger prompt; Figures~\ref{fig:layout4} and~\ref{fig:layout5} show the Formulizer prompt; Figures~\ref{fig:layout9},~\ref{fig:layout10}, and~\ref{fig:layout11} show the Mapper prompt; Figure~\ref{fig:layout12} shows the Speaker prompt; and Figure~\ref{fig:layout13} shows the Refiner prompt.


\section{Example Output from Agents}
\label{sec:all-example}
Each agent produces a structured JSON object that is consumed by the next stage in the pipeline. We include representative outputs below to illustrate the intermediate representations and to help readers understand what each agent contributes.

\begin{figure*}[!tb]
    \centering
    \includegraphics[width=0.83\linewidth,page=1]{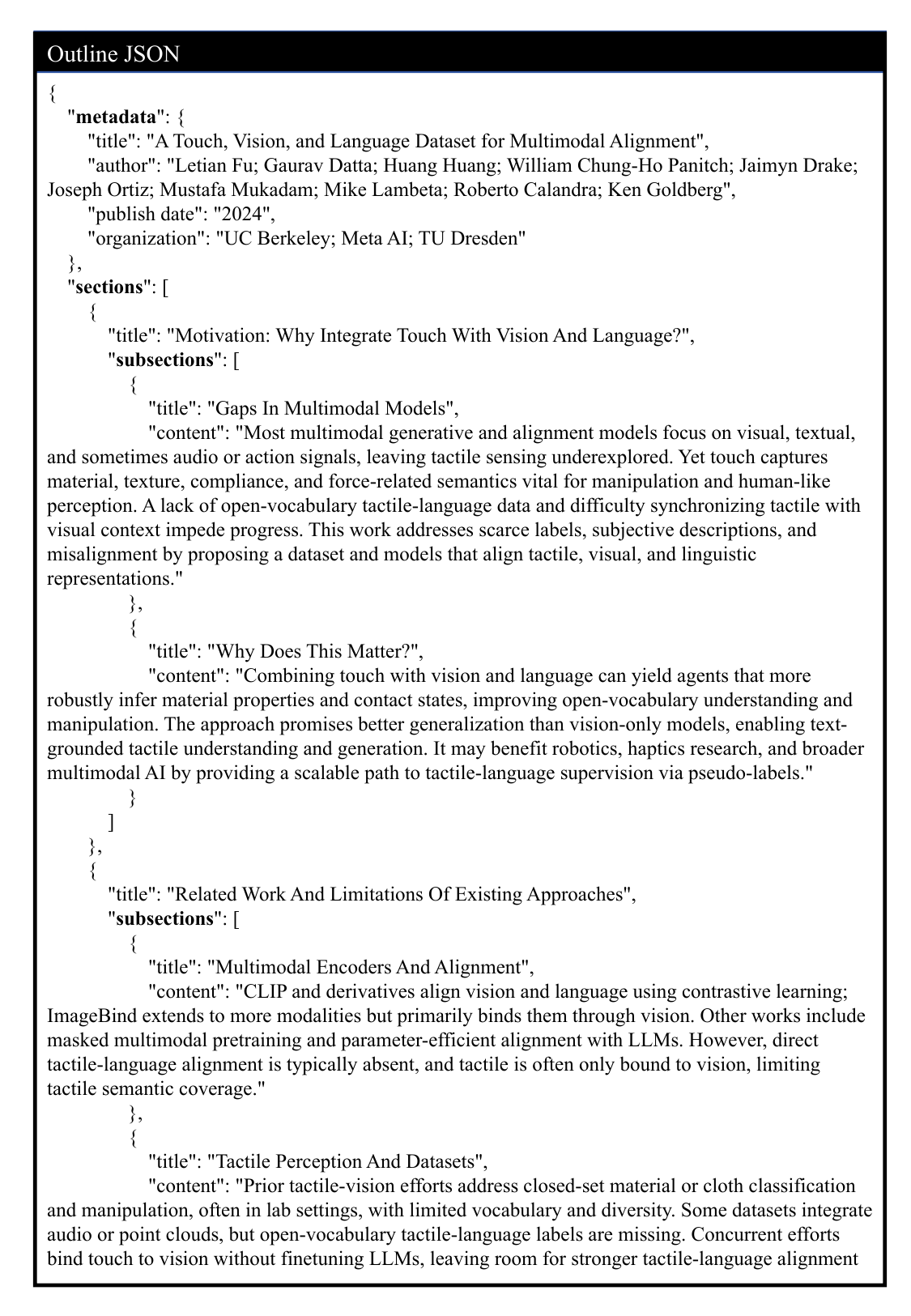}
    \caption{Output JSON of Outliner Agent.}
    \Description{}
    \label{fig:Outliner1}
\end{figure*}

\begin{figure*}[!tb]
    \centering
    \includegraphics[width=0.83\linewidth,page=2]{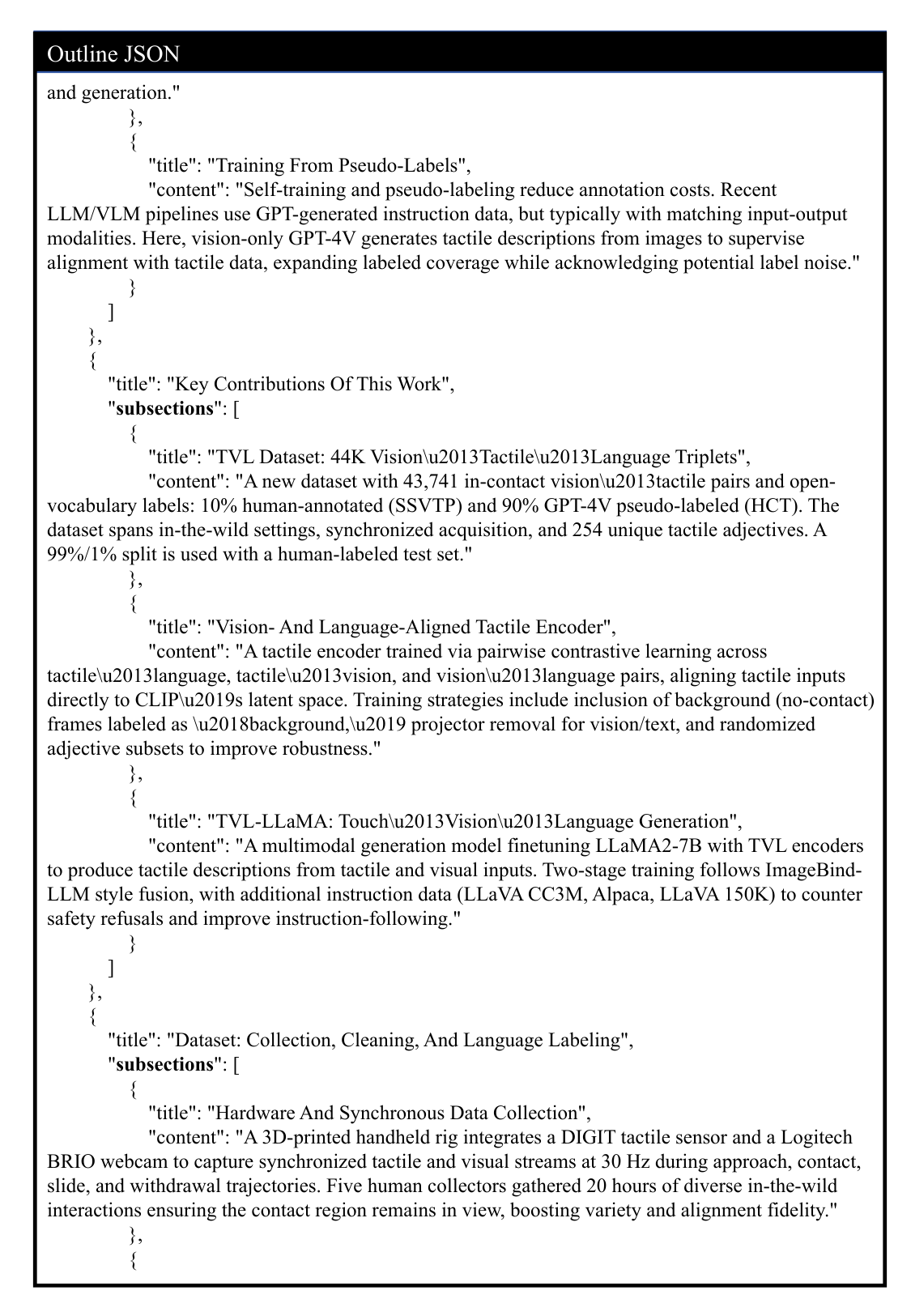}
    \caption{Output JSON of Outliner Agent.}
    \Description{}
    \label{fig:Outliner2}
\end{figure*}

\begin{figure*}[!tb]
    \centering
    \includegraphics[width=0.83\linewidth,page=3]{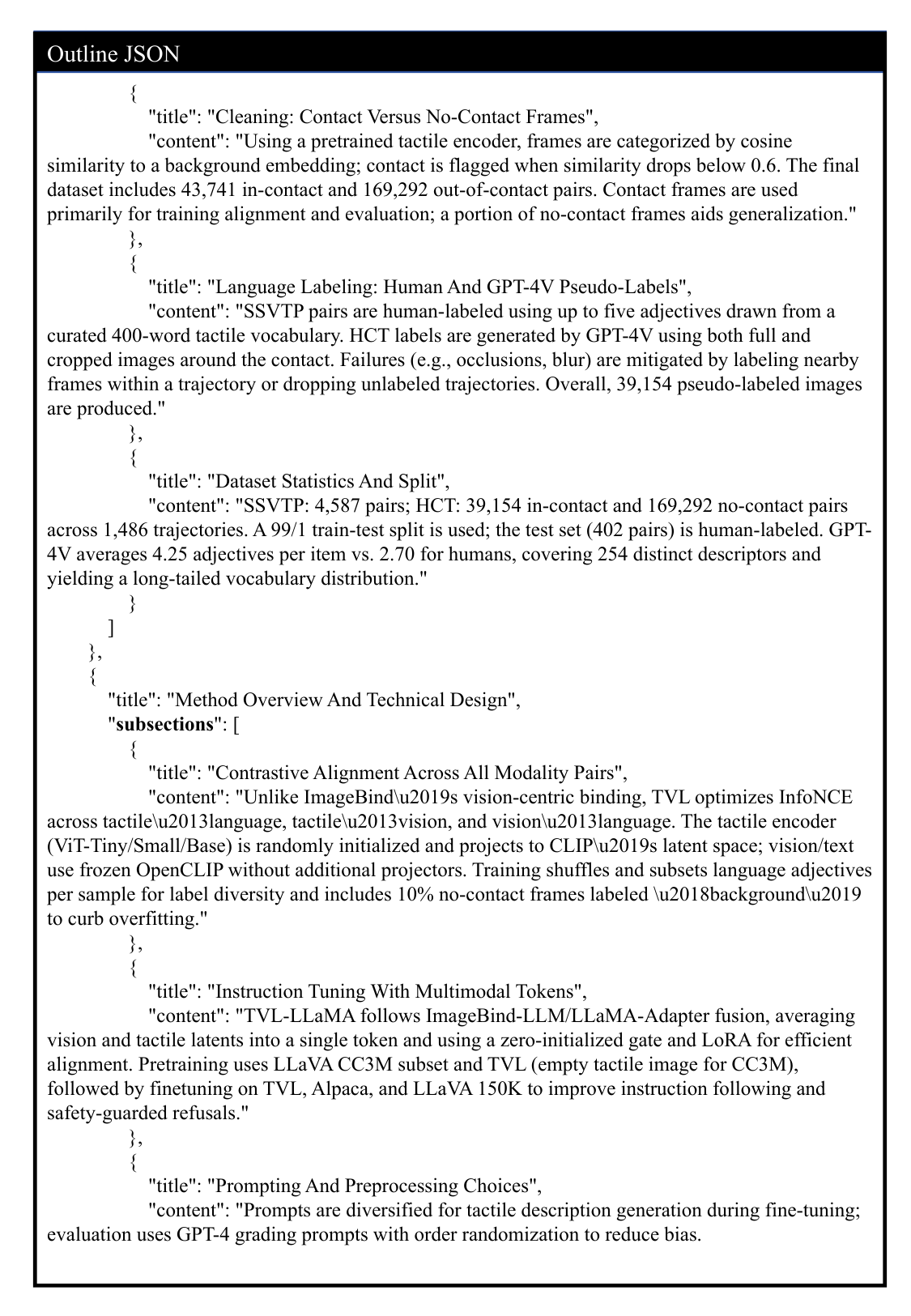}
    \caption{Output JSON of Outliner Agent.}
    \Description{}
    \label{fig:Outliner3}
\end{figure*}

\begin{figure*}[!tb]
    \centering
    \includegraphics[width=0.83\linewidth,page=4]{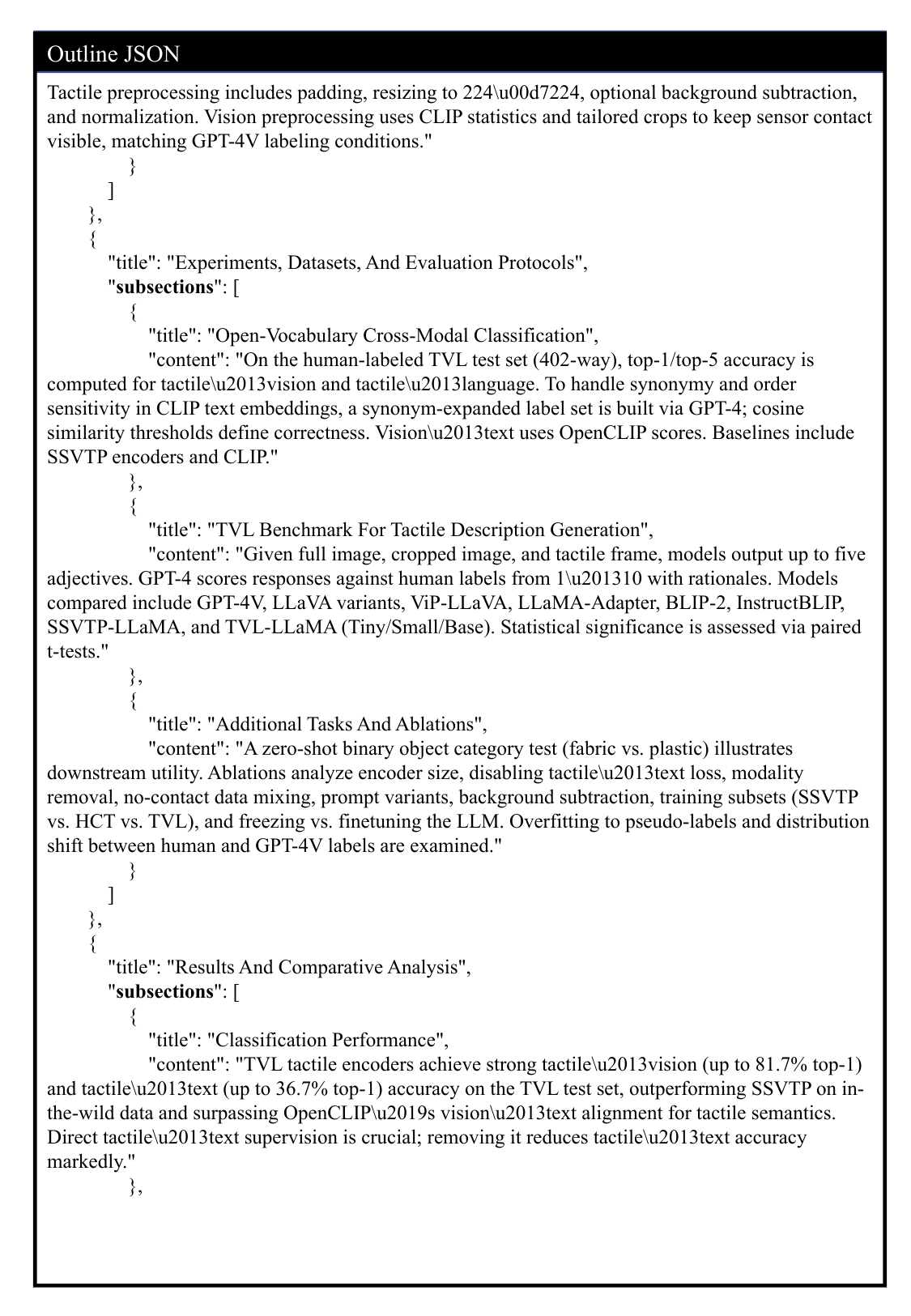}
    \caption{Output JSON of Outliner Agent.}
    \Description{}
    \label{fig:Outliner4}
\end{figure*}

\begin{figure*}[!tb]
    \centering
    \includegraphics[width=0.83\linewidth,page=5]{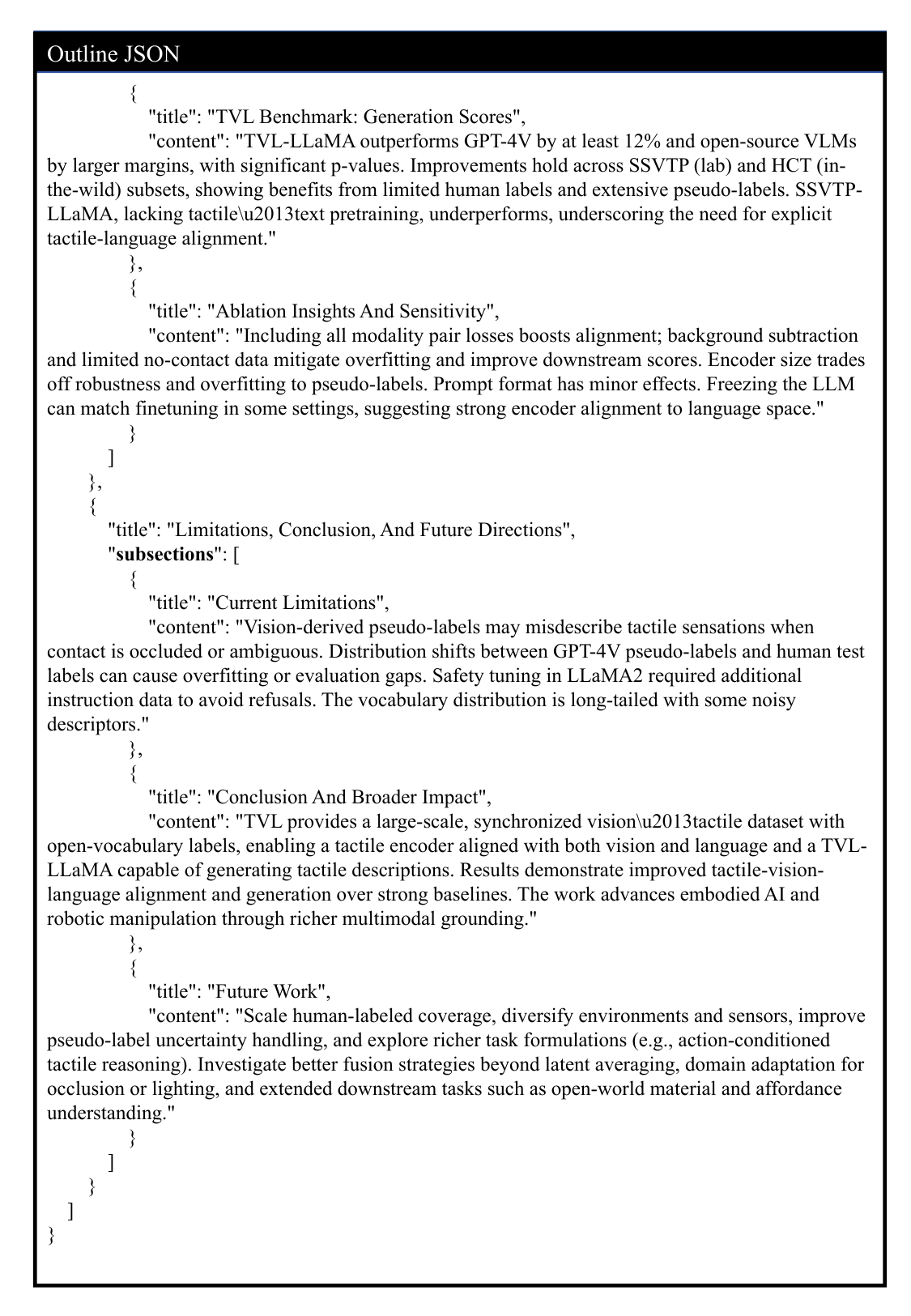}
    \caption{Output JSON of Outliner Agent.}
    \Description{}
    \label{fig:Outliner5}
\end{figure*}

\myparagraph{Outliner Output.}
\label{sec:outliner-example}
Outliner defines the slide-level narrative structure, including section groupings, slide titles, and bullet-point summaries for each slide. \cref{fig:Outliner1,fig:Outliner2,fig:Outliner3,fig:Outliner4,fig:Outliner5} show a complete example.

\begin{figure*}[!tb]
    \centering
    \includegraphics[width=0.83\linewidth,page=1]{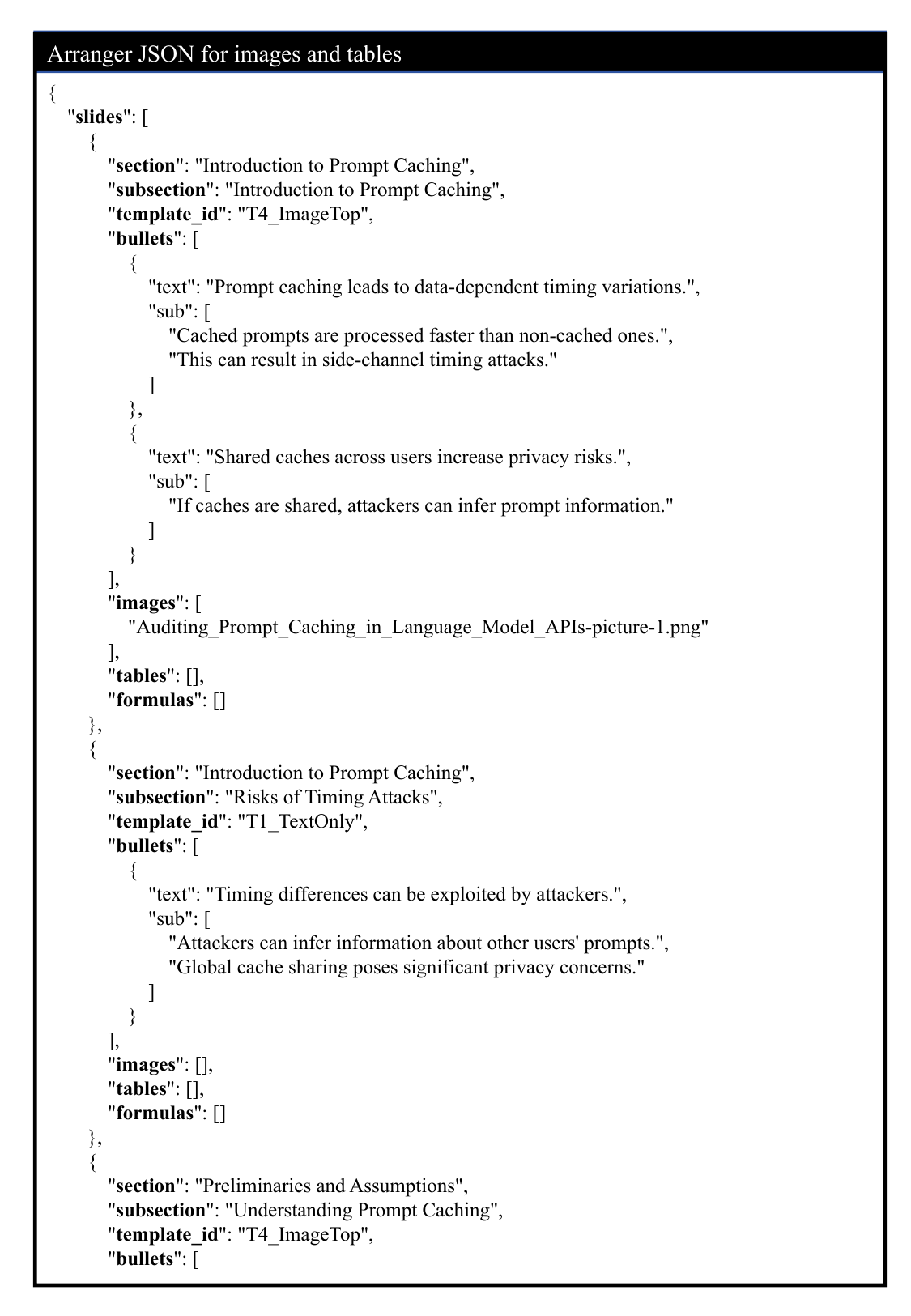}
    \caption{Output JSON of Arranger Agent.}
    \Description{}
    \label{fig:Arranger1}
\end{figure*}

\begin{figure*}[!tb]
    \centering
    \includegraphics[width=0.83\linewidth,page=2]{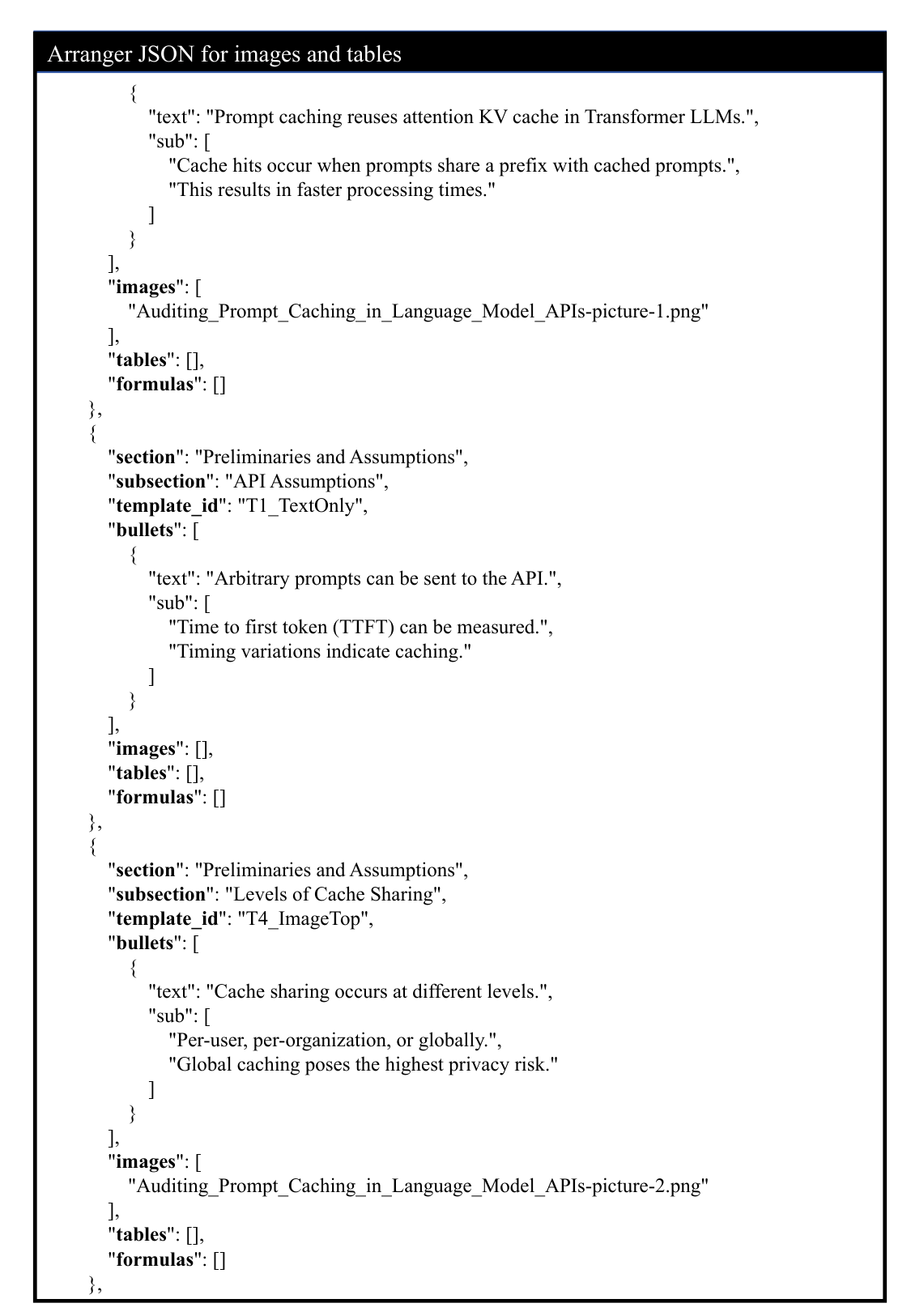}
    \caption{Output JSON of Arranger Agent.}
    \Description{}
    \label{fig:Arranger2}
\end{figure*}

\begin{figure*}[!tb]
    \centering
    \includegraphics[width=0.83\linewidth,page=3]{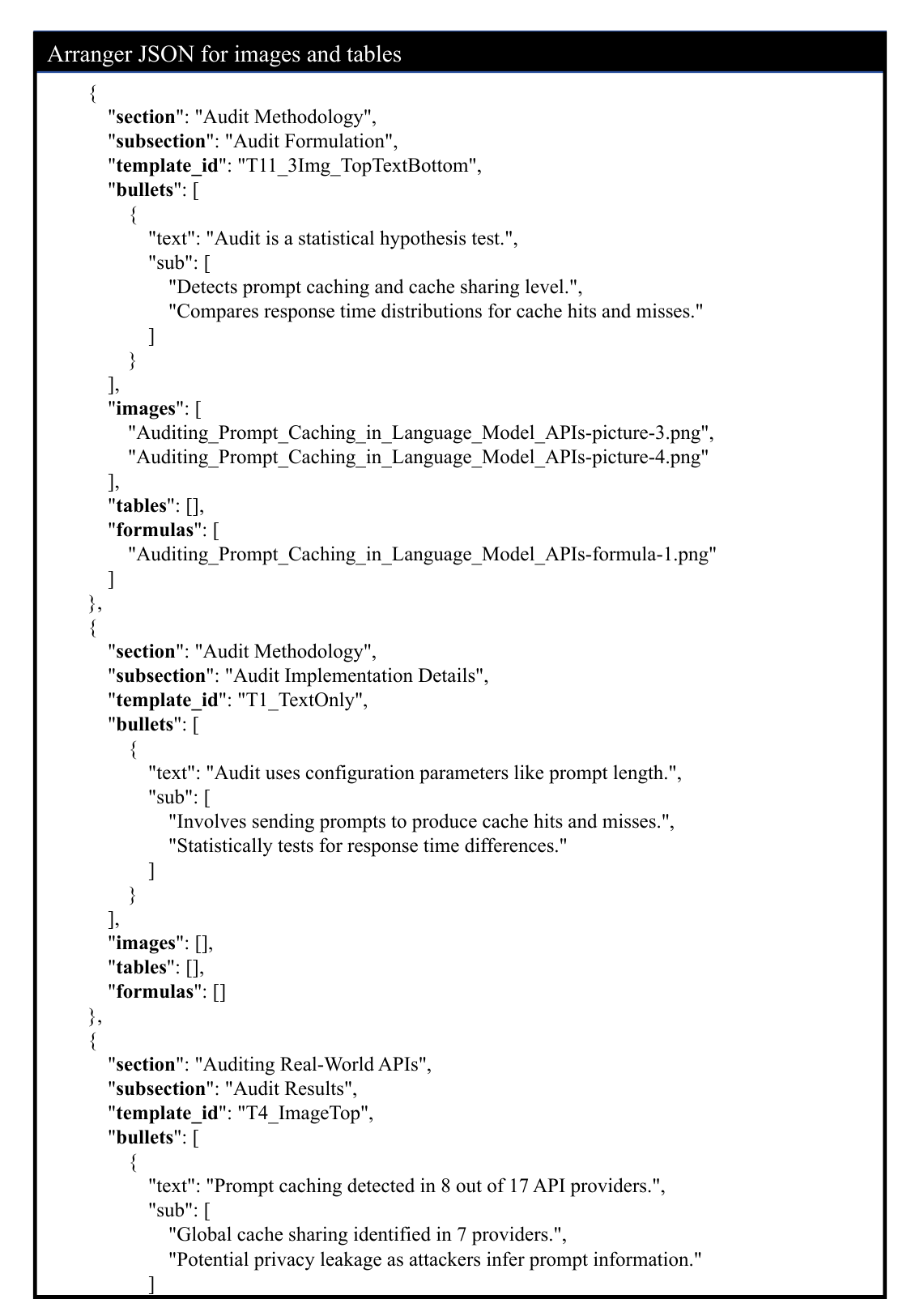}
    \caption{Output JSON of Arranger Agent.}
    \Description{}
    \label{fig:Arranger3}
\end{figure*}

\begin{figure*}[!tb]
    \centering
    \includegraphics[width=0.83\linewidth,page=4]{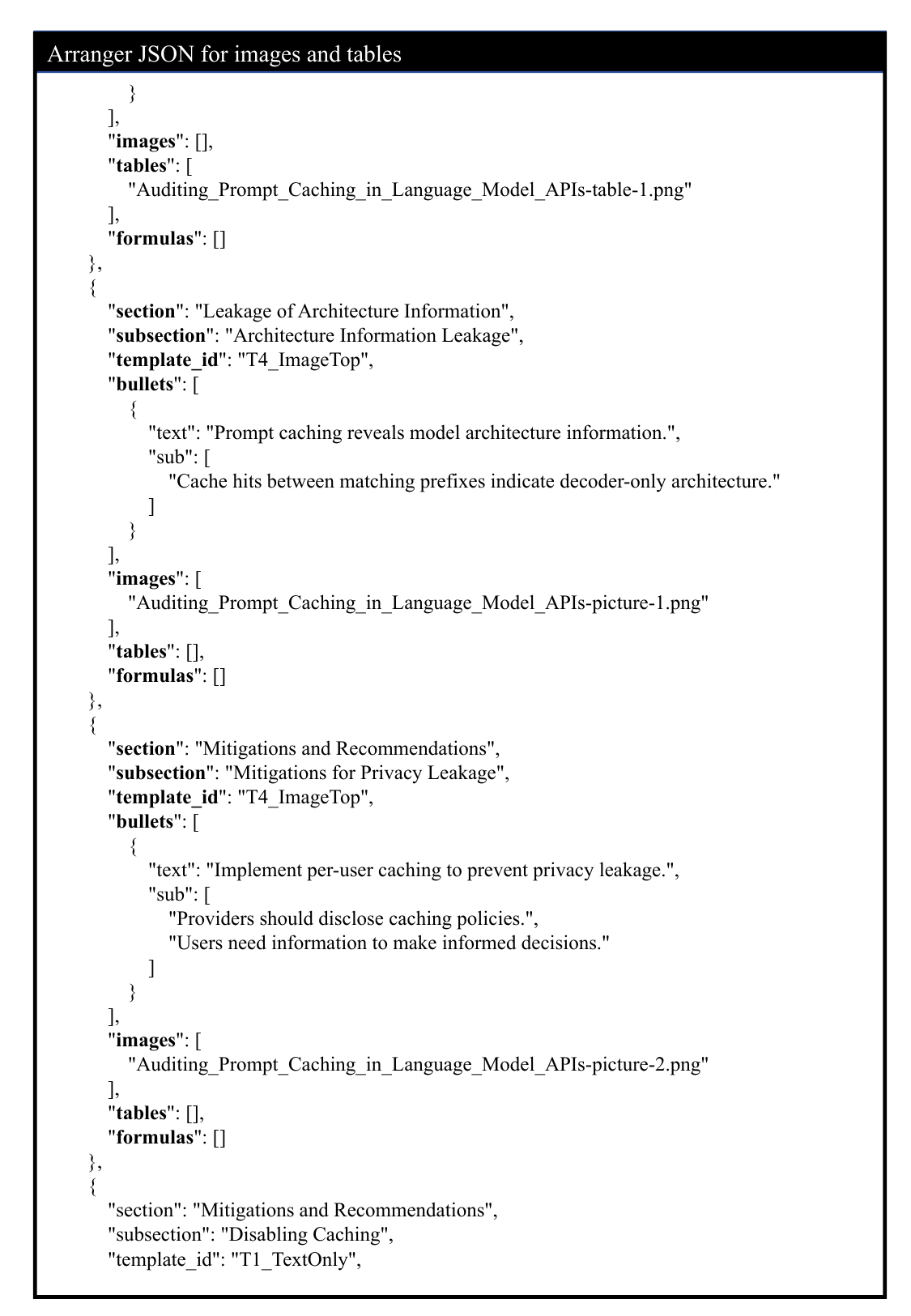}
    \caption{Output JSON of Arranger Agent.}
    \Description{}
    \label{fig:Arranger4}
\end{figure*}

\begin{figure*}[!tb]
    \centering
    \includegraphics[width=0.83\linewidth,page=5]{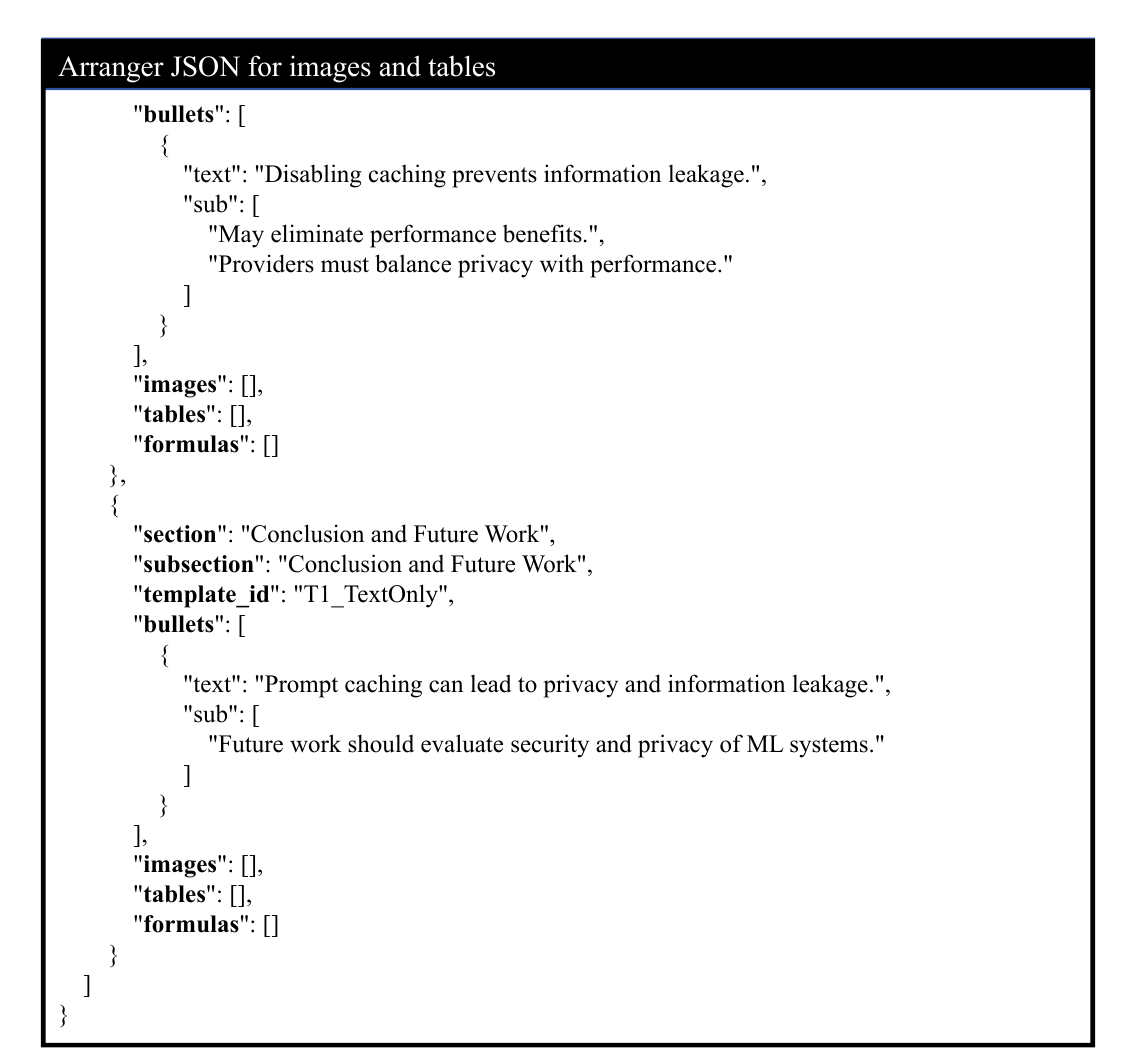}
    \caption{Output JSON of Arranger Agent.}
    \Description{}
    \label{fig:Arranger5}
\end{figure*}

\myparagraph{Arranger Output.}
\label{sec:arranger-example}
Arranger assigns a layout template to each slide and maps content elements (text blocks, images, formulas) to specific template slots. \cref{fig:Arranger1,fig:Arranger2,fig:Arranger3,fig:Arranger4,fig:Arranger5} show a complete example.

\begin{figure*}[!tb]
    \centering
    \includegraphics[width=0.83\linewidth,page=31]{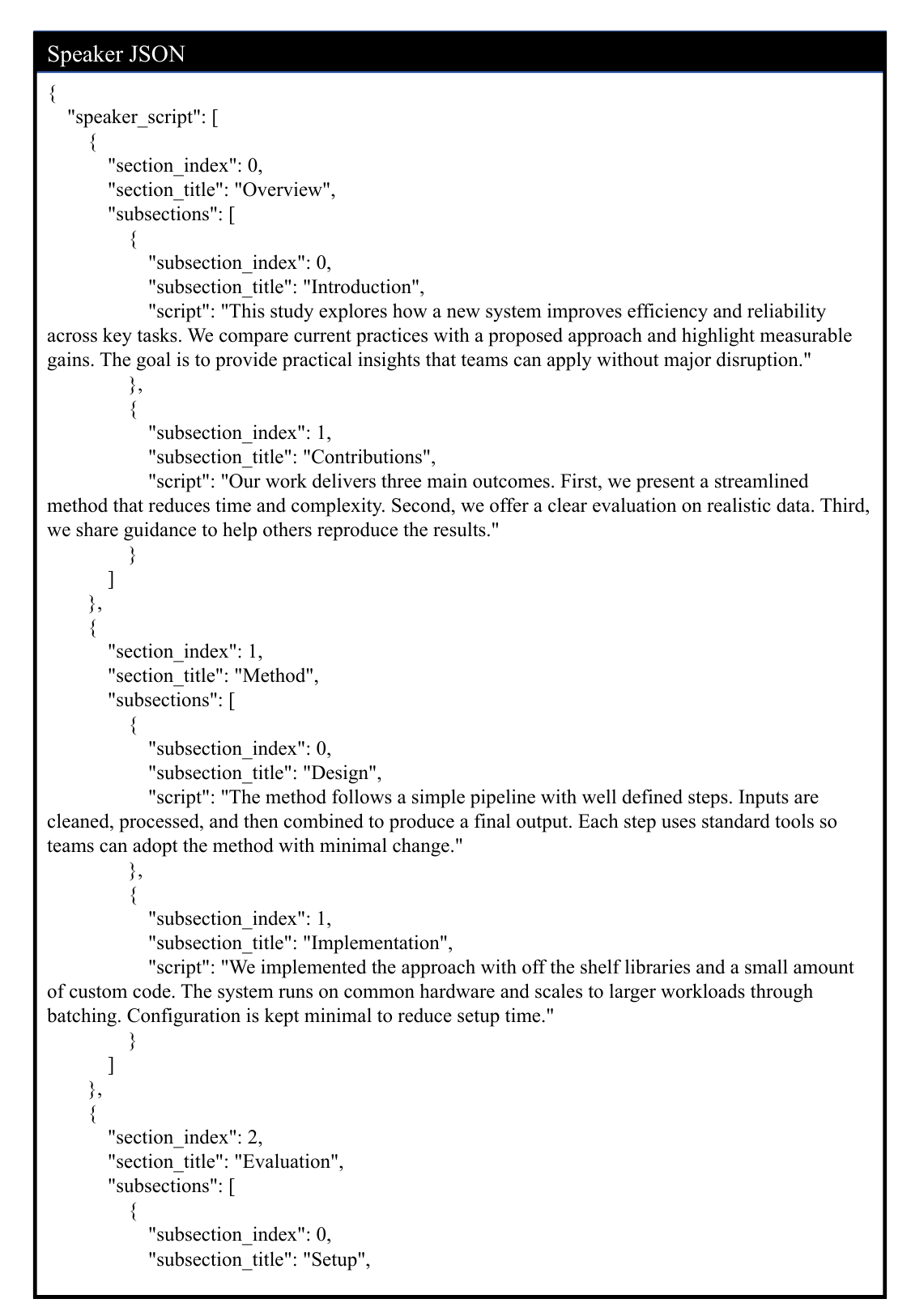}
    \caption{Output JSON of Speaker Agent.}
    \Description{}
    \label{fig:speaker1}
\end{figure*}

\begin{figure*}[!tb]
    \centering
    \includegraphics[width=0.83\linewidth,page=32]{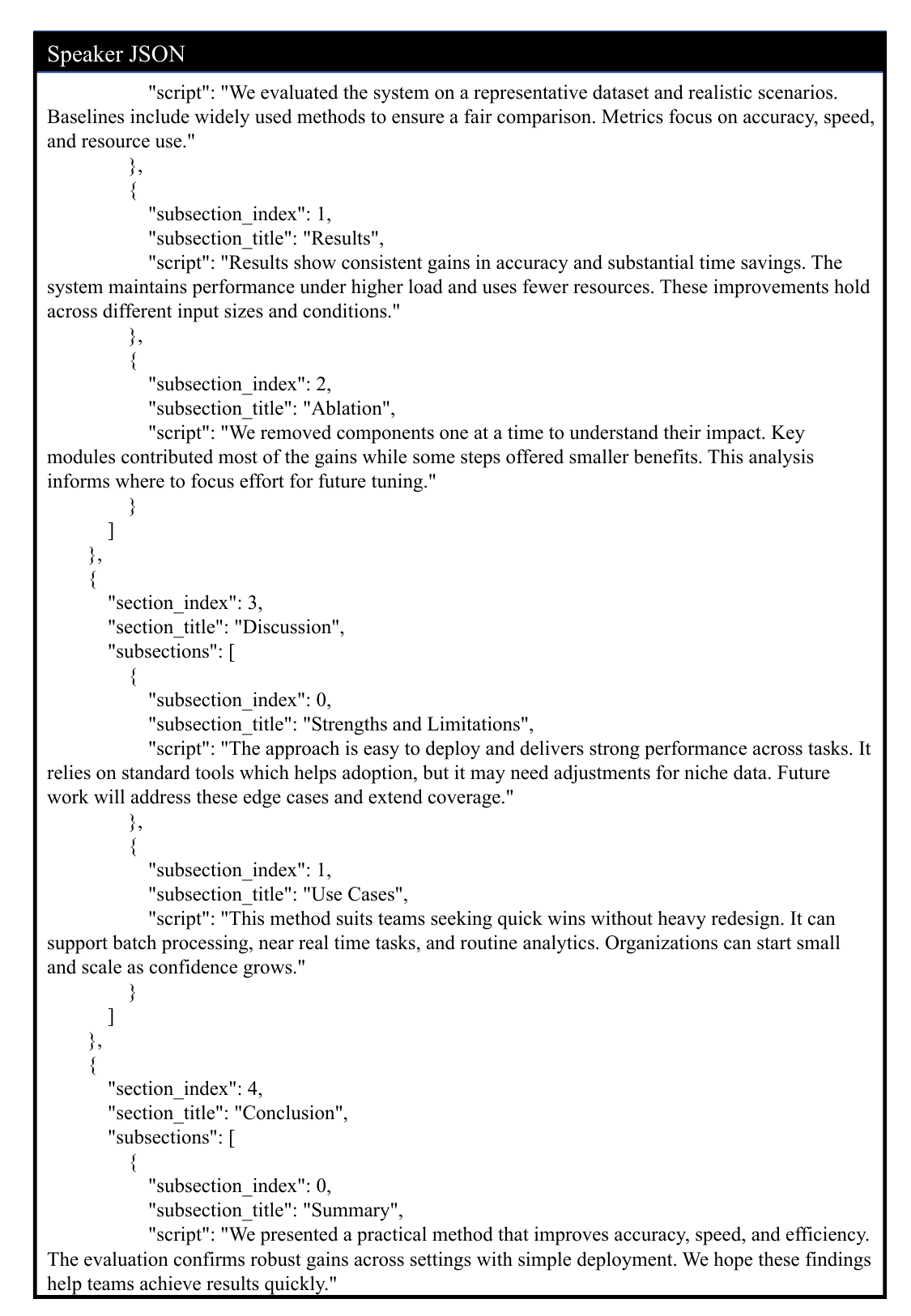}
    \caption{Output JSON of Speaker Agent.}
    \Description{}
    \label{fig:speaker2}
\end{figure*}

\begin{figure*}[!tb]
    \centering
    \includegraphics[width=0.83\linewidth,page=33]{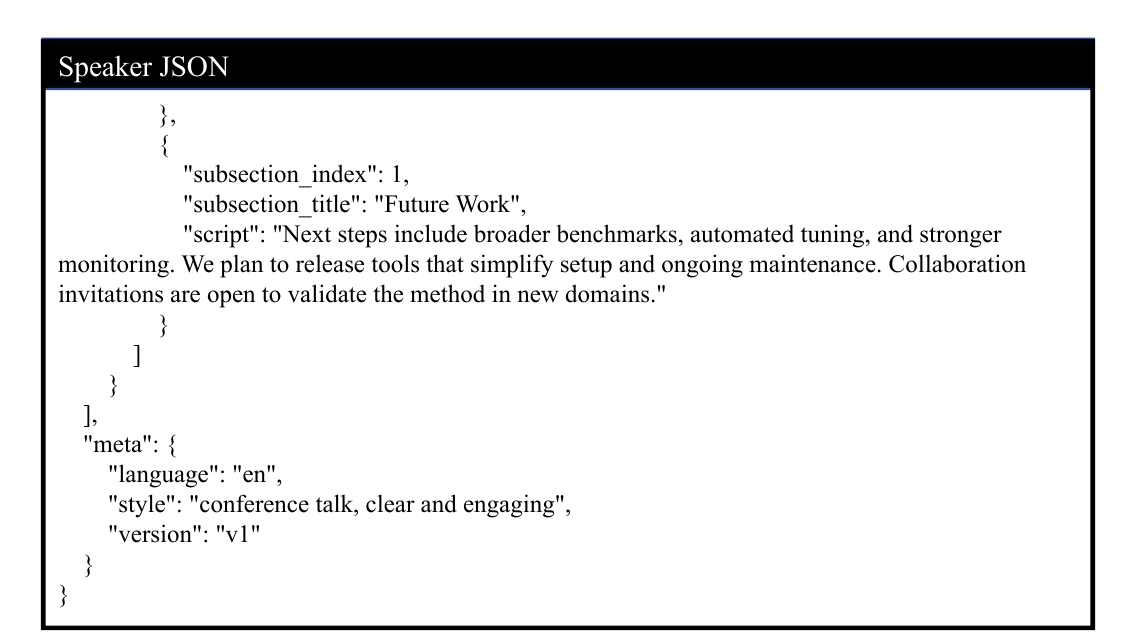}
    \caption{Output JSON of Speaker Agent.}
    \Description{}
    \label{fig:speaker3}
\end{figure*}

\myparagraph{Speaker Output.}
\label{sec:speaker-example}
Speaker generates one paragraph of presenter notes per slide, covering transitions, figure explanations, and narrative flow. \cref{fig:speaker1,fig:speaker2,fig:speaker3} show a complete example.

\begin{figure*}[!tb]
    \centering
    \includegraphics[width=0.83\linewidth,page=1]{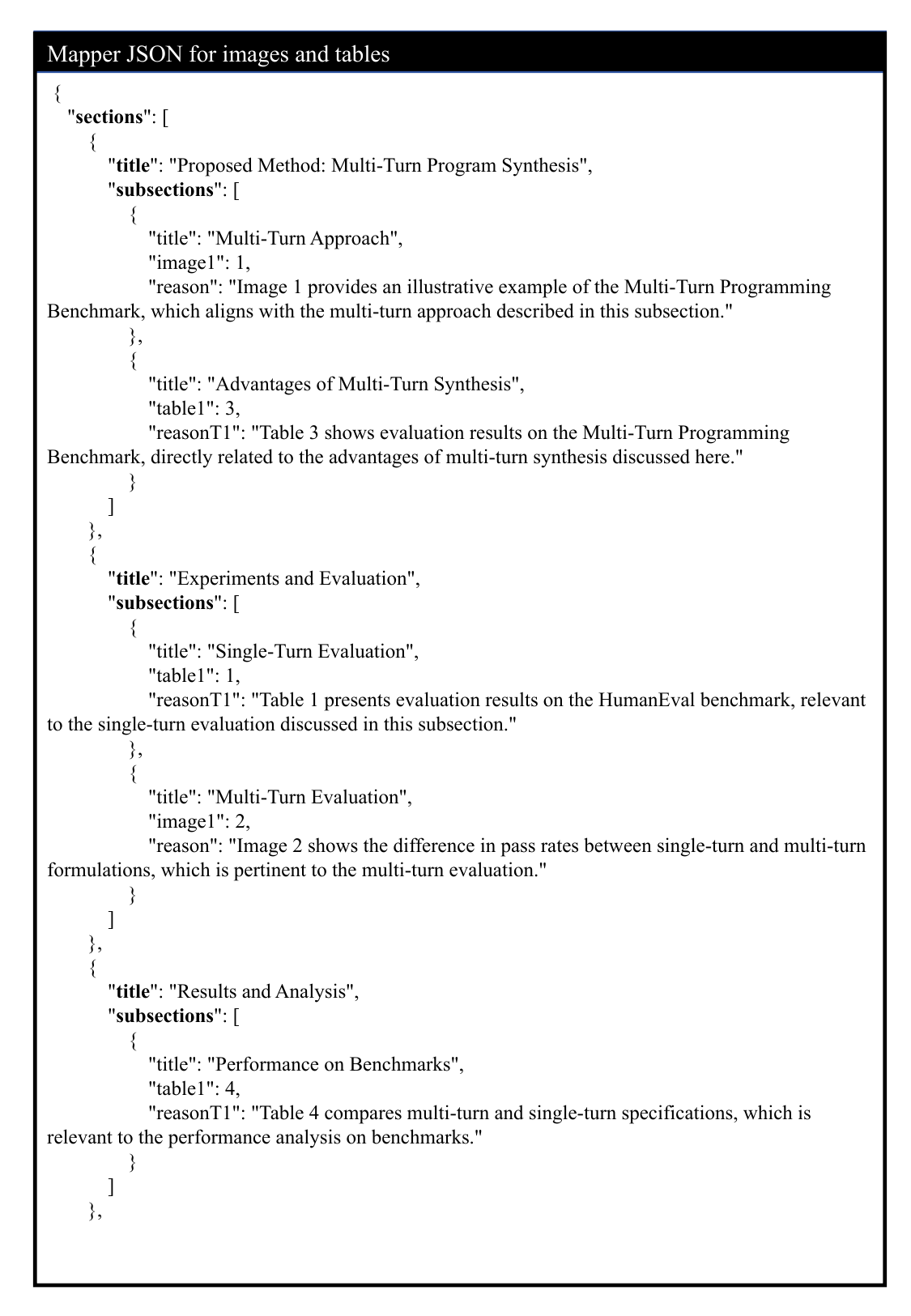}
    \caption{Output JSON of Mapper Agent.}
    \Description{}
    \label{fig:Mapper1}
\end{figure*}

\begin{figure*}[!tb]
    \centering
    \includegraphics[width=0.83\linewidth,page=2]{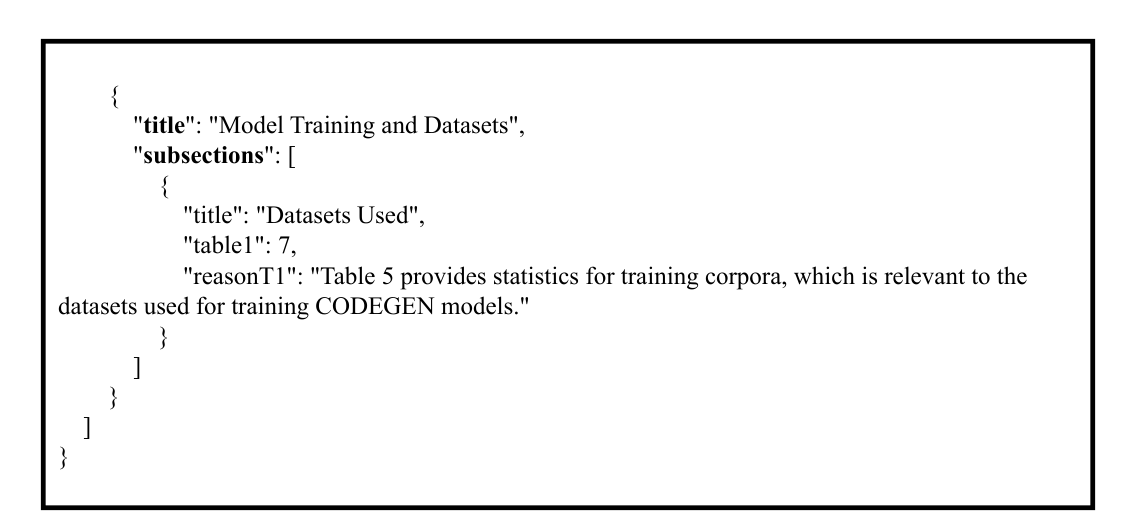}
    \caption{Output JSON of Mapper Agent.}
    \Description{}
    \label{fig:Mapper2}
\end{figure*}

\myparagraph{Mapper Output.}\label{sec:mapper-example}
Mapper lists figure-to-slide assignments, each accompanied by a short justification explaining why the figure supports the narrative. Figures~\ref{fig:Mapper1} and~\ref{fig:Mapper2} show a complete example.

\begin{figure*}[!tb]
    \centering
    \includegraphics[width=0.86\linewidth,page=1]{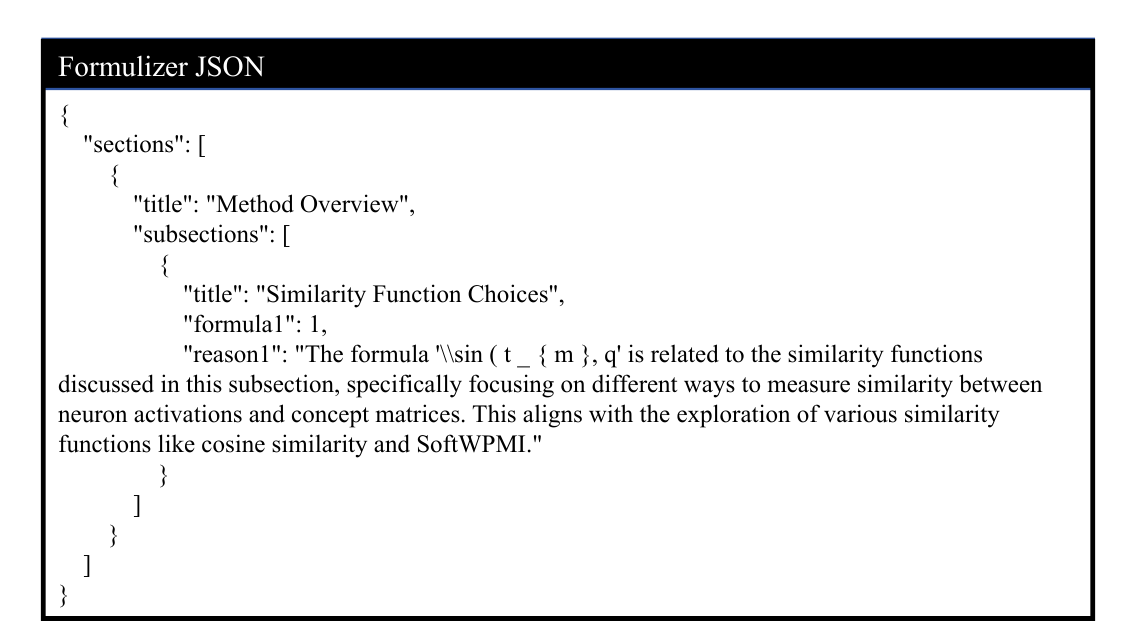}
    \caption{Output JSON of Formulizer Agent.}
    \Description{}
    \label{fig:Formulizer}
\end{figure*}

\myparagraph{Formulizer Output.}
\label{sec:formulizer-example}
Formulizer lists equation-to-slide assignments, pairing each cropped equation image with the slide whose context it supports. Figure~\ref{fig:Formulizer} shows a complete example.
 
\section{Color Setting}\label{sec:colorset}
Selecting good colors for slides involves both technical challenges (such as ensuring text is readable on a dark background) and aesthetic choices (such as creating a calm or professional look). To handle both, we use a simple three-step approach.

\textbf{Step 1: Human-defined baseline.}
We start with a set of baseline parameters that guide color adjustments in HSV space. These include the target saturation level, minimum and maximum saturation values, desired brightness for dark themes, and a fallback hue for colors that are nearly gray. We set this baseline once to ensure stable results: colors avoid becoming too bright, dull, or extreme. This baseline also sets "safe ranges" to prevent later changes from creating unusable colors.

\textbf{Step 2: LLM-based refinement.}
Next, we feed these baseline parameters, along with a short description in everyday language such as ``dark academic, calm, professional,'' into a GPT-based Color Refiner agent. The agent does not create final colors itself. Instead, it suggests small, style-focused tweaks to the parameters (for example, slightly reducing the brightness target or increasing the minimum saturation). These tweaks stay within the safe ranges and are applied only once at the start. This step brings in the language model's sense of style while keeping things controlled.

\textbf{Step 3: Deterministic color generation.}
With the refined parameters in place, all final slide colors are generated deterministically by our HSV adjustment algorithm. The algorithm is in~\cref{alg:color}. Given any input color, the algorithm always produces the same output. This means that while an LLM helps with style choices, the color creation process remains fully reproducible and based on clear math.

In summary, this combined method uses the strengths of language models for interpreting style descriptions, while making the overall slide generation reliable, consistent, and easy to repeat.
\begin{algorithm*}[t]
\caption{Detailed Color Movement Rules on the HSV Plane}\label{alg:color}
\KwIn{HSV $(H,S,V)$ with $S,V\in[0,1]$; parameters: \textup{\texttt{satTarget}}, \textup{\texttt{satFloor}}, \textup{\texttt{satCap}}, \textup{\texttt{satBlend}}, \textup{\texttt{targetV}}, \textup{\texttt{vCap}} (optional), \textup{\texttt{gamma}}, \textup{\texttt{fallbackHue}}}

\KwOut{Updated $(H,S,V)$}
\tcc*[l]{\textbf{Parameters:}}
\tcc*[l]{\textbf{satTarget} (0..1): target saturation you want to push toward. }
\tcc*[l]{\textbf{satFloor}  (0..1): minimum allowed saturation to avoid muddy gray. }
\tcc*[l]{\textbf{satCap}    (0..1): maximum allowed saturation to prevent neon look. }
\tcc*[l]{\textbf{satBlend}  (0..1): how strongly $S$ moves toward \textbf{satTarget} each step.  }
\tcc*[l]{\textbf{targetV}   (0..1): desired brightness “baseline” for a dark theme.  }
\tcc*[l]{\textbf{vCap}      (0..): brightness ceiling after darkening. }
\tcc*[l]{\textbf{gamma}     ($>0$): adaptive strength; the brighter above \textbf{targetV}, the more $V$ moves down.  }

\textbf{If the color is nearly gray} ($S \approx 0$): set $H$ to the hue of fallbackHue (\#2B5FA6) and raise $S$ to at least satFloor\;

\textbf{Move right (make it more vivid, but not too much):}\\
\Indp
\If{$S < \texttt{satTarget}$ \textbf{ or } $S < \texttt{satFloor}$}{
  $S \leftarrow (1-\texttt{satBlend})\cdot S + \texttt{satBlend}\cdot \texttt{satTarget}$\;
}
$S \leftarrow \mathrm{clamp}(S,\, \texttt{satFloor},\, \texttt{satCap})$\;
\Indm

\textbf{Move down (darken adaptively: the brighter it is, the more it moves):}\\
\Indp
\If{$V > \texttt{targetV}$}{
  $d \leftarrow V - \texttt{targetV}$\;
  $a \leftarrow 1 - e^{-\texttt{gamma}\cdot d}$ \tcp*[r]{bigger gap $\Rightarrow$ stronger pull}
  $V \leftarrow V - a\cdot d$ \tcp*[r]{pull $V$ toward \textup{\texttt{targetV}}}
  \If{\texttt{vCap} is set}{ $V \leftarrow \min(V,\, \texttt{vCap})$ }
}
\Indm

\textbf{Optional gentle lift (avoid being too dark):}\\
\Indp
Let $\texttt{vFloor} = \texttt{targetV} - 0.02$. \\
\If{$V < \texttt{vFloor}$}{ $V \leftarrow 0.7\cdot V + 0.3\cdot \texttt{vFloor}$ }
\Indm

\Return $(H,S,V)$\;
\end{algorithm*}